\documentclass{article}

\usepackage{iclr2027_conference,times}
\iclrpreprintcopy 
\usepackage{hyperref}
\usepackage{url}

\usepackage[table]{xcolor}

\usepackage{amsmath}
\usepackage{amsfonts}       
\usepackage{amsthm}         
\usepackage{mathtools}
\usepackage{nicefrac}       
\usepackage[super]{nth}
\usepackage{siunitx}
\usepackage{bm}
\usepackage{pifont}

\usepackage{booktabs}       
\usepackage{tabularx}
\usepackage{makecell}

\usepackage{adjustbox}
\usepackage{svg}
\usepackage{microtype}
\usepackage{graphicx}
\usepackage[font=small]{caption}
\usepackage{subfigure}

\newcommand{\cmark}{\ding{51}}
\newcommand{\xmark}{\ding{55}}

\usepackage[ruled]{algorithm2e} 

\SetAlFnt{\small}
\SetAlCapFnt{\small}
\SetAlCapNameFnt{\small}
\SetAlCapHSkip{0pt}

\usepackage{acro}
\newcommand{\newac}[2]{\DeclareAcronym{#1}{short=#1,long=#2}}
\newac{CAMDM}         {Conditional Autoregressive Motion Diffusion Model}
\newac{CEM}           {Cross-Entropy Method}
\newac{CFG}           {Classifier-Free Guidance}
\newac{EOM}           {Equations of Motion}
\newac{DDPM}          {Denoising Diffusion Probabilistic Model}
\newac{DOF}           {Degrees of Freedom}
\newac{FID}           {Fréchet Inception Distance}
\newac{GD}            {Gradient Descent}
\newac{ID}            {Inverse Dynamics}
\newac{LDJ}           {Log Dimensionless Jerk}
\newac{MDM}           {Motion Diffusion Model}
\newac{MPC}           {Model Predictive Control}
\newac{MSE}           {Mean Squared Error}
\newac{RL}            {Reinforcement Learning}
\newac{RMSE}          {Root Mean Squared Error}
\newac{SAT}           {Separating Axis Theorem}
\newac{Sep-CMA-ES}    {Separable Covariance Matrix Adaptation Evolution Strategy}
\newac{SMPL}          {Skinned Multi-Person Linear}
\newac{SOCP}          {Second-Order Cone Program}
\newac{WE}            {Whittaker–Eilers}

\newcommand{\figref}[1]{Fig.~\ref{#1}}
\newcommand{\secref}[1]{Sec.~\ref{#1}}
\newcommand{\algref}[1]{Alg.~\ref{#1}}
\newcommand{\eqnref}[1]{Eq.~\eqref{#1}}
\newcommand{\tabref}[1]{Tab.~\ref{#1}}

\DeclareMathOperator{\Log}{Log}

\newif\ifshowtodos
\showtodosfalse

\title{PhysibleMDM: Enforcing Inference-Time Physical Feasibility Constraints on Motion Diffusion Models via Noise Optimization}

\author{
  Maximilian St\"olzle \\
  Disney Research\\
  MIT CSAIL, USA
  \And
  David M\"uller \\
  Disney Research\\
  Switzerland
  \And
  Agon Serifi \\
  Disney Research\\
  Switzerland
  \AND
  Ruben Grandia \\
  Disney Research\\
  Switzerland
  \And
  Espen Knoop \\
  Disney Research\\
  Switzerland
  \And
  Moritz B\"acher \\
  Disney Research\\
  Switzerland
}

\begin{document}
\maketitle


\begin{abstract}
    Motion Diffusion Models (MDMs) can generate diverse and visually appealing human motion; yet, the resulting trajectories frequently violate physical constraints, limiting their downstream use in simulation and control. We present \emph{PhysibleMDM}, a framework that promotes physical feasibility of MDM output by optimizing at \emph{inference time} the diffusion noise \emph{without} retraining the generative model.
    We introduce a set of surrogate feasibility constraints that can be evaluated without full trajectory simulation. These constraints combine kinematic checks (e.g., ground penetration, slip, and self-collision) with dynamic feasibility tests computed via an inverse-dynamics inner loop formulated as a cone program. We solve the resulting black-box noise optimization problem using the Separable Covariance Matrix Adaptation Evolution Strategy (Sep-CMA-ES), thereby avoiding reliance on computationally expensive and discontinuous gradients. An autoregressive scheme then chains optimized segments to synthesize long motion sequences. Across 100 motion styles on an SMPL humanoid, PhysibleMDM reduces mean ground penetration, self-collision penetration, and ground slip by 75\%, 92\%, and 74\%, respectively, relative to unoptimized diffusion output. Across 11 sequences for a custom bipedal robot, it achieves median reductions of 89.5\% in per-sequence mean acceleration-reconstruction error and 79.5\% in the mean inverse-dynamics residual.
\end{abstract}

\newlength{\mainpapertextfloatsep}
\newlength{\mainpaperfloatsep}
\newlength{\mainpaperintextsep}
\setlength{\mainpapertextfloatsep}{\textfloatsep}
\setlength{\mainpaperfloatsep}{\floatsep}
\setlength{\mainpaperintextsep}{\intextsep}
\setlength{\textfloatsep}{11pt plus 2pt}
\setlength{\floatsep}{11pt plus 2pt}
\setlength{\intextsep}{11pt plus 2pt}

\section{Introduction}
\acp{MDM}~\citep{tevet2023human,zhang2024motiondiffuse,kim2023flame,rempe2026kimodo} currently achieve state-of-the-art performance in synthesizing human-like motion for physics-based characters and robots. These generators are typically trained on large corpora of human motion capture data and can be conditioned on various signals, including text, style, goals, and semantic tasks~\citep{dabral2023mofusion, chen2024taming, huang2025diffuse, zhao2026ardy}. The generated motion commonly serves as a reference for \ac{RL}-based trackers~\citep{peng2018deepmimic, luo2023perpetual, serifi2024vmp, liao2026beyondmimic, luo2026sonic}. Errors can arise at multiple stages of this pipeline. In particular, generated motions may exhibit \emph{physical artifacts} that hinder their execution. \emph{Kinematic} artifacts include joint-limit violations~\citep{lee2025phuma}, self-collisions~\citep{khazoom2022humanoid}, ground penetration~\citep{yuan2023physdiff}, undesired ground slip~\citep{pan2025spider}, and foot placement in non-traversable regions~\citep{nguyen20163d}. \emph{Dynamic} artifacts include motions requiring joint torques beyond actuator limits~\citep{herzog2014balancing} or contact forces outside friction cones~\citep{grandia2021multi}, as well as trajectories inconsistent with the dynamics of unactuated coordinates, such as the floating base of a legged robot~\citep{herzog2014balancing, grizzle2014models}. These artifacts degrade downstream tracking performance~\citep{araujo2026retargeting, zhang2026physmodpo}, as infeasible reference motions cannot be reliably tracked.

Several methods mitigate these artifacts before or during \emph{training}, for example, by learning from motions tracked in physics simulation~\citep{wu2025uniphys, watanabe2025simdiff, xu2025parc}, or fine-tuning the generator with physics-informed objectives~\citep{serifi2024robot, zhang2026physmodpo}.
While effective in practice, these approaches can be resource-intensive, particularly when generative models and tracking policies are retrained across refinement iterations, and they do not inherently guarantee that all samples from the final diffusion model satisfy the desired physical constraints.

\begin{figure}[t]
    \centering
    \includegraphics[width=0.85\linewidth]{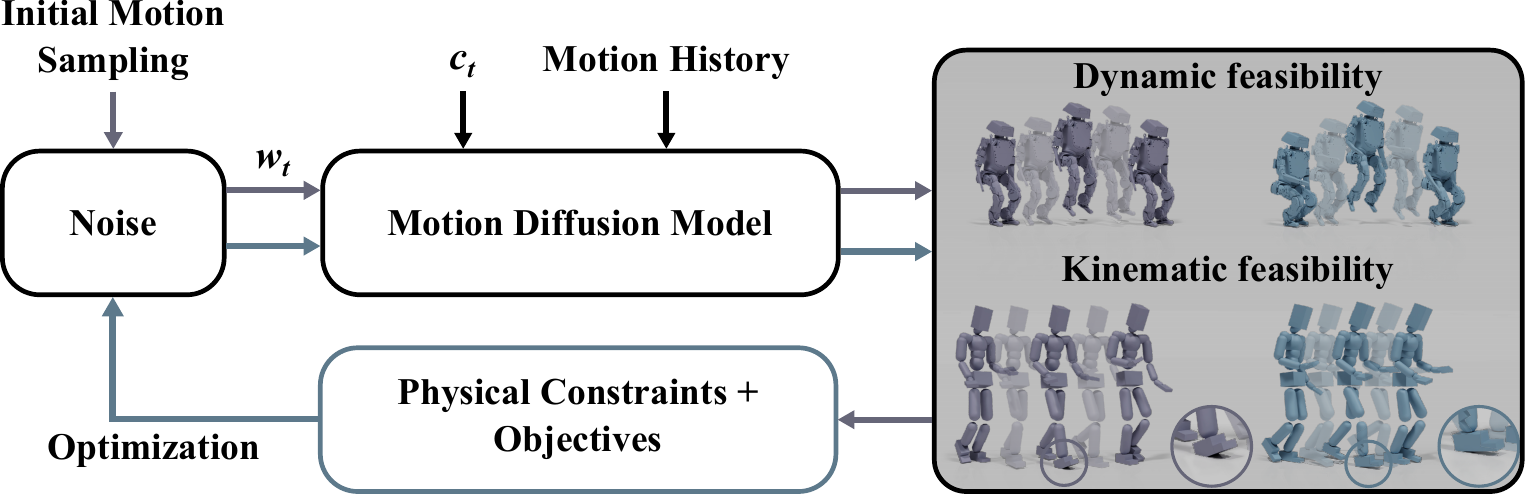}
    \vspace{-0.1cm}
    \caption{\textbf{Overview of PhysibleMDM.} We improve physical feasibility by optimizing the diffusion noise of a frozen motion generator. Kinematic constraint violations and constrained inverse-dynamics residuals serve as physical-feasibility proxies, guiding refinement without simulator rollouts. Initial (purple) and optimized (blue) motions illustrate improvements in dynamic feasibility (top) and kinematic feasibility (bottom). Optimized motion history conditions subsequent generation.}
    \label{fig:overview}
    \vspace{-0.2cm}
\end{figure}

Inference-time optimization incorporates explicit objectives and constraints during generation without retraining the diffusion model. Common approaches steer intermediate denoising states through gradient guidance~\citep{dhariwal2021diffusion,karunratanakul2023guided} or projection onto feasible sets~\citep{yuan2023physdiff,xiao2025safediffuser,watanabe2026projflow}. Noise optimization offers an alternative in which objectives and constraints are evaluated on fully denoised motions~\citep{karunratanakul2024optimizing,zhao2025dartcontrol}. Gradient-based noise optimization requires costly backpropagation through both the diffusion sampler and the physical-feasibility evaluation. Contact-rich motion further introduces non-smooth objectives and constraints for which gradients may be poorly informative~\citep{antonova2023rethinking,paulus2026hard}. Sampling-based optimization has shown promise for optimal control of contact-rich legged locomotion~\citep{xue2025fullorder,alvarez2025real,schramm2026referencefree}. Recent work also applies sampling-based optimization to diffusion noise for image generation and protein design~\citep{kim2026zeno,tan2026dualseed}, but does not address the kinematic and dynamic feasibility requirements of robotic motion. Efficient gradient-free optimization of initial and posterior diffusion noise for these requirements, while retaining characteristic motion style, remains unexplored.


To address this gap, we propose PhysibleMDM, a gradient-free method that jointly optimizes initial and posterior diffusion noise at inference time. Parallel forward passes through the frozen model generate a population of motions for evaluating task objectives, kinematic constraints, and constrained inverse-dynamics residuals. These evaluations guide \ac{Sep-CMA-ES}~\citep{ros2008simple}, whose diagonal covariance supports high-dimensional search, in iteratively refining the noise population. Noise standardization regularizes noise toward its training-time statistics to help limit deviations from the learned motion distribution. An autoregressive scheme chains optimized segments into long motion sequences. 


In summary, the contributions of this paper are:
\begin{itemize}
    \item A physical-feasibility formulation for \acp{MDM} that combines kinematic constraints with constrained inverse-dynamics residuals, enabling inference-time refinement without a tracking controller or full trajectory simulation.
    \item A scalable sampling-based method for high-dimensional noise optimization in \acp{MDM} (realized via \ac{Sep-CMA-ES}), evaluating non-smooth objectives and constraints on fully denoised motions without costly gradient computation.
    \item A diffusion-noise optimization strategy, applicable to any pretrained MDM without architecture-specific changes, that combines joint initial and posterior noise optimization with noise standardization, balancing optimization flexibility against deviations from the model's learned distribution and supporting motion-style retention in \acp{MDM}.
\end{itemize}

\section{Related Work}
\label{sec:related_work}

\paragraph{Inference-Time Optimization}
Guidance and projection steer diffusion sampling by modifying intermediate denoising states or their clean-sample estimates~\citep{karunratanakul2023guided,christopher2024constrained,bodmer2026grounding}. Noise optimization instead evaluates objectives on fully denoised samples and adjusts the stochastic inputs, using gradients~\citep{karunratanakul2024optimizing,zhao2025dartcontrol} or sampling-based updates~\citep{kim2026zeno,tan2026dualseed}. The latter support non-differentiable objectives without backpropagating through the denoising process. Relative to these sampling-based methods, our approach jointly optimizes initial latent and posterior noise with Sep-CMA-ES, whose diagonal covariance yields linear internal time and memory scaling with the number of noise variables~\citep{ros2008simple}, making it suitable for this high-dimensional search.

\paragraph{Physical Feasibility of Generated Motion}
Existing methods incorporate physics at different stages of motion generation. Before training, motion data can be refined through kinematic-constraint-aware retargeting~\citep{yang2026omniretarget} or collected through simulated tracking, with the latter sometimes repeated as the generator and tracker improve over successive training rounds~\citep{watanabe2025simdiff,xu2025parc}. Physical objectives can also guide learning directly through dynamics-consistency losses~\citep{serifi2024robot,xing2026interphys} or preference-based post-training~\citep{zhang2026physmodpo}. Other approaches introduce physics during generation: PhysDiff projects intermediate denoising states through simulated execution~\citep{yuan2023physdiff}, while CLoSD alternates motion planning and execution~\citep{tevet2025closd}. Alongside this choice of when to incorporate physics, methods differ in what they predict. Whereas most motion diffusion models generate kinematic state trajectories, joint state--action models also generate controls, learning both from trajectories collected with tracking controllers~\citep{huang2025diffuse,wu2025uniphys}. However, incorporating physics before or during training does not ensure feasibility of individual generated samples, particularly under distribution shift. Predicting controls alongside states does not remove this limitation, since the predicted states may differ from the motion produced by executing those controls. Furthermore, approaches that use simulated tracking depend on a trained tracking controller. To address these limitations, our method evaluates kinematic constraints and constrained inverse-dynamics residuals on each fully generated motion and optimizes diffusion noise at inference time, requiring neither generator retraining nor tracking-controller rollouts.

Further related work appears in Appendix~\ref{apx:sec:related_work}.

\section{Preliminaries: Motion Diffusion Models}
\label{sec:preliminaries_ddpm}

\paragraph{Motion Representation}

Let $t\geq 0$ denote the current physical frame, $F$ the number of predicted future frames, and $P$ the number of past frames supplied as conditioning. With inclusive ranges, we write
\begin{equation}
 \mathbf M_{\mathrm{fut},t}=(\mathbf m^{\mathrm{enc}}_{t+1},\ldots,\mathbf m^{\mathrm{enc}}_{t+F}),\qquad
 \mathbf M_{\mathrm{pst},t}=(\mathbf m^{\mathrm{enc}}_{t-P+1},\ldots,\mathbf m^{\mathrm{enc}}_t).
\end{equation}
Each $\mathbf m^{\mathrm{enc}}_\ell\in\mathbb R^{n_{\mathrm{enc}}}$ is a full encoded motion feature vector; stacking frames gives $\mathbf M_{\mathrm{fut},t}\in\mathbb R^{F\times n_{\mathrm{enc}}}$. The local future-frame index is $f=1,\ldots,F$, with absolute frame $\ell=t+f$.

For a floating-base character, decoding $\mathbf m^{\mathrm{enc}}_\ell$ gives the motion feature tuple $\mathbf m_\ell=(\mathbf x_\ell,\mathbf r_\ell,\mathbf J_\ell,\mathbf X_\ell,\mathbf R_\ell)$, where $\mathbf x_\ell\in\mathbb R^3$ is the root position, $\mathbf r_\ell\in\mathbb R^6$ is its orientation feature, and $\mathbf J_\ell\in\mathbb R^{J\times6}$ contains the $J$ joint orientations relative to their parents. The link position and orientation features are $\mathbf X_\ell\in\mathbb R^{n_{\mathrm b}\times3}$ and $\mathbf R_\ell\in\mathbb R^{n_{\mathrm b}\times6}$, where $n_{\mathrm b}$ counts rigid bodies. Orientations use the continuous 6D representation~\citep{zhou2019continuity}. Flattening this tuple gives $n_m=9+6J+9n_{\mathrm b}$ features. The full encoding uses six entries per position or orientation block, giving $n_{\mathrm{enc}}=6(2+J+2n_{\mathrm b})$. 

\paragraph{Motion Diffusion Models}

Motion diffusion models generate motion by iteratively denoising a noisy sequence conditioned on past motion, task commands, and style. We consider a frozen \ac{DDPM}~\citep{ho2020denoising} whose denoiser $\mathcal M_\theta$ predicts clean motion from the noisy sequence, diffusion level, and conditioning $\mathbf c_t$. Superscripts $d=0,\ldots,D$ index diffusion levels from clean motion ($d=0$) to initial noise ($d=D$); hats denote clean-motion estimates. Sampling starts from $\mathbf M^D_{\mathrm{fut},t}$ and injects posterior noise $\mathbf Z_d$ at reverse steps with noise scale $\sigma_d^{\mathrm{diff}}>0$. These noise tensors match the future motion's shape and have independent standard-normal entries before optimization. Fixing all injected noise and conditioning makes the complete sampler deterministic, as formalized in \secref{sec:noise_optimization_mdm}. Appendix~\ref{apx:sec:motion_diffusion_model} gives the forward and reverse processes.

\section{Overview}
\label{sec:overview}
PhysibleMDM refines generated motion at inference time by optimizing diffusion noise under task objectives and physical-feasibility penalties (\figref{fig:overview}). Starting from standard-normal noise, we standardize each candidate, generate a complete motion through the frozen \ac{MDM}, and evaluate its kinematic violations and constrained inverse-dynamics residuals. Batched \ac{Sep-CMA-ES}~\citep{ros2008simple} updates the candidate population over a fixed iteration budget. We then commit the first part of the best motion and repeat autoregressively, conditioning on the preceding optimized motion and encouraging consistency between overlapping predictions.

\section{Diffusion Noise Optimization}
\label{sec:noise_optimization_mdm}

We jointly optimize initial and posterior diffusion noise to provide additional control over the denoising trajectory, increasing flexibility to reduce constraint violations and improve task performance. To help preserve motion style, we standardize the noise produced by the optimizer to have zero mean and unit variance, matching the corresponding moments of the model’s standard-normal sampling distribution. This regularizes the search to help limit deviations from the learned motion distribution.

\paragraph{Joint Initial and Posterior Noise Parameterization}
\label{sec:noise_parameterization}
For a frozen model and fixed conditioning, the generated sample is fully determined by the specific sequence of injected noise values. We use these noise values as decision variables.

We collect the initial noise and all $D$ allocated posterior-noise tensors into the sampling-noise vector
\begin{equation}
 \mathbf w=\operatorname{concat}\!\left(\operatorname{vec}(\mathbf M^D_{\mathrm{fut},t}),
 \{\operatorname{vec}(\mathbf Z_d)\}_{d=1}^{D}\right)
 \in\mathbb R^{n_w},\qquad n_w=(D+1)Fn_{\mathrm{enc}}.
\end{equation}
The sampler injects posterior noise only at steps with $\sigma_d^{\mathrm{diff}}>0$. The complete reverse sampler is the deterministic map $\mathbf M^0_{\mathrm{fut},t}=\Phi_\theta(\mathbf w,\mathbf c_t)$.
Here $\Phi_\theta$ comprises all reverse transitions; $\mathcal M_\theta$ denotes a single denoiser evaluation.

\paragraph{Noise Standardization}
\label{sub:noise_norm}
We optimize the unnormalized noise $\mathbf w^{\mathrm{raw}}\in\mathbb R^{n_w}$. For each candidate, we standardize its initial noise and combined posterior noise separately. Let $b\in\{\mathrm{init},\mathrm{post}\}$ index these two groups, with empirical mean $\mu_b$ and sample standard deviation $s_b$ over all entries of $\mathbf w_b^{\mathrm{raw}}$. The sampling noise is
\begin{equation}
\mathbf w=\Psi(\mathbf w^{\mathrm{raw}})
=\operatorname{concat}_{b}\!\left(
\frac{\mathbf w_b^{\mathrm{raw}}-\mu_b}{\max(s_b,\varepsilon_{\mathrm{norm}})}
\right),\qquad \varepsilon_{\mathrm{norm}}>0.
\label{eq:noise_norm}
\end{equation}
The posterior statistics are shared across all $D$ diffusion steps. Correlations and higher-order statistics may still change during optimization.

\paragraph{Diffusion Noise Optimization Problem}
\label{sub:diffusion_noise_optim}

We optimize the noise and evaluate objectives and constraints on each fully denoised sample:
\begin{equation}
\begin{aligned}
 \min_{\mathbf w^{\mathrm{raw}}}\quad &\mathcal J(\mathbf M)\\
 \text{s.t.}\quad &\mathbf M=\Phi_\theta(\Psi(\mathbf w^{\mathrm{raw}}),\mathbf c_t),\\
 &\mathbf h(\mathbf M)=\mathbf0,\qquad \mathbf g(\mathbf M)\leq\mathbf0.
\end{aligned}
\label{eq:noise-opt}
\end{equation}
Here, $\mathcal J$ is the task objective, and $h_j$, $g_j$ are the scalar components of the stacked equality and inequality constraints. Positive values of $g_j$ indicate violations.
Physical constraints follow in \secref{sec:physical_feasibility_constraints}, with task objectives in Appendix~\ref{apx:sub:task_objectives}.

\section{Physical Feasibility Refinement of Generated Motion}
\label{sec:physical_feasibility_constraints}

Under the assumed robot and contact model, physical feasibility requires valid geometry and consistent contact kinematics, together with accelerations realizable by admissible joint torques and contact forces. We encode these requirements as kinematic and dynamic constraints in \eqnref{eq:noise-opt}. An inner \ac{ID} solve enforces actuation and contact-force limits and quantifies the remaining dynamic mismatch in acceleration or generalized-force coordinates for each generated motion at inference time. This provides a proxy for physical feasibility without simulator rollouts or a tracking controller.

\subsection{Robot Configuration and Dynamics}

At each generated frame $\ell$, extracting the physical configuration from $\mathbf m_\ell$  gives $\mathbf s_\ell=(\mathbf x_\ell,\mathbf R_{\mathrm{root},\ell},\mathbf q_\ell)$, comprising root position, root orientation, and joint coordinates. Generalized velocities $\bm\nu_\ell=(\mathbf v_\ell,\bm\omega_\ell,\mathbf v_{\mathrm{joint},\ell})$ and accelerations $\dot{\bm\nu}_\ell$ lie in $\mathbb R^{n_\nu}$, where $n_\nu=6+n_{\mathrm{joint}}$ and $n_{\mathrm{joint}}$ counts joint velocity degrees of freedom. Root positions and velocities use world coordinates; contact geometry follows from forward kinematics.

Of $n_{\mathrm c}$ candidate contact points, $\mathcal C_\ell$ denotes the active set, modeled as sticking: $\mathcal C_\ell=\mathcal C_\ell^{\mathrm{stick}}$. Joint torques and stacked 3D contact forces are denoted by $\bm\tau$ and $\bm\lambda$, respectively. Appendix~\ref{apx:sub:robot_motion_model} details coordinates, dynamics, and contact forces. Smoothing and derivative estimation follow Appendix~\ref{par:whittaker_eilers}.

\subsection{Kinematic Feasibility Constraints}\label{sub:kinematic_feasibility_constraints}

We write kinematic constraints as functions of the generated motion $\mathbf M$ through its physical configurations and derivatives.

\paragraph{Ground Penetration}\label{ssub:ground_penetration}
Let $\delta_{\ell,p}\in\mathbb R$ be signed terrain separation at checked contact point $p$, with negative values indicating penetration. The pointwise nonpenetration condition is $g_{\mathrm{pen},\ell,p}(\mathbf M)=-\delta_{\ell,p}\leq0$ for $p=1,\ldots,n_{\mathrm c}$.

\paragraph{Ground Slip}\label{ssub:contact_slip}
Let $\mathbf J_{\mathrm c,\parallel,p}(\mathbf s)\in\mathbb R^{2\times n_\nu}$ be the contact point's translational Jacobian projected onto the contact tangent plane. Its tangential velocity defines the pointwise sticking condition
\begin{equation}
\mathbf h_{\mathrm{slip},\ell,p}(\mathbf M)
=\mathbf v_{\parallel,p}(\mathbf s_\ell,\bm\nu_\ell)
=\mathbf J_{\mathrm c,\parallel,p}(\mathbf s_\ell)\bm\nu_\ell
=\mathbf0,\qquad p\in\mathcal C_\ell^{\mathrm{stick}}.
\label{eq:h_slip}
\end{equation}
We measure ground penetration and slip by averaging penetration depths and active-contact tangential speeds over all $n_{\mathrm c}$ candidate points (Appendix~\ref{apx:par:ground_violation_aggregation}). On flat ground, tangential velocity comprises the horizontal components, and contact activation is $a_{\ell,p}=\mathbf 1[\delta_{\ell,p}\leq0]$.

\paragraph{Contact Pattern Preservation}\label{ssub:contact_pattern_preservation}
Let $a^{\mathrm{ref}}_{\ell,p}\in\{0,1\}$ denote a reference contact pattern inferred from unoptimized samples and held fixed during each optimization horizon (Appendix~\ref{apx:sub:contact_pattern_identification}). Let $\delta_{\mathrm{pen}}\geq0$ be the allowed penetration and $\delta_{\mathrm{clear}}\geq0$ the clearance required for inactive contacts. With $[x]_+=\max(x,0)$, the pointwise constraint $g_{\mathrm{contact},\ell,p}\leq0$ uses
\begin{equation}
g_{\mathrm{contact},\ell,p}
=a^{\mathrm{ref}}_{\ell,p}
\bigl([\delta_{\ell,p}]_++[-\delta_{\ell,p}-\delta_{\mathrm{pen}}]_+\bigr)
+(1-a^{\mathrm{ref}}_{\ell,p})[\delta_{\mathrm{clear}}-\delta_{\ell,p}]_+.
\label{eq:contact_pattern_preservation}
\end{equation}
This penalizes separation and excessive penetration during expected contact, and insufficient clearance during expected flight. Averaging over contact points gives the per-frame violation $v_{\mathrm{contact},\ell}$.

\subsection{Dynamic Feasibility Constraints}
Dynamic feasibility requires existence of joint torques $\bm\tau$ and contact forces $\bm\lambda$ that realize the prescribed motion under three conditions. \emph{Underactuated dynamics:} the motion must satisfy the equations of motion for the unactuated coordinates, including the floating base; joint torques cannot directly compensate for an imbalance in these equations. \emph{Actuation limits:} joint torques must stay within actuator bounds. \emph{Contact limits:} normal forces must be nonnegative ($\lambda_\perp\geq0$), and tangential forces must lie within the friction cone for the assumed sticking mode; no-slip kinematics are checked separately. Admissible torques and contact forces alone therefore do not establish that the prescribed acceleration is realizable.

\subsubsection{Constrained Inverse Dynamics}\label{sub:id_interface}
For each planned state and its derivatives, constrained \ac{ID} returns joint torques, contact forces, and reconstructed acceleration as $(\bm\tau_\ell^{\mathrm{ID}},\bm\lambda_\ell^{\mathrm{ID}},\dot{\bm\nu}_\ell^{\mathrm{ID}})=\operatorname{ID}(\mathbf s_\ell,\bm\nu_\ell,\dot{\bm\nu}_\ell;\mathcal C_\ell)$.
Following optimization-based control formulations~\citep{righetti2011unified,righetti2013optimal,wensing2023optimization}, the solve minimizes acceleration mismatch subject to actuator bounds $\bm\tau_{\mathrm{lb}}(\mathbf v_{\mathrm{joint}})\leq\bm\tau\leq\bm\tau_{\mathrm{ub}}(\mathbf v_{\mathrm{joint}})$ and unilateral friction cones at active contacts. It accounts for the unactuated floating base through the equations of motion. Appendices~\ref{ssub:id_socp} and~\ref{apx:sec:batched_inverse_dynamics} detail the formulation, velocity-dependent bounds, and batched solver.

\subsubsection{Acceleration Feasibility Constraint}
The inner \ac{ID} solve returns torques and contact forces satisfying its actuator and force-cone constraints. With acceleration residual $\mathbf e_{\mathrm{acc},\ell}=\dot{\bm\nu}_\ell^{\mathrm{ID}}-\dot{\bm\nu}_\ell$, acceleration feasibility additionally requires
\begin{equation}
\mathbf h_{\mathrm{acc},\ell}(\mathbf M)=\mathbf e_{\mathrm{acc},\ell}=\mathbf0,
\qquad \ell=t+1,\ldots,t+F.
\label{eq:h_acc}
\end{equation}
We measure acceleration mismatch as the mean absolute residual over all generalized acceleration components, including the floating base and joints: $v_{\mathrm{acc},\ell}=\|\mathbf e_{\mathrm{acc},\ell}\|_1/n_\nu$.

\subsection{Auto-Regressive Refinement and Motion Consistency}

We generate a long sequence $\mathbf M_{1:T}^{\star}$, with $T\gg F$, by repeatedly optimizing $F$ future frames conditioned on $P$ history frames, as in CAMDM and A-MDM~\citep{chen2024taming,shi2024amdm}. A star denotes the best evaluated candidate in each window. We commit its first $F_{\mathrm{keep}}\leq F$ frames before replanning; the motion-consistency objective encourages agreement over the $F_{\mathrm{ov}}=F-F_{\mathrm{keep}}$ overlapping future frames. Initialization, history updates, and the consistency objective are detailed in Appendices~\ref{apx:sub:diffusion_sampling} and~\ref{apx:sub:task_objectives}.

\section{Diffusion Noise Optimization with Sep-CMA-ES}
\label{sec:diffusion_noise_optimization}

Each candidate noise vector $\mathbf w^{\mathrm{raw}}$ is standardized using \eqnref{eq:noise_norm} and evaluated through the complete sampler $\Phi_\theta(\Psi(\mathbf w^{\mathrm{raw}}),\mathbf c_t)$.

\paragraph{Constraint Softening and Candidate Evaluation}
The resulting soft-constraint objective is
\begin{equation}
\mathcal L(\mathbf M)=\mathcal J(\mathbf M)
+\frac{1}{F}\sum_{f=1}^{F}\left[
\sum_{j\in\mathcal T_{\mathrm h}}\rho_{\mathrm h,j}v_{j,t+f}^{\,2}
+\sum_{j\in\mathcal T_{\mathrm g}}\rho_{\mathrm g,j}v_{j,t+f}^{\,2}
\right],
\label{eq:fitness-unconstrained}
\end{equation}
Here, $\mathcal T_{\mathrm h}$ and $\mathcal T_{\mathrm g}$ index equality and inequality terms, $v_{j,\ell}\geq0$ is the aggregated per-frame violation, and $\rho_{\mathrm h,j},\rho_{\mathrm g,j}\geq0$ weight the squared penalties. See Appendix~\ref{apx:sub:constraint_softening} for aggregation and fitness details.

\paragraph{Batched Sep-CMA-ES Search}\label{sub:sep_cma_es}
Discrete contact decisions and expensive gradients through the diffusion sampler and constrained \ac{ID} motivate sampling-based optimization. We seek convergence in few iterations despite the high-dimensional noise space. We therefore use \ac{Sep-CMA-ES}~\citep{ros2008simple}, whose diagonal covariance keeps computational complexity and memory costs linear in the number of optimization variables for a fixed population size.

We optimize the raw diffusion noise through $K$ population-search iterations:
\begin{equation}
\mathbf w^{\mathrm{raw},\star}\leftarrow\operatorname{SepCMAES}_{K}\!\left[
\mathbf w^{\mathrm{raw}}\mapsto\mathcal L\!\left(\Phi_\theta(\Psi(\mathbf w^{\mathrm{raw}}),\mathbf c_t)\right)
\right].
\label{eq:sep_cma_es_summary}
\end{equation}
\ac{Sep-CMA-ES}~\citep{ros2008simple} samples and ranks candidates, then adapts the search mean, step size, and diagonal covariance. The star denotes the best evaluated candidate after the iteration budget. Appendix~\ref{apx:sec:optimization_baselines} details initialization and updates.

\section{Experiments}
\label{sec:experiments}

\subsection{Experimental Details}
We evaluate the \ac{SMPL} humanoid~\citep{loper2015smpl} and a custom bipedal robot using \acp{MDM} based on \ac{CAMDM}~\mbox{\citep{chen2024taming}}, trained on retargeted 100STYLE motions~\citep{mason2022local}. Generation is conditioned on past motion, a commanded root trajectory, and a style label, with eight denoising steps and autoregressive execution of optimized segments. We evaluate task tracking and physical feasibility from decoded configurations, forward kinematics, and smoothed motion derivatives; evaluation in the dynamic setting additionally uses constrained \ac{ID}. Joint initial- and posterior-noise optimization uses $78{,}840$ scalar optimization parameters per motion horizon for \ac{SMPL} and $103{,}680$ for the bipedal robot in the experimental implementation (Appendix~\ref{apx:sub:noise_parameterization}). Full implementation details appear in Appendices~\ref{apx:sec:motion_diffusion_model}--\ref{apx:sec:experimental_protocol}.

\subsection{Baselines}
We compare inference-time refinements of a frozen generator without requiring a pretrained tracking controller. \emph{Unoptimized} generation establishes the starting quality, and \emph{Noise Best-of-N} tests independent sampling against adaptive search. \emph{Motion GD} and \emph{Motion Sep-CMA-ES} directly refine generated trajectories, motivated by constraint-aware retargeting~\citep{yang2026omniretarget}. \emph{Guided Diffusion} represents objective guidance during denoising~\citep{karunratanakul2023guided}, while \emph{Noise GD} represents gradient-based diffusion-noise optimization~\citep{karunratanakul2024optimizing}. We include \emph{Noise ZeNO}~\citep{kim2026zeno} as a prior sampling-based noise-optimization method, adapting its zeroth-order updates to joint initial and posterior noise for kinematic feasibility. We adapt these technique families to our common objectives and constraints. Two component ablations, \emph{Lat. Noise Sep-CMA-ES} and \emph{Noise Sep-CMA-ES (NNS)}, test posterior-noise optimization and noise standardization, respectively. The latent-only variant disables posterior-noise optimization, holding sampled posterior noise fixed across candidates and iterations within each horizon. NNS denotes no noise standardization: the normalization in \eqnref{eq:noise_norm} is omitted (implementations in Appendix~\ref{apx:sub:baseline_implementations}). 

\subsection{Physical Feasibility Optimization Settings}
We consider two feasibility optimization settings: \textbf{kinematic feasibility optimization}, which penalizes contact-pattern deviations, ground slip, and self-collision; and \textbf{dynamic feasibility optimization}, which penalizes acceleration-reconstruction error, ground slip, and self-collision. Both settings also include velocity tracking and autoregressive motion consistency objectives. The kinematic setting is comparatively inexpensive but requires contact-pattern preservation to discourage floating, whereas the dynamic setting evaluates whether accelerations can be supported under actuator bounds and contact-force cones, at the additional cost of dynamics evaluation and inner \ac{ID} solves. Within each setting, methods share objective \& constraint weights (budgets and weights: Appendix~\ref{apx:sub:optimization_hyperparameters}; active terms: \tabref{tab:optimization_terms}).

\subsection{Evaluation Metrics}
We measure ground penetration, slip, self-collision, acceleration-reconstruction error, and the \ac{ID} force residual, alongside task tracking, autoregressive consistency, and runtime per iteration and motion horizon. Smoothness is measured by \ac{LDJ}~\citep{balasubramanian2015smoothness}, which normalizes jerk energy separately for root translation, root rotation, and joints before averaging and taking the negative logarithm; higher is smoother. For SMPL, a pretrained style classifier provides style accuracy and overall/per-style \ac{FID}~\citep{chen2024taming}.

Framewise metrics pool committed frames over time and, where applicable, styles. Tables report means with the 90th percentile for minimization metrics ($\downarrow$) and the 10th for maximization metrics ($\uparrow$) in brackets. 
 Colors range from red (worse) to green (better); boldface includes means within 3\% of the best value. 
 Appendix~\ref{apx:sub:metrics} defines the metrics and aggregation; classifier and timing details follow in Appendix~\ref{apx:sec:experimental_protocol}.

\subsection{Kinematic Feasibility Optimization Results}
\begin{figure}[t]
    \centering
    \includegraphics[width=0.9\linewidth]{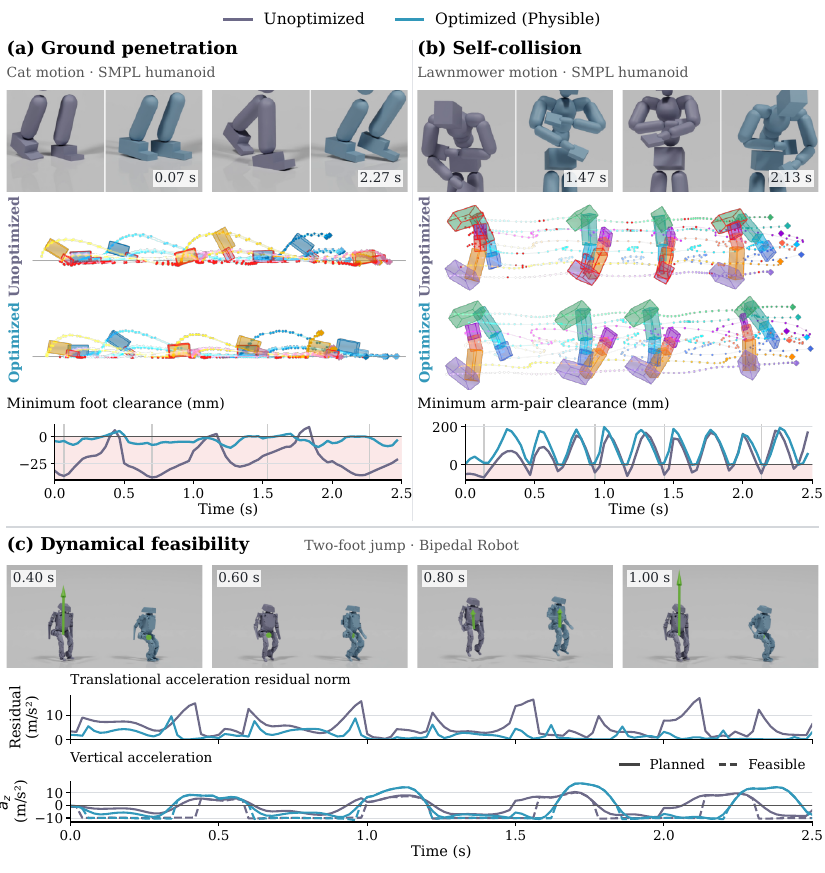}
    \vspace{-0.2cm}
    \caption{
        Qualitative comparison of unoptimized motion (gray-purple) and Physible-optimized motion (cyan-blue).
        (a) Ground-contact behavior for the SMPL humanoid during the \emph{cat} motion, with side-view foot trajectories and minimum foot clearance.
        (b) Arm self-collision behavior during the \emph{lawn-mower} motion, with top-down arm trajectories and minimum arm-pair clearance. In both panels, negative clearance indicates penetration, and red collision geometries mark active collisions or penetrations.
        (c) Dynamic feasibility of the bipedal robot's \emph{two-foot jump}, with rendered snapshots, the translational acceleration residual norm, and planned (solid) versus feasible (dashed) vertical acceleration. Physible reduces the illustrated feasibility violations while preserving the characteristic motion patterns.
    }
    \vspace{-0.4cm}
    \label{fig:combined-feasibility}
\end{figure}

Across 100STYLE and both embodiments, the comparisons in Tabs.~\ref{tab:kinoptim:smpl_humanoid}--\ref{tab:kinoptim:bipedal_robot} show improved kinematic feasibility with Physible; on SMPL, mean ground penetration, self-collision penetration, and slip decrease by 75\%, 92\%, and 74\% relative to unoptimized generation. Figure~\ref{fig:combined-feasibility}(a,b) illustrates the qualitative improvements: Physible reduces foot-ground penetration in the \emph{cat} motion and arm interpenetration in the \emph{lawn-mower} motion while retaining the characteristic poses and motion patterns. Direct motion refinement can introduce jitter (i.e., lower \ac{LDJ}), while Noise GD can degrade motion style. Compared with Noise ZeNO on SMPL, Physible reduces mean ground penetration, self-collision penetration, and slip by 28\%, 67\%, and 39\%, respectively. The component ablations highlight the benefits of standardization for Physible: SMPL style accuracy increases from 76.01\% to 87.73\%, and per-style FID decreases from 74.21 to 43.95. The remaining gap to the unoptimized style accuracy of 96.03\% reflects the tradeoff between physical feasibility and style preservation. Latent-only optimization provides only a slight improvement over unoptimized motion and falls considerably short of Physible (see Appendix~\ref{apx:sec:kinematic_feasibility_optimization_results} for further comparisons and ablations).

\begin{table*}[htbp]
\centering
\caption{Kinematic feasibility on the SMPL humanoid. Metrics show mean [p90 for $\downarrow$; p10 for $\uparrow$], except Style Acc. and Overall FID (means only). NNS: no noise standardization. Green indicates better performance; \textbf{bold} marks means within $3\%$ of the best.}
\label{tab:kinoptim:smpl_humanoid}
\vspace{-0.3cm}
\begingroup
\setlength{\tabcolsep}{1pt}
\renewcommand{\arraystretch}{1}
\fontsize{5}{6}\selectfont
\begin{tabular}{lrrrrrrrrrr}
\toprule
Method & \shortstack[c]{Ground \\ Penetr. [mm] $\downarrow$} & \shortstack[c]{Self-Coll. \\ Penetr. [mm] $\downarrow$} & \shortstack[c]{Ground \\ Slip [m/s] $\downarrow$} & \shortstack[c]{Velocity \\ Tracking $\downarrow$} & \shortstack[c]{Motion \\ Consistency $\downarrow$} & LDJ $\uparrow$ & \shortstack[c]{Style \\ Acc. [\%] $\uparrow$} & \shortstack[c]{Style FID \\ (Overall) $\downarrow$} & \shortstack[c]{Style FID \\ (Per-Style) $\downarrow$} & \shortstack[c]{Comp. Time $\downarrow$ \\ Iter. / Horizon \\ {}[ms] / [s]} \\
\midrule
Unoptimized & \textcolor[RGB]{105,47,28}{2.66e+0 [6.21e+0]} & \textcolor[RGB]{139,26,26}{1.93e-1 [0.00e+0]} & \textcolor[RGB]{139,26,26}{1.85e-1 [4.25e-1]} & \textcolor[RGB]{139,26,26}{7.70e-2 [1.10e-1]} & \textcolor[RGB]{45,83,31}{2.71e-4 [6.92e-4]} & \textcolor[RGB]{27,94,32}{\textbf{-6.42 [-7.15]}} & \textcolor[RGB]{27,94,32}{\textbf{96.03}} & \textcolor[RGB]{27,94,32}{\textbf{7.52}} & \textcolor[RGB]{27,94,32}{\textbf{16.16 [44.54]}} & N/A / \textcolor[RGB]{27,94,32}{\textbf{0.02}} \\
Motion GD & \textcolor[RGB]{139,26,26}{3.59e+0 [9.61e+0]} & \textcolor[RGB]{34,90,32}{1.92e-2 [1.43e-2]} & \textcolor[RGB]{50,80,31}{4.80e-2 [1.45e-1]} & \textcolor[RGB]{47,82,31}{2.01e-2 [5.74e-2]} & \textcolor[RGB]{27,94,32}{\textbf{9.89e-5 [3.33e-4]}} & \textcolor[RGB]{97,52,28}{-8.63 [-9.39]} & \textcolor[RGB]{27,94,32}{\textbf{95.51}} & \textcolor[RGB]{27,94,32}{\textbf{7.64}} & \textcolor[RGB]{27,94,32}{\textbf{15.90 [41.06]}} & \textcolor[RGB]{74,66,30}{57} / \textcolor[RGB]{87,57,29}{5.79} \\
Motion Sep-CMA-ES & \textcolor[RGB]{56,77,30}{1.32e+0 [2.85e+0]} & \textcolor[RGB]{108,45,28}{1.41e-1 [2.36e-1]} & \textcolor[RGB]{95,53,28}{1.17e-1 [2.54e-1]} & \textcolor[RGB]{125,35,27}{6.79e-2 [1.09e-1]} & \textcolor[RGB]{139,26,26}{1.17e-3 [3.77e-3]} & \textcolor[RGB]{139,26,26}{-9.97 [-10.35]} & \textcolor[RGB]{27,94,32}{\textbf{95.54}} & \textcolor[RGB]{27,94,32}{\textbf{7.52}} & \textcolor[RGB]{27,94,32}{16.75 [51.88]} & \textcolor[RGB]{27,94,32}{\textbf{23}} / \textcolor[RGB]{52,79,31}{2.41} \\
Guided Diffusion & \textcolor[RGB]{63,72,30}{1.52e+0 [3.38e+0]} & \textcolor[RGB]{118,39,27}{1.59e-1 [0.00e+0]} & \textcolor[RGB]{112,42,27}{1.44e-1 [3.97e-1]} & \textcolor[RGB]{120,38,27}{6.48e-2 [9.90e-2]} & \textcolor[RGB]{44,84,31}{2.65e-4 [6.85e-4]} & \textcolor[RGB]{46,83,31}{-7.02 [-7.75]} & \textcolor[RGB]{27,94,32}{\textbf{96.33}} & \textcolor[RGB]{27,94,32}{\textbf{7.57}} & \textcolor[RGB]{27,94,32}{\textbf{16.20 [41.30]}} & \textcolor[RGB]{83,60,29}{64} / \textcolor[RGB]{33,90,32}{0.65} \\
Noise Best-of-N & \textcolor[RGB]{81,61,29}{2.03e+0 [5.00e+0]} & \textcolor[RGB]{113,42,27}{1.49e-1 [0.00e+0]} & \textcolor[RGB]{106,46,28}{1.34e-1 [3.21e-1]} & \textcolor[RGB]{131,31,26}{7.19e-2 [1.03e-1]} & \textcolor[RGB]{42,85,31}{2.46e-4 [6.35e-4]} & \textcolor[RGB]{27,94,32}{\textbf{-6.45 [-7.20]}} & \textcolor[RGB]{27,94,32}{\textbf{95.83}} & \textcolor[RGB]{27,94,32}{7.88} & \textcolor[RGB]{27,94,32}{16.99 [42.26]} & \textcolor[RGB]{51,80,31}{40} / \textcolor[RGB]{66,70,30}{3.79} \\
Noise GD & \textcolor[RGB]{110,43,28}{2.80e+0 [2.39e+0]} & \textcolor[RGB]{27,94,32}{\textbf{6.84e-3 [0.00e+0]}} & \textcolor[RGB]{27,94,32}{\textbf{1.27e-2 [4.89e-2]}} & \textcolor[RGB]{27,94,32}{\textbf{7.86e-3 [2.33e-2]}} & \textcolor[RGB]{52,79,31}{3.44e-4 [1.07e-3]} & \textcolor[RGB]{64,72,30}{-7.60 [-8.62]} & \textcolor[RGB]{139,26,26}{15.41} & \textcolor[RGB]{139,26,26}{82.27} & \textcolor[RGB]{139,26,26}{274.81 [396.01]} & \textcolor[RGB]{139,26,26}{105} / \textcolor[RGB]{139,26,26}{10.78} \\
Noise ZeNO & \textcolor[RGB]{41,85,31}{9.39e-1 [2.63e+0]} & \textcolor[RGB]{50,80,31}{4.55e-2 [0.00e+0]} & \textcolor[RGB]{70,68,30}{7.92e-2 [1.96e-1]} & \textcolor[RGB]{108,45,28}{5.81e-2 [8.39e-2]} & \textcolor[RGB]{38,87,31}{2.11e-4 [5.79e-4]} & \textcolor[RGB]{27,94,32}{\textbf{-6.52 [-7.25]}} & \textcolor[RGB]{33,90,32}{91.71} & \textcolor[RGB]{30,92,32}{9.46} & \textcolor[RGB]{31,92,32}{25.76 [81.44]} & \textcolor[RGB]{57,76,30}{45} / \textcolor[RGB]{74,65,29}{4.59} \\
Lat. Noise Sep-CMA-ES & \textcolor[RGB]{103,48,28}{2.62e+0 [6.13e+0]} & \textcolor[RGB]{137,27,26}{1.89e-1 [0.00e+0]} & \textcolor[RGB]{137,27,26}{1.82e-1 [4.22e-1]} & \textcolor[RGB]{138,27,26}{7.62e-2 [1.09e-1]} & \textcolor[RGB]{45,83,31}{2.70e-4 [6.83e-4]} & \textcolor[RGB]{27,94,32}{\textbf{-6.42 [-7.17]}} & \textcolor[RGB]{27,94,32}{\textbf{96.20}} & \textcolor[RGB]{27,94,32}{\textbf{7.53}} & \textcolor[RGB]{27,94,32}{\textbf{16.10 [43.65]}} & \textcolor[RGB]{51,80,31}{40} / \textcolor[RGB]{66,70,30}{3.80} \\
Noise Sep-CMA-ES (NNS) & \textcolor[RGB]{27,94,32}{\textbf{5.41e-1 [1.50e+0]}} & \textcolor[RGB]{28,93,32}{9.22e-3 [0.00e+0]} & \textcolor[RGB]{45,83,31}{4.01e-2 [9.85e-2]} & \textcolor[RGB]{68,69,30}{3.34e-2 [5.01e-2]} & \textcolor[RGB]{37,88,31}{1.92e-4 [5.92e-4]} & \textcolor[RGB]{54,77,31}{-7.29 [-8.11]} & \textcolor[RGB]{55,77,30}{76.01} & \textcolor[RGB]{49,81,31}{22.00} & \textcolor[RGB]{52,79,31}{74.21 [186.84]} & \textcolor[RGB]{51,80,31}{40} / \textcolor[RGB]{66,70,30}{3.80} \\
Physible (ours) & \textcolor[RGB]{32,91,32}{6.76e-1 [1.94e+0]} & \textcolor[RGB]{32,91,32}{1.52e-2 [0.00e+0]} & \textcolor[RGB]{50,80,31}{4.84e-2 [1.21e-1]} & \textcolor[RGB]{78,63,29}{3.92e-2 [5.91e-2]} & \textcolor[RGB]{32,91,32}{1.52e-4 [4.13e-4]} & \textcolor[RGB]{27,94,32}{\textbf{-6.59 [-7.39]}} & \textcolor[RGB]{39,87,31}{87.73} & \textcolor[RGB]{37,88,31}{14.41} & \textcolor[RGB]{39,87,31}{43.95 [124.04]} & \textcolor[RGB]{51,80,31}{40} / \textcolor[RGB]{66,70,30}{3.80} \\
\bottomrule
\end{tabular}
\par\endgroup
\end{table*}

\begin{table*}[htbp]
\centering
\caption{Kinematic feasibility on the bipedal robot. Metrics show mean [p90 for $\downarrow$; p10 for $\uparrow$]. NNS: no noise standardization. Green indicates better performance; \textbf{bold} marks the best results.}
\label{tab:kinoptim:bipedal_robot}
\vspace{-0.3cm}
\begingroup
\setlength{\tabcolsep}{1pt}
\renewcommand{\arraystretch}{1}
\fontsize{5}{6}\selectfont
\begin{tabular}{lrrrrrr}
\toprule
Method & \shortstack[c]{Ground \\ Penetr. [mm] $\downarrow$} & \shortstack[c]{Self-Coll. \\ Penetr. [mm] $\downarrow$} & \shortstack[c]{Ground \\ Slip [m/s] $\downarrow$} & \shortstack[c]{Velocity \\ Tracking $\downarrow$} & \shortstack[c]{Motion \\ Consistency $\downarrow$} & LDJ $\uparrow$ \\
\midrule
Unoptimized & \textcolor[RGB]{72,67,30}{0.56 [1.43]} & \textcolor[RGB]{114,41,27}{1.25e-2 [0.00e+0]} & \textcolor[RGB]{109,44,28}{8.45e-2 [1.68e-1]} & \textcolor[RGB]{88,57,29}{0.07 [0.11]} & \textcolor[RGB]{52,79,31}{2.70e-4 [9.95e-4]} & \textcolor[RGB]{27,94,32}{\textbf{-7.52 [-8.08]}} \\
Motion GD & \textcolor[RGB]{139,26,26}{1.09 [2.91]} & \textcolor[RGB]{57,76,30}{5.85e-3 [0.00e+0]} & \textcolor[RGB]{45,83,31}{2.20e-2 [4.89e-2]} & \textcolor[RGB]{27,94,32}{\textbf{0.03 [0.09]}} & \textcolor[RGB]{27,94,32}{\textbf{6.85e-5 [1.83e-4]}} & \textcolor[RGB]{139,26,26}{-10.53 [-11.19]} \\
Motion Sep-CMA-ES & \textcolor[RGB]{34,90,32}{0.25 [1.28]} & \textcolor[RGB]{92,55,29}{9.99e-3 [0.00e+0]} & \textcolor[RGB]{27,94,32}{\textbf{3.80e-3 [8.74e-3]}} & \textcolor[RGB]{44,84,31}{0.04 [0.08]} & \textcolor[RGB]{48,81,31}{2.34e-4 [8.73e-4]} & \textcolor[RGB]{101,49,28}{-9.52 [-10.31]} \\
Guided Diffusion & \textcolor[RGB]{95,53,28}{0.74 [1.42]} & \textcolor[RGB]{105,47,28}{1.15e-2 [0.00e+0]} & \textcolor[RGB]{133,30,26}{1.09e-1 [2.38e-1]} & \textcolor[RGB]{86,58,29}{0.07 [0.10]} & \textcolor[RGB]{49,81,31}{2.44e-4 [9.27e-4]} & \textcolor[RGB]{58,75,30}{-8.35 [-9.29]} \\
Noise Best-of-N & \textcolor[RGB]{63,72,30}{0.49 [1.30]} & \textcolor[RGB]{89,56,29}{9.68e-3 [0.00e+0]} & \textcolor[RGB]{76,64,29}{5.24e-2 [1.19e-1]} & \textcolor[RGB]{81,61,29}{0.06 [0.10]} & \textcolor[RGB]{46,83,31}{2.22e-4 [8.34e-4]} & \textcolor[RGB]{27,94,32}{\textbf{-7.55 [-8.12]}} \\
Noise GD & \textcolor[RGB]{91,55,29}{0.70 [1.53]} & \textcolor[RGB]{31,92,32}{2.85e-3 [0.00e+0]} & \textcolor[RGB]{139,26,26}{1.15e-1 [2.70e-1]} & \textcolor[RGB]{139,26,26}{0.10 [0.18]} & \textcolor[RGB]{139,26,26}{9.67e-4 [4.71e-3]} & \textcolor[RGB]{45,83,31}{-8.00 [-8.51]} \\
Noise ZeNO & \textcolor[RGB]{67,69,30}{0.52 [1.40]} & \textcolor[RGB]{139,26,26}{1.55e-2 [0.00e+0]} & \textcolor[RGB]{79,62,29}{5.56e-2 [1.27e-1]} & \textcolor[RGB]{92,55,29}{0.07 [0.11]} & \textcolor[RGB]{48,81,31}{2.35e-4 [9.12e-4]} & \textcolor[RGB]{54,77,31}{-8.25 [-8.79]} \\
Lat. Noise Sep-CMA-ES & \textcolor[RGB]{71,67,30}{0.55 [1.42]} & \textcolor[RGB]{107,45,28}{1.18e-2 [0.00e+0]} & \textcolor[RGB]{104,47,28}{8.04e-2 [1.59e-1]} & \textcolor[RGB]{87,57,29}{0.07 [0.10]} & \textcolor[RGB]{51,80,31}{2.59e-4 [9.78e-4]} & \textcolor[RGB]{27,94,32}{\textbf{-7.52 [-8.08]}} \\
Noise Sep-CMA-ES (NNS) & \textcolor[RGB]{27,94,32}{\textbf{0.20 [0.69]}} & \textcolor[RGB]{77,64,29}{8.18e-3 [0.00e+0]} & \textcolor[RGB]{49,81,31}{2.54e-2 [7.44e-2]} & \textcolor[RGB]{51,79,31}{0.05 [0.07]} & \textcolor[RGB]{91,55,29}{5.78e-4 [2.52e-3]} & \textcolor[RGB]{71,67,30}{-8.70 [-9.20]} \\
Physible (ours) & \textcolor[RGB]{41,85,31}{0.31 [0.85]} & \textcolor[RGB]{27,94,32}{\textbf{2.36e-3 [0.00e+0]}} & \textcolor[RGB]{45,83,31}{2.16e-2 [5.59e-2]} & \textcolor[RGB]{31,92,32}{0.04 [0.06]} & \textcolor[RGB]{27,94,32}{7.40e-5 [2.63e-4]} & \textcolor[RGB]{35,89,32}{-7.75 [-8.49]} \\
\bottomrule
\end{tabular}
\vspace{-0.2cm}
\par\endgroup
\end{table*}

\subsection{Dynamic Feasibility Optimization Results}
Across 11 bipedal robot sequences spanning motion styles, slopes, and friction settings, Physible achieves median reductions of 89.5\% in mean acceleration-reconstruction error and 79.5\% in mean \ac{ID} residual relative to unoptimized motion (Appendix~\ref{apx:sub:dynamic_sequence_results}). Qualitative results in Appendix~\ref{apx:sub:dynamic_case_studies} illustrate how Physible reduces acceleration mismatch while retaining characteristic motion and contact behavior. The two-foot-jump comparison (\tabref{tab:dynoptim:bipedal_robot_two_foot_jump_dynoptim} and \figref{fig:dynoptim-methods}) highlights the tradeoffs among methods. Guided Diffusion produces substantial ground penetration, large dynamic residuals, and degraded motion style. Motion Sep-CMA-ES reduces residuals at the expense of smoothness, while Noise GD achieves the smallest residuals but requires more computation and produces persistent floating instead of the intended jumping motion. Noise ZeNO reduces dynamic residuals with an LDJ of $-8.49$. Physible achieves 72\% lower mean acceleration-reconstruction error and 65\% lower mean \ac{ID} residual than Noise ZeNO at comparable smoothness, while retaining the characteristic two-foot-jump behavior (\figref{fig:combined-feasibility}(c)). Across ten optimization seeds for the two-foot jump, Physible achieves consistent dynamic feasibility and motion smoothness, with moderate variability in the resulting motion and dynamic residuals (Appendix~\ref{apx:sub:dynamic_convergence_variability}).

\begin{table*}[!htbp]
\centering
\caption{Dynamic feasibility for the bipedal robot during a two-foot jump. Metrics show mean [p90 for $\downarrow$; p10 for $\uparrow$]. Green indicates better performance; \textbf{bold} marks the best results.}
\label{tab:dynoptim:bipedal_robot_two_foot_jump_dynoptim}
\vspace{-0.3cm}
\begingroup
\setlength{\tabcolsep}{1pt}
\renewcommand{\arraystretch}{1}
\fontsize{5}{6}\selectfont
\begin{tabular}{lrrrr}
\toprule
Method & \shortstack[c]{Acceleration \\ Reconstruction $\downarrow$} & \shortstack[c]{Inverse Dynamics \\ Residual $\downarrow$} & LDJ $\uparrow$ & \shortstack[c]{Comp. Time $\downarrow$ \\ Iter. / Horizon \\ {}[s] / [s]} \\
\midrule
Unoptimized & \textcolor[RGB]{29,93,32}{0.454 [0.954]} & \textcolor[RGB]{32,91,32}{8.44e+1 [1.87e+2]} & \textcolor[RGB]{27,94,32}{\textbf{-7.71 [-8.11]}} & N/A / \textcolor[RGB]{27,94,32}{$<$\textbf{1}} \\
Motion Sep-CMA-ES & \textcolor[RGB]{27,94,32}{0.032 [0.056]} & \textcolor[RGB]{27,94,32}{7.34e+0 [1.29e+1]} & \textcolor[RGB]{139,26,26}{-11.67 [-11.83]} & \textcolor[RGB]{28,93,32}{1.66} / \textcolor[RGB]{58,75,30}{4145} \\
Guided Diffusion & \textcolor[RGB]{139,26,26}{19.163 [34.988]} & \textcolor[RGB]{139,26,26}{1.86e+3 [3.32e+3]} & \textcolor[RGB]{89,56,29}{-9.91 [-10.37]} & \textcolor[RGB]{139,26,26}{17.65} / \textcolor[RGB]{28,93,32}{141} \\
Noise GD & \textcolor[RGB]{27,94,32}{\textbf{0.013 [0.022]}} & \textcolor[RGB]{27,94,32}{\textbf{1.94e+0 [4.57e+0]}} & \textcolor[RGB]{93,54,28}{-10.03 [-10.69]} & \textcolor[RGB]{58,75,30}{5.93} / \textcolor[RGB]{139,26,26}{14828} \\
Noise ZeNO & \textcolor[RGB]{28,93,32}{0.289 [0.545]} & \textcolor[RGB]{31,91,32}{7.51e+1 [1.39e+2]} & \textcolor[RGB]{49,81,31}{-8.49 [-8.76]} & \textcolor[RGB]{27,94,32}{\textbf{1.50}} / \textcolor[RGB]{55,77,30}{3751} \\
Physible (ours) & \textcolor[RGB]{27,94,32}{0.081 [0.174]} & \textcolor[RGB]{28,93,32}{2.65e+1 [6.27e+1]} & \textcolor[RGB]{49,80,31}{-8.51 [-8.83]} & \textcolor[RGB]{29,93,32}{1.81} / \textcolor[RGB]{61,73,30}{4515} \\
\bottomrule
\end{tabular}
\par\endgroup
\end{table*}

\subsection{Training RL Controller on Optimized vs. Unoptimized Motion}
DeepMimic-style policies~\citep{peng2018deepmimic} trained on the optimized bipedal two-foot jump reduce body position, body orientation, and joint position tracking errors by 16.4\%, 11.9\%, and 7.4\%, respectively, relative to training on the unoptimized reference. Root position and orientation errors increase by 39.6\% and 41.3\%, revealing a tradeoff between root-relative body tracking and accumulated global root-pose drift. Appendices~\ref{apx:sub:rl_tracking_protocol} and~\ref{apx:sub:rl_tracking_results} detail the setup and results, respectively.

\section{Conclusion}
Physible refines frozen motion diffusion models through batched, gradient-free optimization of initial and posterior noise, combining kinematic penalties with constrained inverse-dynamics evaluation. Across 100 motion styles on SMPL, it reduces mean ground penetration, self-collision penetration, and slip by 75\%, 92\%, and 74\% relative to unoptimized generation. Across 11 bipedal robot sequences, median reductions in per-sequence mean acceleration-reconstruction error and inverse-dynamics residual are 89.5\% and 79.5\%, respectively, supporting inference-time noise optimization as a means of improving physical feasibility without retraining the generator or requiring tracking-controller rollouts.

\paragraph{Limitations \& Future Work}
Our soft penalties do not guarantee strict physical feasibility, and optimized noise can drift from the learned motion distribution and degrade style. Noise standardization favors style preservation in this tradeoff at the cost of optimization expressivity, a restriction partly alleviated by jointly optimizing initial and posterior noise rather than initial latent noise alone. Evaluation is limited to bipedal embodiments with assumed dynamics and sticking contacts, while repeated feasibility evaluations and inverse-dynamics solves can be costly for real-time use. Appendix~\ref{apx:sec:limitations} expands on limitations and future work.

\ificlrpreprint\else
\vfill
\clearpage
\fi

\setlength{\textfloatsep}{\mainpapertextfloatsep}
\setlength{\floatsep}{\mainpaperfloatsep}
\setlength{\intextsep}{\mainpaperintextsep}

\ificlrpreprint\else
\section*{AI Disclosure}
Generative AI tools were used for limited implementation and editorial assistance. Coding assistants were used to generate draft software code, and AI tools were used to assist with manuscript formatting, readability improvements, consistency checks, identification of potentially relevant literature, and reference formatting. All AI-assisted outputs were independently reviewed, verified, and, where appropriate, modified by the authors. Generative AI tools did not contribute to the conception or development of the research ideas, hypotheses, methods, experimental design, analyses, interpretation of results, conclusions, or other substantive scientific or technical contributions reported in this work. All such contributions were made by the human authors, who retain full responsibility for the content of the manuscript and all associated artifacts.

\section*{Reproducibility Statement}
We support reproducibility through detailed descriptions of our methodology, implementation, and experimental protocol. Appendix~\ref{apx:sec:motion_diffusion_model} documents data preparation, motion retargeting, diffusion-model training, and autoregressive motion generation. Appendix~\ref{apx:sec:objectives_constraints} specifies the task objectives and physical feasibility constraints, while Appendix~\ref{apx:sec:batched_inverse_dynamics} describes the numerical conditioning and efficient batched solution of the inverse-dynamics optimization. Appendix~\ref{apx:sec:optimization_baselines} provides optimization hyperparameters and baseline implementation details. Appendix~\ref{apx:sec:experimental_protocol} specifies the robot models, experimental configurations, evaluation metrics, and computational environment. Appendices~\ref{apx:sec:kinematic_feasibility_optimization_results} and~\ref{apx:sec:dynamic_feasibility_optimization_results} provide supplementary kinematic and dynamic feasibility results, including convergence comparisons, ablations, and detailed motion and feasibility diagnostics covering all robot degrees of freedom for the illustrated sequences. Together, these descriptions and results support independent implementation of the method and examination of the behavior underlying the aggregate metrics and examples presented in the main text.

\fi

\bibliographystyle{iclr2027_conference}
\bibliography{main}

\clearpage

\appendix

\section{Extended Related Work}
\label{apx:sec:related_work}

\subsection{Motion Diffusion Models}
Diffusion models~\citep{ho2020denoising} serve as strong generative priors for humanoid motion, enabling high-quality synthesis under rich conditioning~\citep{tevet2023human, chen2024taming, dabral2023mofusion} and interactive control via guided diffusion~\citep{karunratanakul2023guided, huang2025diffuse, wu2025uniphys, liao2026beyondmimic}. Motion diffusion models commonly generate kinematic trajectories for a downstream tracking controller~\citep{serifi2024robot,yuan2023physdiff,truong2024pdp,watanabe2025simdiff,tevet2025closd}. Diffusion policies instead generate action sequences~\citep{chi2025diffusion}, while joint state--action diffusion models predict both motion and control, allowing objectives on predicted states to guide action generation~\citep{huang2025diffuse,wu2025uniphys}. While these methods generate visually compelling humanoid motions, ensuring strict kinematic validity and contact-rich dynamical feasibility remains a core challenge. Our method optimizes the stochastic inputs of a kinematic motion generator and evaluates physical feasibility before downstream tracking.

\subsection{Inference-Time Optimization of Diffusion Models}\label{sub:diffusion_inference_time_optimization}
A major practical advantage of diffusion models is that sampling can be modified \emph{at inference time} to satisfy additional objectives or constraints without retraining the base model. A common approach is \emph{gradient-based guidance}, which biases the denoising process using gradients of an auxiliary signal. Building on classifier guidance~\citep{dhariwal2021diffusion}, prior work in robotics has explored constraint-aware guidance for kinematic constraints~\citep{ma2025constraint, cho2026safeflow} as well as Lyapunov-based objectives to ensure safety and stability~\citep{cheng2025safe, zhang2025constrained}. Gradient-based guidance requires differentiable objectives, while sampling-based guidance can also incorporate non-differentiable constraint evaluations~\citep{hui2026trainingfree}. Both modify the intermediate sampling dynamics and can bias samples away from the model distribution.

Alternatively, \emph{projection} methods enforce constraints by mapping intermediate denoising states or their clean-sample estimates onto feasible sets~\citep{christopher2024constrained, zampini2025trainingfree}. Related strategies have also been explored in flow-matching frameworks, including exact linear spatial constraints for motion control~\citep{li2026hardflow, utkarsh2025physics, watanabe2026projflow}. While these approaches can be effective when well-defined projection operators exist, their success depends on the informativeness of intermediate sample estimates and the degree to which the projection conflicts with the learned prior.

More recently, \emph{noise optimization} has emerged, which directly optimizes the initial diffusion noise and, for stochastic samplers, noise injected along the sampling trajectory, to improve the generated output under a given objective~\citep{chen2024tinoedit, karunratanakul2024optimizing, eyring2024reno, uehara2025inference, ben2024d, pu2026noiseasier}. This approach evaluates objectives on generated outputs and can regularize the noise to limit deviations from the prior.
Gradient-based approaches update the noise by backpropagating through the generator~\citep{ben2024d, karunratanakul2024optimizing, pu2026noiseasier, zhao2025dartcontrol}. Recent work introduces sampling-based noise updates, including evolutionary search, using only objective evaluations~\citep{kim2026zeno, tan2026dualseed}. For example, ZeNO uses a path-integral update of the latent noise, whereas our \ac{Sep-CMA-ES} strategy adapts a diagonal search covariance and step size using ranked candidates for both latent and posterior noise.

Guidance and projection also require choosing how strongly constraints influence each diffusion step: intermediate trajectories may give incomplete feasibility information, while strong corrections can conflict with the learned prior and compromise motion style. Noise optimization instead evaluates fully denoised outputs and permits regularization of noise statistics~\citep{ben2024d,karunratanakul2024optimizing,su2025noise,eyring2025noise}. Gradient-based noise updates require backpropagation through both the denoising graph and the objective and constraint evaluations. This can be expensive, and contact decisions introduce non-smoothness that can make gradients poorly informative~\citep{antonova2023rethinking,paulus2026hard}.

Efficient gradient-free optimization of high-dimensional (motion) diffusion model noise for non-smooth objectives and constraints remains underexplored. We address this gap by jointly optimizing initial and posterior noise with \ac{Sep-CMA-ES}.

\subsection{Physical Feasibility of Generated Robotic Character Motion}
Physically infeasible references hinder downstream motion tracking and can distort the style and semantics of generated motion~\citep{araujo2026retargeting}. Approaches to improve feasibility generally focus on (i) enhancing the training data and (ii) enforcing constraints at \emph{inference-time} during motion generation.

\paragraph*{Physics-Informed Training}\label{ssub:motion_retargeting}
A common strategy is to retarget motion capture data to new morphologies while enforcing kinematic and contact constraints via constraint-aware optimization~\citep{luo2023perpetual, zhang2023simulation, ze2025twist, yang2026omniretarget, lee2025phuma, pan2025spider, dhedin2026dynaretarget}. SimDiff~\citep{watanabe2025simdiff} constructs simulator-augmented motion data using a pretrained tracking policy and trains adapters around a frozen diffusion backbone, enabling inference without repeated simulator calls. PARC~\citep{xu2025parc} instead iteratively trains a motion generator and tracking controller: the generator proposes motions, the tracker executes them in simulation, and successful executions augment the dataset for subsequent training rounds. Both use physics-based execution to improve the training pipeline. Joint state--action diffusion models, such as Diffuse-CLoC and UniPhys~\citep{huang2025diffuse,wu2025uniphys}, similarly use tracking controllers to collect simulated state trajectories and associated actions for training. They learn to predict motion and control together. InterPhys trains motion generation for consistency with predicted torques and contact parameters supervised by offline dynamics fitting~\citep{xing2026interphys}.
	SafeFlow distills samples guided by kinematic penalties on joint-limit violations, self-collisions, and joint and center-of-mass velocities and accelerations~\citep{cho2026safeflow}. Preference-based post-training methods such as PhysMoDPO~\citep{zhang2026physmodpo} also rely on a physics-based tracking controller. Physics-based corrections can introduce style drift~\citep{zhang2026physmodpo}, and incorporating physics into the training pipeline does not guarantee feasible generation at inference time. For joint state--action models, this limitation also manifests as potential disagreement between predicted states and the motion produced by executing the generated actions, particularly under out-of-distribution conditioning. Our refinement evaluates explicit constraints on generated motion at inference time without training additional generator components or requiring a tracking controller.

\paragraph*{Motion Diffusion Model Optimization}
Finally, physical feasibility of \acp{MDM} can be enforced \emph{at inference time} by modifying diffusion sampling.
\emph{Gradient-based guidance} steers sampling using differentiable penalties for kinematic constraints~\citep{cho2026safeflow}.
PhysDiff~\citep{yuan2023physdiff} injects physics through \emph{intermittent simulator projection}: at selected denoising steps, the partially denoised motion is tracked in simulation and denoising resumes from the resulting closed-loop trajectory.
CLoSD~\citep{tevet2025closd} adopts a related approach in an \emph{autoregressive} setting, where short motion segments are repeatedly planned and executed. The motion actually executed by the simulated character becomes the prefix for the next diffusion plan, closing the loop through the tracking controller; our autoregressive procedure instead conditions on the preceding optimized motion. While both PhysDiff and CLoSD use physics-based tracking to project kinematic into dynamically consistent motions, PhysDiff applies projection within the diffusion trajectory, whereas CLoSD applies it between successive planning steps.

Improving both kinematic and contact-rich dynamical feasibility under non-smooth constraints while preserving motion quality, style, and diversity remains challenging, particularly without retraining the generator or relying on a physics-based tracking controller. Our approach combines gradient-free noise optimization with kinematic penalties and constrained inverse-dynamics evaluations on fully generated motions.

\clearpage

\section{Motion Diffusion Model: Data, Training, and Generation}
\label{apx:sec:motion_diffusion_model}

\subsection{Motion Data and Preprocessing}
We train on the \emph{100STYLE} dataset~\citep{mason2022local}, which contains locomotion sequences performed in 100 distinct styles. The human motion capture is retargeted to the SMPL skeleton and subsequently to the bipedal robot.


This pipeline follows SMooDi~\citep{zhong2024smoodi}, using Rokoko~\citep{rokoko} to map the source motion capture onto the SMPL skeleton and obtain retargeted motions for all 100 styles. SMPL's human-like proportions allow this mapping to be performed kinematically.

The bipedal robot differs substantially from the human body in size, proportions, and kinematic structure. We therefore retarget the SMPL motions using ReActor~\citep{mueller2026reactor}. The simulated retargeting controller tracks scaled SMPL motion. A fixed correspondence maps the pelvis, hips, knees, feet, shoulders, elbows, neck, and head between the two embodiments. The reference motion is scaled by $0.5$; the controller tracks root pose, root velocity, and corresponding body orientations, with auxiliary root forces and torques penalized in the retargeting reward. The resulting robot-motion files provide training data for the bipedal robot. This retargeting stage is distinct from the downstream tracking comparison in Appendix~\ref{apx:sub:rl_tracking_protocol}.

The motion loader resamples the SMPL and bipedal robot clips to \SI{30}{Hz} and \SI{50}{Hz}, respectively. It constructs every contiguous window of $P+F$ frames with stride one and omits clips shorter than this window. The past and future parts are expressed in the heading frame associated with the history. Position features retain metric units; each three-component position occupies a six-entry feature block with three zero-padding entries. Root, joint, and body orientations use the 6D representation described below. Style labels are obtained from the motion-file style identifiers.

\subsection{Motion Representation, Architecture, and Conditioning}
\label{apx:sub:mdm_architecture}
Our \ac{MDM} builds on the \ac{CAMDM} transformer architecture~\citep{chen2024taming} with latent width $256$ (4 layers, 4 heads, feed-forward size 1024, dropout 0.2) and predicts motions over a horizon of $F=20$ frames at \SI{30}{Hz} for the \ac{SMPL} humanoid and $F=30$ at \SI{50}{Hz} for the bipedal robot. It is trained on the \emph{100STYLE} dataset~\citep{mason2022local}. The model is conditioned on (i) the past-motion window $\mathbf M_{\mathrm{pst},t}$, with $P=10$ frames for SMPL and $P=15$ for the bipedal robot (expressed in the current heading frame $\mathcal H_t$), (ii) a desired root pose sequence $\mathbf C_{\mathrm{fut},t}$ in the same frame derived from velocity commands (local $x$--$y$ translational velocity and yaw rate), (iii) a discrete motion-style label $\ell_{\mathrm{syl}}$, and (iv) scalar attention biases $b_{\mathrm{pst}},b_{\mathrm{syl}}$ that modulate the influence of past-motion and style conditioning. We collect these inputs as $\mathbf c_t=(\mathbf M_{\mathrm{pst},t},\mathbf C_{\mathrm{fut},t},\ell_{\mathrm{syl}},b_{\mathrm{pst}},b_{\mathrm{syl}})$. Conditioning is encoded in a tokenized representation: at each denoising step, we embed a timestep token, a global style/condition token, and per-frame tokens for the commanded root trajectory and the past-motion context, concatenate them with the per-frame future-motion tokens to be denoised, add positional encodings, and process the resulting sequence with a Transformer encoder; the scalar conditioning weights enter as learned attention-mask biases on the corresponding token blocks, and \ac{CFG} is implemented by randomly masking the past-motion condition during training (mask probability 0.15).

Orientations use the continuous 6D rotation representation~\citep{zhou2019continuity}. The conversion from generated motion features to physical robot configurations is described in Appendix~\ref{apx:sub:robot_motion_model}.

The two models share the transformer dimensions above. SMPL uses $J=23$ and $n_{\mathrm b}=24$, whereas the bipedal robot uses $J=20$ and $n_{\mathrm b}=21$. Consequently, $n_{\mathrm{enc}}=438$ for SMPL and $n_{\mathrm{enc}}=384$ for the bipedal robot. Although the bipedal robot has revolute joints, the denoiser represents each joint orientation with six features.

\subsection{Training Procedure and Hyperparameters}
\paragraph{Forward Diffusion}
The forward diffusion process gradually perturbs a clean motion sequence $\mathbf{M}^{0}_{\mathrm{fut},t}$ over $D$ diffusion steps into noise $\mathbf{M}^{D}_{\mathrm{fut},t}$ via
\begin{equation}
    \mathbf{M}^{d}_{\mathrm{fut},t} \;=\; \sqrt{\bar\alpha_d}\,\mathbf{M}^{0}_{\mathrm{fut},t} \;+\; \sqrt{1-\bar\alpha_d}\,\bm{\epsilon},
    \quad \bm{\epsilon} \sim \mathcal{N}(\mathbf{0},\mathbf{I}),
    \label{eq:forward-diff}
\end{equation}
with variance schedule $\{\beta_d\}_{d=1}^D$, $\alpha_d=1-\beta_d$, and $\bar\alpha_d:=\prod_{j=1}^d\alpha_j$. Gaussian matrix noise denotes independent standard-normal entries, or equivalently a standard-normal distribution after vectorization.

We train on 100STYLE motions retargeted to the target embodiment. The training objective combines MSE reconstruction of clean motion features with losses on consecutive-frame feature differences, body positions obtained by forward kinematics, and consecutive-frame differences of these body positions. Following the geometric-loss construction of \citet{tevet2023human}, a foot-contact term penalizes predicted foot displacement at frames where the reference foot displacement falls below a contact threshold.

\subsection{Diffusion Sampling and Autoregressive Motion Generation}
\label{apx:sub:diffusion_sampling}
The learned reverse process uses an ancestral \ac{DDPM} sampler with an $x_0$-prediction parameterization.
Concretely, we initialize with latent noise $\mathbf{M}^{D}_{\mathrm{fut},t} \sim \mathcal{N}(\mathbf{0},\mathbf{I})$ and for $d=D,D{-}1,\dots,1$ perform
\begin{align}
    \widehat{\mathbf{M}}^{0}_{\mathrm{fut},t\mid d}
    &=
    \mathcal{M}_\theta\!\left(\mathbf{M}^{d}_{\mathrm{fut},t},\,d,\,\mathbf{c}_t\right),
    \label{eq:x0-pred}\\
    \mathbf{M}^{d-1}_{\mathrm{fut},t}
    &=
    \bm{\mu}^{\mathrm{diff}}_d\!\left(\mathbf{M}^{d}_{\mathrm{fut},t},\widehat{\mathbf{M}}^{0}_{\mathrm{fut},t\mid d}\right)
    +
    \sigma^{\mathrm{diff}}_d\,\mathbf{Z}_d,
    \quad \mathbf{Z}_d \sim \mathcal{N}(\mathbf{0},\mathbf{I}),
    \label{eq:ddpm-reverse}
\end{align}
where $\mathcal{M}_\theta$ is the denoising neural network predicting the clean sequence $\widehat{\mathbf{M}}^{0}_{\mathrm{fut},t\mid d}$.
The reverse-transition mean $\bm{\mu}^{\mathrm{diff}}_d$ is given in closed form by the diffusion posterior,
\begin{equation}
    \bm{\mu}^{\mathrm{diff}}_d
    \;=\;
    a_d\,\widehat{\mathbf{M}}^{0}_{\mathrm{fut},t\mid d}
    \;+\;
    b_d\,\mathbf{M}^{d}_{\mathrm{fut},t},
    \label{eq:posterior-mean}
\end{equation}
with scalar schedule coefficients $\{a_d,b_d\}$, and $(\sigma^{\mathrm{diff}}_d)^2$ set by the (clipped) posterior variance schedule~\citep{ho2020denoising}. In many implementations, the final step returns the mean (i.e., no noise is added for $d=1$).
The conditioning $\mathbf{c}_t$ is intentionally kept general and may include, for example, past motion, task descriptions, the current robot state, sensor measurements, style commands, or any combination thereof.

We use a \ac{DDPM}~\citep{ho2020denoising} with a cosine variance schedule $\beta_d$ and $D=8$ diffusion steps. The forward process perturbs $\mathbf M^0_{\mathrm{fut},t}$ according to \eqnref{eq:forward-diff}. Reverse sampling starts from $\mathbf M^D_{\mathrm{fut},t}$ and proceeds through $d=D,\ldots,1$, using a denoiser trained to predict $\widehat{\mathbf M}^0_{\mathrm{fut},t\mid d}$ and the corresponding posterior with fixed ``small'' variance.

At each autoregressive planning call at frame $t$, the sampler receives $\mathbf M_{\mathrm{pst},t}$ and predicts $F$ future frames. For unoptimized generation, we select the motion whose aggregated loss is closest to the median loss of the generated batch. Optimization methods return the motion specified by their respective procedures in Appendix~\ref{apx:sub:baseline_implementations}: the lowest-loss motion from the final batch or across evaluated candidates, as applicable, except that Motion Sep-CMA-ES returns its final search mean. This selection is performed separately at each autoregressive planning call. We commit the first $F_{\mathrm{keep}}$ frames of the selected motion before replanning, leaving $F_{\mathrm{ov}}=F-F_{\mathrm{keep}}$ overlapping future frames. The next history is $\mathbf M_{t+F_{\mathrm{keep}}-P+1:t+F_{\mathrm{keep}}}$, using the supplied initial history where needed. At $t=0$, the procedure accepts an initial history or the model's neutral history condition. In our experiments, we set $F_{\mathrm{keep}}=15$ for the \ac{SMPL} humanoid and $F_{\mathrm{keep}}=25$ for the bipedal robot, and use a diffusion inference batch size of $64$ for all optimization methods. The experimental history initialization is specified below.

We use full conditional sampling with a classifier-free guidance multiplier of one. The attention biases $b_{\mathrm{pst}}$ and $b_{\mathrm{syl}}$ are specified in Appendix~\ref{apx:sub:optimization_hyperparameters}; constraint guidance is a separate baseline described in Appendix~\ref{apx:sub:baseline_implementations}. At the first horizon, $\mathbf M_{\mathrm{pst},0}$ is an all-zero encoded history condition. Subsequent calls take the last $P$ committed frames as history. The zero feature condition is an input to the denoiser, rather than a prescribed physical joint configuration.

At each planning call, the planar velocity command is integrated over the prediction horizon in the current heading frame. Yaw is accumulated from $\omega_z^{\mathrm{cmd}}$, and the commanded planar displacements are rotated by the evolving yaw before accumulation. The command is held fixed within that horizon. After committing $F_{\mathrm{keep}}$ frames, the heading frame is rebuilt from the last committed root pose and the reference trajectory is constructed again. The remaining $F_{\mathrm{ov}}$ frames of the previous prediction supply the overlap-consistency target. 

\clearpage
\section{Task Objectives and Physical Feasibility Constraints}
\label{apx:sec:objectives_constraints}
This appendix formalizes all relevant optimization objectives and physical feasibility constraints in depth, unless this has already been done in the main manuscript. 
Table~\ref{tab:optimization_terms} reports the active optimization terms for the two physical feasibility optimization settings considered in this paper: (1) kinematic feasibility optimization, and (2) dynamic feasibility optimization.

\begin{table}[ht]
\centering
\caption{Objectives and constraints actively included in the optimization loss for the two feasibility optimization settings. A checkmark indicates a nonzero loss weight; terms without a checkmark may still be evaluated for reporting.}
\label{tab:optimization_terms}
\setlength{\tabcolsep}{6pt}
\begin{tabular}{lcc}
\toprule
Term & Kinematic & Dynamic \\
\midrule
\multicolumn{3}{l}{\textit{Objectives}} \\
Velocity tracking $\mathcal J_{\mathrm{track}}$ & \cmark & \cmark \\
Autoregressive motion consistency $\mathcal J_{\mathrm{cons}}$ & \cmark & \cmark \\
\midrule
\multicolumn{3}{l}{\textit{Equality constraints}} \\
Ground slip $\mathbf h_{\mathrm{slip}}$ & \cmark & \cmark \\
Acceleration reconstruction $\mathbf h_{\mathrm{acc}}$ & \xmark & \cmark \\
\midrule
\multicolumn{3}{l}{\textit{Inequality constraints}} \\
Ground penetration $g_{\mathrm{pen}}$ & \xmark & \xmark \\
Contact-pattern preservation $g_{\mathrm{contact}}$ & \cmark & \xmark \\
Self-collision $g_{\mathrm{self}}$ & \cmark & \cmark \\
Joint-torque limits $g_{\mathrm{torque}}$ & \xmark & \xmark \\
Friction cone $g_{\mathrm{fric}}$ & \xmark & \xmark \\
Unilateral contact $g_{\mathrm{uni}}$ & \xmark & \xmark \\
\midrule
\multicolumn{3}{l}{\textit{Diagnostics (not optimized)}} \\
Smoothness diagnostics (LDJ and jerk) & \xmark & \xmark \\
Reported inverse-dynamics residual $E_{\mathrm{ID}}$ & \xmark & \xmark \\
\bottomrule
\end{tabular}
\end{table}

\subsection{Robot Configuration and Dynamics}
\label{apx:sub:robot_motion_model}
For physical-feasibility evaluation, we decode the generated motion features $\mathbf m_\ell$ into the robot configuration $\mathbf s_\ell=\mathcal P(\mathbf m_\ell)$:
\begin{equation}
 \mathbf s_\ell=(\mathbf x_\ell,\mathbf R_{\mathrm{root},\ell},\mathbf q_\ell),
 \qquad \mathbf R_{\mathrm{root},\ell}\in SO(3),\quad \mathbf q_\ell\in\mathcal Q_{\mathrm{joint}}.
\end{equation}
Here $\mathbf x_\ell$ and $\mathbf R_{\mathrm{root},\ell}$ describe the root pose, and $\mathbf q_\ell$ collects the joint configurations: 23 spherical-joint rotations for SMPL and 20 revolute-joint angles for the bipedal robot. The decoder $\mathcal P$ recovers rotations from the generated 6D features. For the bipedal robot, it extracts each scalar joint angle from the decoded axis-angle vector, assigning its magnitude the sign of its dot product with the configured joint axis.

Generated root poses are expressed in the heading frame $\mathcal H_t$ of the last history frame. Its origin is the ground-projected root position and its yaw is aligned with the root; it remains fixed across the feature window. The decoder transforms the root poses into the world frame $\mathcal W$, in which physical positions are expressed. The rotation $\mathbf R_{\mathrm{root},\ell}$ maps root-frame vectors into $\mathcal W$. Physical-feasibility evaluation uses link and contact geometry computed by forward kinematics from $\mathbf s_\ell$. The predicted per-body position and orientation features enter the motion-consistency objective.

We use generalized velocities and accelerations~\citep{featherstone2008rigid}
\begin{equation}
 \bm\nu_\ell=(\mathbf v_\ell,\bm\omega_\ell,\mathbf v_{\mathrm{joint},\ell})\in\mathbb R^{n_\nu},
 \qquad \dot{\bm\nu}_\ell\in\mathbb R^{n_\nu},\qquad n_\nu=6+n_{\mathrm{joint}}.
\end{equation}
Here $\mathbf v_{\mathrm{joint},\ell}\in\mathbb R^{n_{\mathrm{joint}}}$ collects the joint generalized velocities, and $n_{\mathrm{joint}}$ is the number of joint velocity degrees of freedom: $69$ for SMPL and $20$ for the bipedal robot. Root linear and angular velocities are expressed in world coordinates.
Feasibility evaluation uses the decoded configurations together with the velocity and acceleration estimates described in Appendix~\ref{par:whittaker_eilers}.

\paragraph*{Rigid Body Dynamics}

For the per-frame dynamics equations below, we omit the time subscript $\ell$ for readability. The floating-base equations of motion are~\citep{featherstone2008rigid}
\begin{equation}
 \mathbf B(\mathbf s)\dot{\bm\nu}+\mathbf n(\mathbf s,\bm\nu)
 =\mathbf S^\top\bm\tau+\mathbf J_{\mathrm c}(\mathbf s)^\top\bm\lambda.
 \label{eq:fbdyn}
\end{equation}
The mass matrix is $\mathbf B\in\mathbb R^{n_\nu\times n_\nu}$, and $\mathbf n\in\mathbb R^{n_\nu}$ contains gravity, Coriolis, centrifugal, and any other modeled bias forces. Assuming all joint velocity degrees of freedom are actuated, $\mathbf S=[\mathbf0\ \mathbf I_{n_{\mathrm{joint}}}]\in\mathbb R^{n_{\mathrm{joint}}\times n_\nu}$ selects joint torques $\bm\tau\in\mathbb R^{n_{\mathrm{joint}}}$.

At frame $\ell$, $\mathcal C_\ell\subseteq\{1,\ldots,n_{\mathrm c}\}$ is the active contact set, with $n_{\lambda,\ell}=3|\mathcal C_\ell|$. The stacked contact Jacobian has dimension $n_{\lambda,\ell}\times n_\nu$, and $\bm\lambda$ stacks 3D point-contact forces. Force components and Jacobian rows use the same contact-local frames.
For each active contact, impose the unilateral friction cone
\begin{equation}
 \mathcal K_p=\{\bm\lambda_p:\lambda_{\perp,p}\geq0,\quad
 \|\bm\lambda_{\parallel,p}\|_2\leq\mu_{\mathrm c,p}\lambda_{\perp,p}\},
 \qquad \bm\lambda_p\in\mathcal K_p.
 \label{eq:contact-force-stick-constraints}
\end{equation}
Here $\lambda_{\perp,p}\in\mathbb R$ and $\bm\lambda_{\parallel,p}\in\mathbb R^2$ are normal and tangential force components, and $\mu_{\mathrm c,p}$ is the friction coefficient. We model all active contacts as sticking, $\mathcal C_\ell=\mathcal C_\ell^{\mathrm{stick}}$, $\mathcal C_\ell^{\mathrm{slide}}=\emptyset$. Sticking additionally requires contact kinematics, e.g., $\mathbf J_{\mathrm c,p}\bm\nu=\mathbf0$; its acceleration-level condition is $\mathbf J_{\mathrm c,p}\dot{\bm\nu}+\dot{\mathbf J}_{\mathrm c,p}\bm\nu=\mathbf0$.

\subsection{Motion Smoothing and Derivative Estimation}
\paragraph{Whittaker--Eilers Smoothing}\label{par:whittaker_eilers}
We apply \ac{WE} filtering~\citep{eilers2003perfect} to the coordinate signals used for derivative estimation. Root positions are filtered directly, and revolute joint angles are unwrapped before filtering. For each root or spherical-joint orientation sequence $\mathbf R_\ell$, we form the cumulative incremental rotation-vector signal
\[
\bm\phi_{t+1}=\mathbf0,\qquad
\bm\phi_\ell=\sum_{k=t+2}^{\ell}\Log\!\left(\mathbf R_k\mathbf R_{k-1}^{\top}\right),
\qquad \ell=t+2,\ldots,t+F,
\]
where $\Log:SO(3)\to\mathbb R^3$ returns the rotation vector. For the root, $\mathbf R_\ell=\mathbf R_{\mathrm{root},\ell}$ and the increments use world coordinates; for spherical joints, $\mathbf R_\ell$ is the parent-relative joint rotation and the increments use parent-frame coordinates. The cumulative sum is a signal for derivative estimation, rather than the logarithm of the total relative rotation. Let $\bm\Xi=\bm\Xi_{t+1:t+F}\in\mathbb R^{F\times n_\Xi}$ stack the position, unwrapped-angle, and cumulative rotation-vector signals over the predicted frames, with $n_\Xi$ coordinate channels. The filter gives
\begin{equation}
\bm\Xi^{\mathrm{sm}}=\mathcal S_{\mathrm{WE}}\bm\Xi,
\qquad
\mathcal S_{\mathrm{WE}}
=(\mathbf I_F+\kappa_{\mathrm{WE}}\mathbf D_2^\top\mathbf D_2)^{-1}.
\label{eq:we_solution}
\end{equation}
The superscript $\mathrm{sm}$ denotes the smoothed signals, and $\mathbf D_2\in\mathbb R^{(F-2)\times F}$ is the second-difference matrix with row stencil $(1,-2,1)$. The parameter $\kappa_{\mathrm{WE}}>0$ controls smoothing strength. The filtered signals provide the basis for velocity and acceleration estimation.

\paragraph{Motion Derivatives Estimation}
We estimate the coordinate-signal derivatives by repeated application of \ac{WE} filtering and finite differences:
\begin{equation}
\dot{\bm\Xi}=\mathbf D_1\bm\Xi^{\mathrm{sm}},
\qquad
\ddot{\bm\Xi}=\mathbf D_1\mathcal S_{\mathrm{WE}}\dot{\bm\Xi}.
\label{eq:motion_derivatives}
\end{equation}
The differentiation matrix $\mathbf D_1\in\mathbb R^{F\times F}$ uses the central stencil $(-1,0,1)/(2\Delta t)$ at interior frames and the stencil $(-1,1)/\Delta t$ on the first and last pairs of frames for the boundary rows. Here, $\Delta t=1/f_s$, with $f_s=30\,\mathrm{Hz}$ for SMPL and $f_s=50\,\mathrm{Hz}$ for the bipedal robot. We assemble the resulting root and joint components into the generalized velocities $\bm\nu_\ell$ and accelerations $\dot{\bm\nu}_\ell$.

Smoothing is used to estimate derivatives; it does not replace the configurations used for geometric evaluation. Ground separation, contact-pattern preservation, and self-collision use forward kinematics of the decoded $\mathbf s_\ell$. The tracking and autoregressive-consistency objectives compare the generated motion features directly. Ground slip uses the estimated generalized velocity $\bm\nu_\ell$, and inverse dynamics uses $\mathbf s_\ell$, $\bm\nu_\ell$, and $\dot{\bm\nu}_\ell$, including the velocity-dependent torque bounds. Jerk diagnostics use a separate estimator with unsmoothed finite differences, since smoothing would attenuate rapid acceleration changes and obscure the jerk present in the generated motion.

Each optimization evaluation estimates derivatives over its $F$ predicted frames, with the boundary stencils stated above. The previous horizon's overlap enters $\mathcal J_{\mathrm{cons}}$ but is not prepended to the derivative-estimation window. The committed per-frame diagnostics are concatenated across horizons.

\subsection{Task Objectives}
\label{apx:sub:task_objectives}
In our experiments, the task objective is $\mathcal J=\rho_{\mathrm{track}}\mathcal J_{\mathrm{track}}+\rho_{\mathrm{cons}}\mathcal J_{\mathrm{cons}}$, combining velocity tracking with the motion-consistency objective in \eqnref{eq:motion_consistency_objective}.

\paragraph{Velocity Tracking Objective}
We track the planar root-velocity command $\mathbf v^{\mathrm{cmd}}=(v_x^{\mathrm{cmd}},v_y^{\mathrm{cmd}},\omega_z^{\mathrm{cmd}})$.
We integrate the commanded planar velocity and yaw rate to obtain a reference root trajectory. Each candidate is evaluated in the heading frame of its first predicted frame, denoted here by $\mathcal H=\mathcal H_{t+1}$. The trajectory-tracking objective combines errors in horizontal root position and 6D root orientation:
\begin{equation}
\mathcal J_{\mathrm{track}}(\mathbf M)
=\frac{1}{F}\sum_{f=1}^{F}\left[
\frac{\|{}^{\mathcal H}\mathbf x_{\mathrm{xy},t+f}-{}^{\mathcal H}\mathbf x^{\mathrm{ref}}_{\mathrm{xy},t+f}\|_2}{\sqrt{2}}
+\frac{\|{}^{\mathcal H}\mathbf r_{t+f}-{}^{\mathcal H}\mathbf r^{\mathrm{ref}}_{t+f}\|_2}{\sqrt{6}}
\right].
\label{eq:velocity_tracking_objective}
\end{equation}
The left superscript $\mathcal H$ indicates the frame in which a feature is expressed. Here, ${}^{\mathcal H}\mathbf x^{\mathrm{ref}}_{\mathrm{xy},\ell}$ and ${}^{\mathcal H}\mathbf r^{\mathrm{ref}}_\ell$ are the command-derived reference position and orientation features, expressed in $\mathcal H$. The subscript $\mathrm{xy}$ selects the two horizontal position components. Reference positions accumulate the commanded planar displacements in the evolving heading frame, and reference orientations encode the integrated yaw. Position features are expressed in meters, and orientation features are dimensionless.

\paragraph{Auto-Regressive Motion Consistency Objective}
We encourage agreement between successive predictions over their overlapping future frames. We express the full encoded motion features in the common world frame. With $\mathbf m^{\mathrm{enc},\mathrm{prev}}_\ell$ denoting the prediction from the preceding horizon, we define
\begin{equation}
\mathcal J_{\mathrm{cons}}(\mathbf M)
=\frac{1}{F n_{\mathrm{enc}}}\sum_{f=1}^{F_{\mathrm{ov}}}
\|\mathbf m^{\mathrm{enc}}_{t+f}-\mathbf m^{\mathrm{enc},\mathrm{prev}}_{t+f}\|_2^2,
\qquad F_{\mathrm{ov}}=F-F_{\mathrm{keep}}.
\label{eq:motion_consistency_objective}
\end{equation}
The objective measures the feature discrepancy over the overlap, averaged over the full prediction horizon. We set $\mathcal J_{\mathrm{cons}}=0$ for the first horizon and whenever $F_{\mathrm{ov}}=0$.

Objective weights are given in Appendix~\ref{apx:sub:optimization_hyperparameters}.

\subsection{Kinematic Feasibility Constraints}
\paragraph{Ground Penetration and Slip}\label{apx:par:ground_violation_aggregation}
We measure ground penetration and slip by averaging their magnitudes over the $n_{\mathrm c}$ candidate contact points. With $[x]_+=\max(x,0)$, the per-frame violations are
\begin{equation}
v_{\mathrm{pen},\ell}
=\frac{1}{n_{\mathrm c}}\sum_{p=1}^{n_{\mathrm c}}[-\delta_{\ell,p}]_+,
\qquad
v_{\mathrm{slip},\ell}
=\frac{1}{n_{\mathrm c}}\sum_{p=1}^{n_{\mathrm c}}
a_{\ell,p}\|\mathbf v_{\parallel,\ell,p}\|_2.
\label{eq:ground_violation_aggregation}
\end{equation}
Here, $\mathbf v_{\parallel,\ell,p}=\mathbf v_{\parallel,p}(\mathbf s_\ell,\bm\nu_\ell)$ and $a_{\ell,p}$ indicates an active contact. For flat ground, $a_{\ell,p}=\mathbf 1[\delta_{\ell,p}\leq0]$, where $\mathbf 1[\cdot]$ denotes the indicator function. Tangential velocity comprises the horizontal world-frame components.

\paragraph{Contact Pattern Preservation}\label{apx:sub:contact_pattern_identification}
We infer a reference contact pattern by a strict majority vote over an initial population of $N_{\mathrm{pop}}$ unoptimized motion samples:
\begin{equation}
a^{\mathrm{ref}}_{\ell,p}
=\mathbf 1\!\left[
\sum_{i=1}^{N_{\mathrm{pop}}}\mathbf 1[\delta^{(i)}_{\ell,p}\leq\delta_{\mathrm{vote}}]
>\frac{N_{\mathrm{pop}}}{2}
\right],
\label{eq:contact_pattern_reference}
\end{equation}
where $\delta^{(i)}_{\ell,p}$ is the signed terrain separation in sample $i$, and $\delta_{\mathrm{vote}}$ is the separation threshold used for pattern identification. The reference pattern is held fixed during optimization of each horizon and inferred again at the next autoregressive step. The thresholds $\delta_{\mathrm{vote}}$, $\delta_{\mathrm{pen}}$, and $\delta_{\mathrm{clear}}$ govern reference-pattern identification and preservation; they do not change the physical contact-activation rule $a_{\ell,p}=\mathbf 1[\delta_{\ell,p}\leq0]$ defined in \secref{ssub:contact_slip}.
The indicator $\mathbf 1[\cdot]$ equals one when its argument is true and zero otherwise.
The pointwise penalty is defined in \eqnref{eq:contact_pattern_preservation}. The per-frame violation is $v_{\mathrm{contact},\ell}=\frac{1}{n_{\mathrm c}}\sum_{p=1}^{n_{\mathrm c}}g_{\mathrm{contact},\ell,p}$, averaging these nonnegative components over all contact points.

An example of the contact pattern identification based on the initial motion and subsequent enforcement during optimization, including the thresholds, can be found in Figs.~\ref{fig:kinoptim_results:behavior_comparison:smpl_neutral:contact_pattern_preservation:part1}-\ref{fig:kinoptim_results:behavior_comparison:smpl_neutral:contact_pattern_preservation:part2} of the appendix.

\paragraph{Self-Collision}
For self-collision, let $\mathcal B_{\mathrm{col}}$ be the set of checked body pairs and $\delta^{\mathrm{self}}_{\ell,bb'}$ their signed separation. The pointwise condition is $g_{\mathrm{self},\ell,bb'}=-\delta^{\mathrm{self}}_{\ell,bb'}\leq0$ for $(b,b')\in\mathcal B_{\mathrm{col}}$.
Each checked body is approximated by an oriented bounding box. Native boxes retain their dimensions; a sphere is enclosed by a cube of side twice its radius, and a cylinder by a box spanning its diameter and length. Mesh geometry is enclosed by a box aligned with principal axes computed from its convex hull. The box pose includes the geometry's local offset and orientation and the body's forward-kinematic transform. We apply \ac{SAT}~\citep{dyn4jSAT} to each pair in $\mathcal B_{\mathrm{col}}$ and use the minimum overlap of the tested axes as penetration depth when the boxes intersect. The per-frame mean self-collision violation is
\begin{equation}
 v_{\mathrm{self},\ell}
 =\frac{1}{|\mathcal B_{\mathrm{col}}|}
 \sum_{(b,b')\in\mathcal B_{\mathrm{col}}}
 [-\delta^{\mathrm{self}}_{\ell,bb'}]_+.
\end{equation}
The maximum over checked pairs is also available as a diagnostic. These box approximations and the selected pair set define the collision measure; it is not a mesh-intersection volume.

\subsection{Dynamic Feasibility Constraints}
\subsubsection{Nominal Constrained Inverse Dynamics}\label{ssub:id_socp}
Following the procedure established for optimization-based control of legged robots~\citep{righetti2011unified, righetti2013optimal, wensing2023optimization}, we estimate for each generated motion state with associated time derivatives $(\mathbf{s}, \bm{\nu}, \dot{\bm{\nu}})$, joint torques and contact forces approximating the motion-implied dynamics by solving floating-base \ac{ID} as a constrained least-squares problem based on the tuple of generalized coordinates, velocities and accelerations.
Let $\mathbf{y} \coloneqq [\bm{\tau}^\top, \bm{\lambda}^\top]^\top$ stack the joint torques $\bm{\tau}\in\mathbb{R}^{n_{\mathrm{joint}}}$ and the 3D contact forces $\bm{\lambda}\in\mathbb{R}^{3 |\mathcal{C}|}$ (one force per active contact point $p \in \mathcal{C}$).
Using a root-first row ordering, we rewrite the \ac{EOM} in the linear form
\begin{equation}
\begin{split}
    \mathbf{A}_\mathrm{ID} \, \mathbf{y} = \mathbf{b}_\mathrm{ID},
    \quad
    \mathbf{A}_\mathrm{ID} =
    \begin{bmatrix}
        \mathbf{0} & \mathbf{J}_{\mathrm{c},\mathrm{root}}^\top(\mathbf{s}) \\
        \mathbf{I}_{n_{\mathrm{joint}}} & \mathbf{J}_{\mathrm{c},\mathrm{joint}}^\top(\mathbf{s})
    \end{bmatrix},
    \quad
    \mathbf{b}_\mathrm{ID} = \mathbf{B}(\mathbf{s})\,\dot{\bm{\nu}} + \mathbf{n}(\mathbf{s},\bm{\nu}),
    \label{eq:Az_equals_b}
\end{split}
\end{equation}
where $\mathbf{J}_{\mathrm{c},\mathrm{root}} \in \mathbb{R}^{n_{\lambda} \times 6}$ and $\mathbf{J}_{\mathrm{c},\mathrm{joint}} \in \mathbb{R}^{n_{\lambda} \times n_{\mathrm{joint}}}$ denote the contact Jacobian columns for the root and joint coordinates, respectively. Equivalently, $\mathbf A_{\mathrm{ID}}=[\mathbf S^\top\ \mathbf J_{\mathrm c}^\top]$. We solve for $\mathbf y$ by minimizing the squared generalized acceleration mismatch
\begin{equation}
\begin{aligned}
    \mathbf{y}^{\mathrm{ID}} \in \arg\min_{\mathbf{y}}\quad
    &
    \lVert \mathbf{B}^{-1}(\mathbf{s}) \left ( \mathbf{A}_\mathrm{ID} \, \mathbf{y} - \mathbf{b}_\mathrm{ID} \right ) \rVert_2^2 \;
    \\
    \text{s.t.}\quad
    &
    \bm{\tau}_{\mathrm{lb}}(\mathbf v_{\mathrm{joint}})\;\le\;\bm{\tau}\;\le\;\bm{\tau}_{\mathrm{ub}}(\mathbf v_{\mathrm{joint}}),\\
    &
    \bm{\lambda}_{p} \in \mathcal{K}_p,
    \qquad \forall p \in \mathcal{C}.
\end{aligned}
\label{eq:id_socp}
\end{equation}

The inner inverse-dynamics solution $\mathbf y^{\mathrm{ID}}$ supplies joint torques and contact forces subject to the above constraints. The generalized force residual compares the forces supplied by this solution with those required by the generated motion. Applying the inverse mass matrix converts this residual into the corresponding acceleration mismatch:
\begin{align}
 \mathbf e_{\mathrm{dyn}}&=
 \underbrace{\mathbf A_{\mathrm{ID}}\mathbf y^{\mathrm{ID}}}_{\text{ID generalized force}}
 -\underbrace{\mathbf b_{\mathrm{ID}}}_{\text{required generalized force}},\\
 \dot{\bm\nu}^{\mathrm{ID}}&=\mathbf B(\mathbf s)^{-1}(\mathbf A_{\mathrm{ID}}\mathbf y^{\mathrm{ID}}-\mathbf n),\\
 \mathbf e_{\mathrm{acc}}&=\mathbf B(\mathbf s)^{-1}\mathbf e_{\mathrm{dyn}}
 =\underbrace{\dot{\bm\nu}^{\mathrm{ID}}}_{\text{ID acceleration}}
 -\underbrace{\dot{\bm\nu}}_{\text{motion acceleration}}.
\end{align}
Here, $\dot{\bm\nu}^{\mathrm{ID}}$ is the acceleration induced by the returned torques and contact forces at the same state $(\mathbf s,\bm\nu)$, whereas $\dot{\bm\nu}$ is the acceleration of the generated motion. The superscript $\mathrm{ID}$ denotes quantities obtained from the inner inverse-dynamics solve.

\paragraph{Joint Torque Limits} The joint torque limits are either static bounds $[\bm{\tau}_{\min},\bm{\tau}_{\max}]$ or velocity-dependent bounds with joint-specific slopes $\bm\kappa_\tau\in\mathbb R^{n_{\mathrm{joint}}}$, offsets $\mathbf o_\tau\in\mathbb R_{\geq0}^{n_{\mathrm{joint}}}$ and saturation to the interval $[\bm{\tau}_\mathrm{min}, \bm{\tau}_\mathrm{max}]$
\begin{equation}
\begin{split}
    \bm{\tau}_{\mathrm{lb}}(\mathbf v_{\mathrm{joint}})
    =
    \max \bigl(\bm\kappa_\tau\odot\mathbf v_{\mathrm{joint}} - \mathbf o_\tau, -\bm{\tau}_{\max}\bigr),
    \quad
    \bm{\tau}_{\mathrm{ub}}(\mathbf v_{\mathrm{joint}})
    =
    \min \bigl(\bm\kappa_\tau\odot\mathbf v_{\mathrm{joint}} + \mathbf o_\tau, \bm{\tau}_{\max}\bigr),
    \label{eq:vel_dependent_torque_bounds}
\end{split}
\end{equation}
where the displayed saturation assumes symmetric static bounds $\bm\tau_{\min}=-\bm\tau_{\max}$, $\odot$ denotes componentwise multiplication, and $\max$ and $\min$ act componentwise. The slopes and offsets can be determined via actuator system identification~\citep{grandia2024design, miller2025high}.


\subsubsection{Relaxed Inverse Dynamics with a Virtual Wrench}

For analysis, we also consider a relaxed inverse-dynamics problem that allows a virtual force and torque to act directly on the floating base. This removes the restriction imposed by floating-base underactuation while retaining the joint actuation limits and contact-force constraints. The formulation therefore lets us examine the remaining dynamical feasibility conditions when direct root support is available. Its objective is to find the smallest virtual wrench compatible with these constraints and the planned motion, quantifying the additional root support required. We use this formulation only for analysis and diagnostics; physical-feasibility optimization uses the nominal inverse-dynamics problem in \eqnref{eq:id_socp}.

Let $\bm\xi_{\mathrm{virt}}\in\mathbb R^6$ denote the virtual floating-base wrench, and define
\begin{equation}
 \mathbf y^{\mathrm{virt}}=[\bm\tau^\top,\bm\lambda^\top,\bm\xi_{\mathrm{virt}}^\top]^\top,
 \qquad
 \mathbf A_{\mathrm{ID}}^{\mathrm{virt}}=
 \begin{bmatrix}
 \mathbf0&\mathbf J_{\mathrm c,\mathrm{root}}^\top(\mathbf s)&\mathbf I_6\\
 \mathbf I_{n_{\mathrm{joint}}}&\mathbf J_{\mathrm c,\mathrm{joint}}^\top(\mathbf s)&\mathbf0
 \end{bmatrix}.
 \label{eq:relaxed_AID}
\end{equation}
The virtual wrench acts on the root rows of the equations of motion. Solve
\begin{equation}
\begin{aligned}
 \mathbf y^{\mathrm{virt},\mathrm{ID}}\in\arg\min_{\mathbf y^{\mathrm{virt}}}\quad
 &\|\bm\xi_{\mathrm{virt}}\|_{\mathbf W_{\mathrm{virt}}}^2\\
 \text{s.t.}\quad &\mathbf A_{\mathrm{ID}}^{\mathrm{virt}}\mathbf y^{\mathrm{virt}}=\mathbf b_{\mathrm{ID}},\\
 &\bm\tau_{\mathrm{lb}}(\mathbf v_{\mathrm{joint}})\leq\bm\tau\leq\bm\tau_{\mathrm{ub}}(\mathbf v_{\mathrm{joint}}),\\
 &\bm\lambda_p\in\mathcal K_p,\qquad p\in\mathcal C.
\end{aligned}
\label{eq:relaxed_id_socp}
\end{equation}
Here $\mathbf W_{\mathrm{virt}}\succeq0$ is a diagonal weighting matrix and $\|\bm\xi\|_{\mathbf W}^2=\bm\xi^\top\mathbf W\bm\xi$. For a feasible solution, denote its wrench by $\bm\xi_{\mathrm{virt}}^{\mathrm{ID}}$. The zero-wrench constraint and its mean-absolute violation are
\begin{equation}
\begin{aligned}
 \mathbf h_{\mathrm{virt},\ell}(\mathbf M)&=\bm\xi_{\mathrm{virt},\ell}^{\mathrm{ID}}=\mathbf0,\\
 v_{\mathrm{virt},\ell}&=\frac{1}{6}\|\mathbf h_{\mathrm{virt},\ell}(\mathbf M)\|_1,
 \qquad \ell=t+1,\ldots,t+F.
\end{aligned}
 \label{eq:h_virtual_wrench}
\end{equation}
The horizon mean $F^{-1}\sum_{\ell=t+1}^{t+F}v_{\mathrm{virt},\ell}$ is zero exactly when the returned virtual wrench vanishes at every evaluated frame.

\clearpage
\section{Numerical Implementation of Batched Inverse Dynamics}
\label{apx:sec:batched_inverse_dynamics}

\subsection{Solver Formulation and Numerical Conditioning}
The differentiable floating-base dynamics in \eqnref{eq:fbdyn} are implemented in PyTorch using a custom version of \texttt{adam}~\citep{nori2015idyntree, lerario2026adam} that supports spherical joints. Each spherical joint has four quaternion position coordinates and three angular-velocity coordinates; the implementation accounts for these different position and velocity dimensions in the rigid-body dynamics.

In the \ac{ID} formulation, we assume that $\mathcal C_\ell=\mathcal C_\ell^{\mathrm{stick}}$, $\mathcal C_\ell^{\mathrm{slide}}=\emptyset$. Equation~\eqref{eq:id_socp} has a convex quadratic objective and second-order cone constraints (friction cones). We solve it as an inner-loop \ac{SOCP} with Clarabel~\citep{goulart2024clarabel}, using DiffCP~\citep{diffcp2019} and the differentiable CVXPYlayers PyTorch interface~\citep{agrawal2019differentiable}.


Motion generation, forward kinematics, and dynamics tensors use single precision. The CVXPYlayers problem data are converted to double precision for the CPU solve, and the returned solution is converted back to the dynamics tensor precision and device. Acceleration reconstruction uses a linear solve with $\mathbf B$. 

\subsection{Batching and Computational Efficiency}
At each diffusion noise optimization step, we group the \acp{SOCP} instances (i.e., per individual in the population and per motion-plan time step) by the tuple of active contact-point identities. For each group, we compile a dedicated CVXPY problem and then solve it for all samples in the batch in parallel on the CPU. This approach avoids recompilation as stances change, while also sidestepping the numerical issues that can arise from ``Big-M'' reformulations of disjunctive contact constraints.

The CVXPY layer is cached for each active-contact tuple and reused when that tuple recurs. 
All candidate/frame instances with the same active-contact tuple are stacked into one layer call. The CPU backend uses eight forward workers by default, with an environment-variable override available. 

\subsection{Differentiation for Gradient-Based Baselines}
Gradient-based methods differentiate the scalar batch-mean loss through the optimized motion or noise, rotation decoding, forward kinematics, derivative estimation, and, when active, the CVXPYlayers/DiffCP inverse-dynamics solve. Denoiser parameters remain fixed. Hard contact masks and the grouping of active contacts are discrete selections; derivatives follow the selected problem and do not differentiate a change of contact set. Absolute values, clamping, vector norms, and separating-axis minima use the piecewise derivatives provided by PyTorch.

\section{Physical Feasibility Optimization and Baselines}
\label{apx:sec:optimization_baselines}

\subsection{Constraint Softening and Candidate Evaluation}
\label{apx:sub:constraint_softening}
The soft objective in \eqnref{eq:fitness-unconstrained} adds weighted squared feasibility violations to the task objective $\mathcal J$.
The sets $\mathcal T_{\mathrm h}$ and $\mathcal T_{\mathrm g}$ contain the equality- and inequality-constraint term labels, respectively, including $\mathrm{slip},\mathrm{acc}\in\mathcal T_{\mathrm h}$ and $\mathrm{pen},\mathrm{contact},\mathrm{self}\in\mathcal T_{\mathrm g}$. Each $v_{j,\ell}$ is a nonnegative per-frame violation aggregated over the relevant contacts, collision pairs, or generalized coordinates. We square these scalar violations and average over future frames $f=1,\ldots,F$, with absolute frame $\ell=t+f$. The nonnegative weights $\rho_{\mathrm h,j}$ and $\rho_{\mathrm g,j}$ determine each term's contribution.
The fitness of unnormalized noise is
\begin{equation}
	f(\mathbf w^{\mathrm{raw}};\mathbf c_t)=\mathcal L(\Phi_\theta(\Psi(\mathbf w^{\mathrm{raw}}),\mathbf c_t)).
\end{equation}

\subsection{Noise Parameterization and Optimizer Initialization}
\label{apx:sub:noise_parameterization}
The noise vector in \secref{sec:noise_optimization_mdm} contains $(D+1)Fn_{\mathrm{enc}}$ scalar optimization parameters per candidate. With $D=8$, this gives $9\cdot20\cdot438=78{,}840$ for \ac{SMPL} and $9\cdot30\cdot384=103{,}680$ for the bipedal robot. The sampler directly injects $D-1$ posterior-noise tensors, since its final transition returns the mean, but all $D$ allocated tensors enter posterior-noise standardization.

For each candidate, \eqnref{eq:noise_norm} standardizes the initial-noise tensor and the combined posterior-noise tensor separately, using their empirical means and sample standard deviations. The denominator floor $\varepsilon_{\mathrm{norm}}$ is the machine epsilon of the noise dtype.

For each run (\secref{sec:diffusion_noise_optimization}), initialize the optimizer mean with independent standard-normal entries and set the remaining state to
\begin{equation}
\begin{aligned}
\bm\mu_0^{\mathrm{ES}}&\sim\mathcal N(\mathbf0,\mathbf I_{n_w}),
&\sigma_0^{\mathrm{ES}}&=1,\\
\mathbf a_0^{\mathrm{ES}}&=\mathbf1,
&\mathbf C_0^{\mathrm{ES}}&=\mathbf I_{n_w},\\
\mathbf p_{\sigma,0}&=\mathbf0,
&\mathbf p_{C,0}&=\mathbf0.
\end{aligned}
\label{eq:sep_cma_es_initialization}
\end{equation}
Use $N_{\mathrm{pop}}$ candidates, $N_{\mathrm{sel}}=N_{\mathrm{pop}}/2$ parents, and the recombination weights and adaptation hyperparameters in Appendix~\ref{apx:sub:sep_cma_es_updates}. The experimental population size is specified in Appendix~\ref{apx:sub:optimization_hyperparameters}.

The methods draw latent and posterior noise from the model's standard-normal distributions within each run. Method-specific handling of these noise samples is described in Appendix~\ref{apx:sub:baseline_implementations}.

\subsection{Sep-CMA-ES Updates}\label{apx:sub:sep_cma_es_updates}
We minimize $f$ with \ac{Sep-CMA-ES}~\citep{ros2008simple}, which restricts the CMA-ES covariance~\citep{hansen2001cmaes} to a diagonal matrix. At iteration $k=0,\ldots,K-1$, the optimizer maintains mean $\bm\mu_k^{\mathrm{ES}}\in\mathbb R^{n_w}$, global step size $\sigma_k^{\mathrm{ES}}>0$, and covariance $\mathbf C_k^{\mathrm{ES}}=\operatorname{diag}((\mathbf a_k^{\mathrm{ES}})^{\odot2})$, with positive diagonal scales $\mathbf a_k^{\mathrm{ES}}$. It samples $N_{\mathrm{pop}}$ candidates:
\begin{equation}
 \bm\zeta_{k,i}\sim\mathcal N(\mathbf0,\mathbf I),\qquad
 \mathbf w^{\mathrm{raw}}_{k,i}=\bm\mu_k^{\mathrm{ES}}+
 \sigma_k^{\mathrm{ES}}\mathbf a_k^{\mathrm{ES}}\odot\bm\zeta_{k,i},\qquad i=1,\ldots,N_{\mathrm{pop}}.
\end{equation}
Here $\odot$ and $(\cdot)^{\odot2}$ denote entrywise multiplication and squaring. Rank candidates by $f(\mathbf w^{\mathrm{raw}}_{k,i};\mathbf c_t)$ and recombine the best $N_{\mathrm{sel}}$:
\begin{equation}
 \bm\mu_{k+1}^{\mathrm{ES}}=\sum_{i=1}^{N_{\mathrm{sel}}}\eta_i\mathbf w^{\mathrm{raw}}_{k,i:N_{\mathrm{pop}}},
 \qquad \eta_i\propto\log(N_{\mathrm{sel}}+\tfrac12)-\log i,\qquad \sum_i\eta_i=1.
\label{eq:sep_cma_es_mean_update}
\end{equation}
The index $i:N_{\mathrm{pop}}$ denotes rank within the population, and $\mu_{\mathrm{eff}}=(\sum_i\eta_i^2)^{-1}$ is the effective number of parents. Define the normalized mean shift and step-size evolution path by
\begin{align}
 \mathbf u_k^{\mathrm{ES}}&=(\bm\mu_{k+1}^{\mathrm{ES}}-\bm\mu_k^{\mathrm{ES}})/\sigma_k^{\mathrm{ES}},\\
 \mathbf p_{\sigma,k+1}&=(1-c_\sigma)\mathbf p_{\sigma,k}
 +\sqrt{c_\sigma(2-c_\sigma)\mu_{\mathrm{eff}}}\,(\mathbf C_k^{\mathrm{ES}})^{-1/2}\mathbf u_k^{\mathrm{ES}}.
\end{align}
The step size is updated using cumulative step-size adaptation, with $\chi_{n_w}=\sqrt{n_w}(1-1/(4n_w)+1/(21n_w^2))$:
\begin{equation}
 \sigma_{k+1}^{\mathrm{ES}}=\sigma_k^{\mathrm{ES}}\exp\!\left[
 \frac{c_\sigma}{d_\sigma}\left(\frac{\|\mathbf p_{\sigma,k+1}\|_2}
 {\chi_{n_w}}-1\right)\right].
\end{equation}
For the diagonal covariance update, let $\widetilde{\mathbf u}_{k,i}^{\mathrm{ES}}=(\mathbf w^{\mathrm{raw}}_{k,i:N_{\mathrm{pop}}}-\bm\mu_k^{\mathrm{ES}})/\sigma_{k+1}^{\mathrm{ES}}$ and define the path gate
\[
h_{\sigma,k}=\mathbf1\!\left[
\|\mathbf p_{\sigma,k+1}\|_2<
\left(1.4+\frac{2}{n_w+1}\right)\chi_{n_w}
\sqrt{1-(1-c_\sigma)^{2(k+1)}}\right].
\]
With $\delta_k=(1-h_{\sigma,k})c_c(2-c_c)$, the updates are
\begin{align}
 \mathbf p_{C,k+1}&=(1-c_c)\mathbf p_{C,k}+h_{\sigma,k}\sqrt{c_c(2-c_c)\mu_{\mathrm{eff}}}\,\mathbf u_k^{\mathrm{ES}},\\
 (\mathbf a_{k+1}^{\mathrm{ES}})^{\odot2}&=
 (1-c_1-c_\mu+\delta_k)(\mathbf a_k^{\mathrm{ES}})^{\odot2}
 +c_1\mathbf p_{C,k+1}^{\odot2}
 +c_\mu\sum_{i=1}^{N_{\mathrm{sel}}}\eta_i(\widetilde{\mathbf u}_{k,i}^{\mathrm{ES}})^{\odot2}.
\label{eq:sep_cma_es_covariance_update}
\end{align}
The next scale vector is the entrywise square root. All search vectors and evolution paths have dimension $n_w$. Writing $n=n_w$ and $m=\mu_{\mathrm{eff}}$, the adaptation hyperparameters are
\[
\begin{aligned}
c_\sigma&=\frac{m+2}{n+m+5},&
d_\sigma&=1+2\max\!\left(0,\sqrt{\frac{m-1}{n+1}}-1\right)+c_\sigma,\\
c_c&=\frac{4+m/n}{n+4+2m/n},&
c_1&=\min\!\left(1,\frac{2n}{(n+1.3)^2+m}\right),\\
\bar c_\mu&=\min\!\left(1-c_1,\frac{2(m-2+1/m)}{(n+2)^2+m}\right),&
c_\mu&=\min(1-c_1,n\bar c_\mu).
\end{aligned}
\]

\subsection{Optimization Procedure and Autoregressive Integration}
Each population is evaluated with $f(\mathbf w^{\mathrm{raw}};\mathbf c_t)=\mathcal L(\Phi_\theta(\Psi(\mathbf w^{\mathrm{raw}}),\mathbf c_t))$, using the generator in Appendix~\ref{apx:sec:motion_diffusion_model}, the objectives and constraints in Appendix~\ref{apx:sec:objectives_constraints}, and the numerical evaluation in Appendix~\ref{apx:sec:batched_inverse_dynamics}. The updates above are repeated for the fixed iteration budget. The best evaluated candidate supplies the committed motion frames and the next history window according to the autoregressive procedure in the main text.

\begin{algorithm}[t]
\SetKwComment{tcp}{}{}
\SetCommentSty{textnormal}
\caption{Autoregressive motion diffusion noise optimization with Sep-CMA-ES}
\label{alg:autoregressive_noise_optimization}
\KwIn{Frozen sampler $\Phi_\theta$, target length $T$, history length $P$, horizon $F$, commitment $F_{\mathrm{keep}}$, iteration budget $K$}
\tcp{Initialize time and zero encoded history}
$t\leftarrow0$, $\mathbf M_{\mathrm{pst},0}\leftarrow\mathbf0$\;
\While{$t<T$ \textnormal{(autoregressive motion generation)}}{
\tcp{Construct the heading frame $\mathcal H_t$ and conditioning from the committed history and current command}
$\mathbf c_t\leftarrow(\mathbf M_{\mathrm{pst},t},\mathbf C_{\mathrm{fut},t},\ell_{\mathrm{syl}},b_{\mathrm{pst}},b_{\mathrm{syl}})$\;
Initialize the Sep-CMA-ES state using \eqnref{eq:sep_cma_es_initialization}\;
\tcp{Generate an initial population; infer the reference contacts when contact-pattern preservation is used}
Generate initial motions with $\Phi_\theta\circ\Psi$ at fixed $\mathbf c_t$; compute $a^{\mathrm{ref}}_{\ell,p}$ using \eqnref{eq:contact_pattern_reference}\;
Hold the inferred contact pattern and previous prediction's overlap fixed within this horizon\;
\For{$k=0,\ldots,K-1$ \textnormal{(Sep-CMA-ES optimization)}}{
\ForEach{$i=1,\ldots,N_{\mathrm{pop}}$ \textnormal{(population evaluation in parallel)}}{
\tcp{Sample raw noise}
$\bm\zeta_{k,i}\sim\mathcal N(\mathbf0,\mathbf I_{n_w})$,\quad
$\mathbf w^{\mathrm{raw}}_{k,i}\leftarrow\bm\mu_k^{\mathrm{ES}}+\sigma_k^{\mathrm{ES}}\mathbf a_k^{\mathrm{ES}}\odot\bm\zeta_{k,i}$\;
\tcp{Standardize the noise}
$\mathbf w_{k,i}\leftarrow\Psi(\mathbf w^{\mathrm{raw}}_{k,i})$\;
\tcp{Generate motion}
$\mathbf M^{0,(k,i)}_{\mathrm{fut},t}\leftarrow\Phi_\theta(\mathbf w_{k,i},\mathbf c_t)$\;
\tcp{Evaluate the fitness}
$f(\mathbf w^{\mathrm{raw}}_{k,i};\mathbf c_t)\leftarrow\mathcal L(\mathbf M^{0,(k,i)}_{\mathrm{fut},t})$\;
}
Update the Sep-CMA-ES state using Eqs.~\eqref{eq:sep_cma_es_mean_update}--\eqref{eq:sep_cma_es_covariance_update}\;
\tcp{Retain the best candidate's noise and already evaluated motion across iterations}
$(s^\star,i^\star)\leftarrow\displaystyle\operatorname*{arg\,min}_{0\leq s\leq k,\,1\leq i\leq N_{\mathrm{pop}}} f(\mathbf w^{\mathrm{raw}}_{s,i};\mathbf c_t)$\;
$\mathbf w^{\mathrm{raw},\star}\leftarrow\mathbf w^{\mathrm{raw}}_{s^\star,i^\star}$,\quad
$\mathbf M^{0,\star}_{\mathrm{fut},t}\leftarrow\mathbf M^{0,(s^\star,i^\star)}_{\mathrm{fut},t}$\;
}
Commit its first $F_{\mathrm{keep}}$ frames and retain the remaining $F_{\mathrm{ov}}$ frames as the next overlap target\;
\tcp{Advance time and update the history from the last $P$ committed frames}
$t\leftarrow t+F_{\mathrm{keep}}$,\quad $\mathbf M_{\mathrm{pst},t}\leftarrow\mathbf M^\star_{t-P+1:t}$\;
}
\end{algorithm}
Optimization uses the fixed $K$-iteration budget. The initial population establishes the reference contact pattern and initial diagnostics; the retained solution is updated by the subsequent optimization iterations. Since $\Phi_\theta$ is deterministic for fixed noise and conditioning, \algref{alg:autoregressive_noise_optimization} retains the winning motion together with its noise, avoiding an additional sampling pass. 

\subsection{Inference-Time Optimization Baselines and Comparison Budgets}
\label{apx:sub:baseline_implementations}
We compare inference-time methods that improve the physical feasibility of generated motion while keeping the \ac{MDM} fixed. Within each experimental setting, the optimization methods use the same task objectives and feasibility constraints. Approaches that require retraining or finetuning the motion model~\citep{xu2025parc, watanabe2025simdiff}, or a pretrained tracking controller for closed-loop simulator rollouts~\citep{yuan2023physdiff, watanabe2025simdiff, tevet2025closd}, fall outside this comparison and are, therefore, out-of-scope for this paper. Appendix~\ref{apx:sub:optimization_hyperparameters} specifies the experimental settings.

\subsubsection{Baseline Methods}
\paragraph{Unoptimized Generation}
We generate a batch of stochastic motions from the trained \ac{MDM} and select the sample whose aggregated loss is closest to the batch median as a representative sample by loss. For an even batch size, the median is the average of the two middle losses. 

\paragraph{Motion-Space Optimization}
We directly post-process the generated motion sequence, analogous in spirit to constraint-aware, physics-informed retargeting of motion capture to robot embodiments~\citep{luo2023perpetual, zhang2023simulation, ze2025twist, yang2026omniretarget, lee2025phuma, pan2025spider, dhedin2026dynaretarget}. \emph{Motion GD} updates the generated batch of motions to minimize the aggregated loss and returns the lowest-loss motion after its final update. \emph{Motion Sep-CMA-ES} initializes its search mean with the lowest-loss generated motion, optimizes in motion space, and returns and evaluates the final search mean.

\paragraph{Guided Diffusion}
We use gradients of the aggregated loss to guide diffusion sampling~\citep{dhariwal2021diffusion, karunratanakul2023guided, ma2025constraint, cheng2025safe, zhang2025constrained, wu2025uniphys, liao2026beyondmimic}. At each denoising step, we evaluate the loss on the predicted clean sample $\widehat{\mathbf{M}}^0_{\mathrm{fut},t\mid d}$ and shift the posterior mean:
\begin{equation}
    \bm{\mu}^{\mathrm{diff}}_d \;=\;
    a_d \,\widehat{\mathbf{M}}^{0}_{\mathrm{fut},t\mid d} + b_d\,\mathbf{M}^{d}_{\mathrm{fut},t} - \eta_d^{\mathrm{guide}} \, \nabla_{\widehat{\mathbf{M}}^{0}_{\mathrm{fut},t\mid d}} \mathcal L \left (\widehat{\mathbf{M}}^{0}_{\mathrm{fut},t\mid d} \right ).
\end{equation}
Here $\eta_d^{\mathrm{guide}}$ combines the fixed guidance scale $s^{\mathrm{guide}}$, the clipped posterior variance at step $d$, and the batch-mean reduction of the loss used for differentiation. Guidance is applied throughout the $D$ denoising steps, and the final sample with the lowest aggregated loss is returned.

\paragraph{Noise Best-of-N}
We sample successive populations of diffusion noise tensors from the nominal training distribution and rank their generated motions by the aggregated loss. The method retains the best noise sampled during its optimization iterations.

\paragraph{Noise GD}
We optimize the batch of diffusion noise tensors by differentiating through the sampler, following prior work on diffusion noise optimization~\citep{chen2024tinoedit, karunratanakul2024optimizing, eyring2024reno, uehara2025inference, ben2024d}. 
Learning rates and Adam parameters are specified in Appendix~\ref{apx:sub:optimization_hyperparameters}.

\paragraph{Noise ZeNO}\label{apx:par:noise_zeno}
ZeNO~\citep{kim2026zeno} optimizes a frozen generator's initial noise using only rewards evaluated on generated outputs. Its update is derived from path-integral control with an Ornstein--Uhlenbeck reference process whose stationary distribution is standard Gaussian. Centering rewards and linearizing their exponential weights under a small-variance approximation yields a control estimate that favors perturbations producing above-average rewards. Each iteration evaluates perturbed noise candidates, combines this estimate with an independent Gaussian perturbation, and projects the updated state onto a sphere.

We adapt this strategy, originally evaluated for image generation and protein design, to the physical-feasibility constraint formulation and task objectives of Physible, extending the optimization variables to both initial and posterior diffusion noise. In the kinematic setting, the reward is the negative of our shared loss, combining contact-pattern preservation, ground slip, and self-collision penalties with velocity tracking and autoregressive motion consistency under the same weights as the other methods.
Let $\mathbf z_k\in\mathbb R^{n_z}$ collect the active noise coordinates, $\mathcal L_{k,i}$ denote the loss of candidate $i$, and $\overline{\mathcal L}_k$ its population mean. The update is
\begin{equation}
\begin{aligned}
\mathbf z_{k,i} &= \sqrt{1-\beta^{\mathrm Z}}\,\mathbf z_k + \sqrt{\beta^{\mathrm Z}}\,\bm\epsilon_{k,i}
&& \text{(candidate sampling)},\\
\mathbf u_k &= -\frac{1}{N_{\mathrm{pop}}}\sum_{i=1}^{N_{\mathrm{pop}}}(\mathcal L_{k,i}-\overline{\mathcal L}_k)\,\bm\epsilon_{k,i}
&& \text{(control estimate)},\\
\widetilde{\mathbf z}_{k+1} &= \sqrt{1-\beta^{\mathrm Z}}\,\mathbf z_k + \sqrt{\beta^{\mathrm Z}}\,\bm\xi_k + \eta^{\mathrm Z}\mathbf u_k
&& \text{(state update)},\\
\mathbf z_{k+1} &= \sqrt{n_z}\,\frac{\widetilde{\mathbf z}_{k+1}}{\|\widetilde{\mathbf z}_{k+1}\|_2}
&& \text{(spherical projection)}.
\end{aligned}
\label{eq:zeno_update}
\end{equation}
where $\bm\epsilon_{k,i}$ and $\bm\xi_k$ are independent standard Gaussian vectors. We initialize $\mathbf z_0\sim\mathcal N(\mathbf0,\mathbf I_{n_z})$, evaluate candidates without projection, and retain the lowest-loss evaluated motion across initialization, candidate populations, and the terminal state. The Langevin interpretation of this practical update is approximate.

The active coordinates comprise the initial noise and all seven consumed posterior-noise tensors, omitting the unused final posterior slot; thus, $n_z=8\cdot20\cdot438=70{,}080$ for SMPL and $n_z=8\cdot30\cdot384=92{,}160$ for the bipedal robot, compared with $78{,}840$ and $103{,}680$ allocated coordinates, respectively, in our standardized formulation (Appendix~\ref{apx:sub:noise_parameterization}). We omit \eqnref{eq:noise_norm}: ZeNO's spherical projection acts on the combined active vector, whereas Physible standardizes initial and posterior noise separately. Hyperparameters are specified in Appendix~\ref{apx:sub:optimization_hyperparameters}.

\subsubsection{Ablations of the Proposed Method}
Both ablations use the Sep-CMA-ES updates and autoregressive integration described above. Their experimental implementations retain the best sampled noise and regenerate its motion after optimization.

\paragraph{Latent Noise Only}
In \emph{Lat. Noise Sep-CMA-ES}, we optimize only the latent noise $\mathbf{M}_{\mathrm{fut},t}^D$. One posterior-noise tensor is sampled and held fixed across candidates and optimization iterations within each horizon.

\paragraph{No Noise Standardization (NNS)}
In \emph{Noise Sep-CMA-ES (NNS)}, NNS denotes \emph{no noise standardization}: we omit the noise normalization in \eqnref{eq:noise_norm} and pass $\mathbf w=\mathbf w^{\mathrm{raw}}$ directly to the sampler, without centering or rescaling either the initial or posterior noise.

\subsubsection{Computational Budgets}
\tabref{tab:appendix_sampling_budgets} counts full batched reverse-sampling passes per motion horizon. Motion-space methods generate their initial batch once and then perform $K$ updates without further denoiser calls. Guided Diffusion uses one pass with guidance at each denoising step. Noise GD uses one pass per update and one final pass to regenerate the motion after the last noise update, giving $K+1$ passes.
We report the same $K+1$ sampling-pass budget for all sampling-based noise methods: Noise Best-of-N, Noise ZeNO, Noise Sep-CMA-ES, and its two ablations.


Runtime measurement is reported in Appendix~\ref{apx:sub:runtime_measurement}.

\begin{table}[t]
\centering
\caption{Number of full \ac{MDM} reverse-sampling passes per horizon, each over the inference batch. Each pass contains $D$ batched denoiser evaluations. Numerical budgets are specified in Appendix~\ref{apx:sub:optimization_hyperparameters}. Motion-space objective evaluations require no additional denoiser calls.}
\label{tab:appendix_sampling_budgets}
\small
\begin{tabularx}{\linewidth}{lcX}
\toprule
Method & Batched sampling passes & Additional evaluation structure\\
\midrule
Unoptimized & $1$ & Score generated batch\\
Motion GD & $1$ & $K$ motion updates and final scoring\\
Motion Sep-CMA-ES & $1$ & $K$ populations and final mean scoring\\
Guided Diffusion & $1$ & Score/guide at each step; score final motion\\
Noise GD & $K+1$ & Regenerate after final noise update\\
Sampling-based noise methods & $K+1$ & Initial population, $K$ populations\\
\bottomrule
\end{tabularx}
\end{table}

\section{Experimental Setup and Evaluation Protocol}
\label{apx:sec:experimental_protocol}

\subsection{Robot Models and Contact Specifications}
\paragraph{Robotic Characters}
We consider two robotic characters in this work: the \ac{SMPL} humanoid~\citep{loper2015smpl} and a custom bipedal robot.
The \ac{SMPL} humanoid has $75$ \acp{DOF} in total, comprising six root \acp{DOF} and $23$ spherical joints. We assume these spherical joints are actuated by simulated Unitree actuators with symmetric torque limits $\tau_{\min}=-\tau_{\max}$, where $\tau_{\max}=\SI{500}{Nm}$, and with a velocity-dependent torque bound parameterized by $\kappa_\tau=\SI{-6.25}{Nms}$ and offset $o_\tau=\SI{625}{Nm}$.
The bipedal robot has $20$ revolute joints actuated by $10$ Unitree A1 motors (symmetric torque limits $\tau_{\max}=\SI{34}{Nm}$, $\kappa_\tau=\SI{-2.70}{Nms}$, $o_\tau=\SI{53.97}{Nm}$) and $10$ Dynamixel XH540-V150 motors (symmetric torque limits $\tau_{\max}=\SI{4.8}{Nm}$, $\kappa_\tau=\SI{-0.71}{Nms}$, $o_\tau=\SI{4.94}{Nm}$).

\paragraph{Contact Points}
For both characters, contact-related constraints are evaluated with a differentiable contact-point module using hard contact masks.
For the \ac{SMPL} character, we model contacts at the four corners of the bottom faces of the \emph{ankle} and \emph{toe} rigid bodies on each leg, yielding $16$ possible contact points in total. For the bipedal robot, we consider four candidate contact points per foot: two at the toe edge and two at the heel edge, identified from the foot-geometry mesh.

The SMPL model has 24 bodies and 23 spherical joints; the bipedal robot model has 21 bodies and 20 revolute joints. For the bipedal robot, the ten hip, knee, and ankle joints use Unitree A1 actuators; the six shoulder joints, two elbows, and two neck joints use Dynamixel XH540-V150 actuators. Unless specified otherwise for the slope and friction experiments, the ground is flat, with gravity $(0,0,-9.80665)$ in meters per second squared, and the dynamic solve uses friction coefficient $\mu_{\mathrm c,p}=0.5$. Joint damping is zero.

For SMPL, the set of body pairs checked for self-collision $\mathcal B_{\mathrm{col}}$ contains all 16 left--right combinations of hip, knee, ankle, and toe bodies, and all 16 left--right combinations of shoulder, elbow, wrist, and hand bodies. For the bipedal robot, it contains all 16 left--right combinations of hip-roll, hip-pitch, knee-pitch, and ankle-pitch bodies, and all 16 left--right combinations of the three shoulder bodies and the elbow. Each pair is checked once. No within-limb or limb--torso pairs are added to these sets.

\subsection{Experimental Details for the Optimization Settings}
\paragraph{Kinematic Feasibility Optimization}
We present a quantitative evaluation in the kinematic setting on an optimized set of $1500$ time steps per style from the \emph{100STYLE} dataset, corresponding to a total optimized motion duration of \SI{83.3}{min} for the \ac{SMPL} humanoid and \SI{50}{min} for the bipedal robot. We resample a velocity command every third motion horizon, drawing from $v_x^{\mathrm{cmd}} \in [0.5, 2.0]$ (forward velocity), $v_y^{\mathrm{cmd}} = 0$ (sideway velocity), and $\omega_z^{\mathrm{cmd}} \in [-2.0, 2.0]$ (yaw angular velocity).

\paragraph{Dynamic Feasibility Optimization}
For the quantitative evaluation in the dynamic setting, we compare Physible and unoptimized generation on 11 bipedal robot sequences spanning six flat-ground motion styles, two slopes, and three friction settings (Appendix~\ref{apx:sub:dynamic_sequence_results}). The \emph{two-foot jump} sequence contains $250$ time steps (\SI{5}{s}); the other ten sequences each contain $150$ time steps (\SI{3}{s}), all sampled at \SI{50}{Hz}. The comparison across optimization methods uses the two-foot-jump sequence. Unlike in the kinematic setting, we do not resample the root-reference velocity command, but use the fixed command $(v_x^{\mathrm{cmd}},v_y^{\mathrm{cmd}},\omega_z^{\mathrm{cmd}})=(\SI{1}{\meter\per\second},\SI{0}{\meter\per\second},\SI{0}{\radian\per\second})$ throughout each sequence.


\paragraph{Selected Kinematic Experiments}
\label{apx:par:kinematic_case_settings}
The targeted self-collision experiment for the SMPL \emph{lawn-mower} motion and the neutral-walking contact-preservation ablation use attention-bias weights $b_{\mathrm{syl}}=1.5$ and $b_{\mathrm{pst}}=-1$, with \ac{WE} smoothing strength $\kappa_{\mathrm{WE}}=1$. The self-collision experiment optimizes only the self-collision constraint for $100$ iterations. Penetration depths are evaluated using \ac{SAT} tests on predefined body pairs belonging to different leaves of the kinematic tree, with spheroid bodies approximated by enclosing cuboids. The contact-preservation ablation compares three selected constant-command runs: unoptimized generation, optimization with a ground-penetration loss, and optimization with contact pattern preservation.

\paragraph{Slope Experiments}
\label{apx:par:slope_settings}
The uphill and downhill experiments use neutral walking at slope angles $\theta=\pm25^\circ$, with $150$ frames at \SI{50}{Hz} and the fixed command $(v_x^{\mathrm{cmd}},v_y^{\mathrm{cmd}},\omega_z^{\mathrm{cmd}})=(\SI{1}{\meter\per\second},\SI{0}{\meter\per\second},\SI{0}{\radian\per\second})$. In the slope-aligned coordinate frame, inverse dynamics uses $\mathbf g=(-g\sin\theta,0,-g\cos\theta)^\top$, with $g=\SI{9.80665}{\meter\per\second\squared}$. Both slopes use friction coefficient $\mu_{\mathrm c,p}=1.0$, attention-bias weights $b_{\mathrm{syl}}=2.5$ and $b_{\mathrm{pst}}=-1$, and \ac{WE} smoothing strength $\kappa_{\mathrm{WE}}=1$. Physible uses \ac{Sep-CMA-ES} with $2500$ iterations per horizon, a population of $64$, and the dynamic loss weights in Appendix~\ref{apx:sub:optimization_hyperparameters}.

\subsection{Optimization Hyperparameters and Objective Weights}
\label{apx:sub:optimization_hyperparameters}
The active objective and constraint terms are summarized in \tabref{tab:optimization_terms}, and the shared numerical settings are collected in \tabref{tab:appendix_optimization_settings}.
We use an inference batch of $N_{\mathrm{pop}}=64$ samples, matching the population size of the sampling-based methods, and $D=8$ denoising steps for every method. The \ac{GD}-based methods update their batch of motion or noise candidates independently, and Guided Diffusion guides each sample in the batch.
For \emph{kinematic feasibility optimization}, we optimize each motion horizon for $K=100$ iterations unless stated otherwise. Since \emph{dynamic feasibility optimization} is substantially more challenging and convergence is slower, we run it for $K=2500$ iterations per horizon. Within each scenario, all iterative sampling-based and \ac{GD}-based methods use the same outer iteration budget, and all optimization methods use the same loss/fitness weights. Guided Diffusion applies guidance during its single $D$-step sampling pass. Method-specific learning rates and guidance scales are given below.

\paragraph{Gradient-Based Baselines}
For all \ac{GD} optimization approaches, we employ an Adam~\citep{kingma2014adam} optimizer with $\beta_1^{\mathrm{Adam}} = 0.9$ and $\beta_2^{\mathrm{Adam}} = 0.999$.
For \emph{Noise GD}, we use a learning rate of $10^{-1}$ (SMPL) and $10^{-2}$ (bipedal robot). For \emph{Motion GD}, we use a learning rate of $10^{-3}$.

\paragraph{Guided Diffusion}
We use a fixed guidance scale $s^{\mathrm{guide}}=10$ in the guidance coefficient $\eta_d^{\mathrm{guide}}$ defined in Appendix~\ref{apx:sub:baseline_implementations}.

\paragraph{Noise ZeNO}
We use $\beta^{\mathrm Z}=0.03$ and $\eta^{\mathrm Z}=50$, selected in preliminary $25$-update kinematic tests, with the shared population size, iteration budgets, loss weights, and generation settings for each experiment.

\paragraph{Kinematic Feasibility Optimization}
Unless specified otherwise for an individual experiment, we use attention-bias weights $b_{\mathrm{syl}}=1.0$ and $b_{\mathrm{pst}}=-0.5$ for \ac{SMPL}, and $b_{\mathrm{syl}}=2.5$ and $b_{\mathrm{pst}}=-1.0$ for the bipedal robot. The corresponding default \ac{WE} smoothing strengths are $\kappa_{\mathrm{WE}}=20$ and $\kappa_{\mathrm{WE}}=1$, respectively. We use a weighted-sum loss with no loss-weight curriculum and no dynamic-feasibility objectives or constraints. The nonzero objective weights are $\rho_{\mathrm{track}}=5$ and $\rho_{\mathrm{cons}}=300$ for \ac{SMPL}, and $\rho_{\mathrm{track}}=1$ and $\rho_{\mathrm{cons}}=300$ for the bipedal robot. The nonzero equality-constraint weight is $\rho_{\mathrm h,\mathrm{slip}}=\SI{10}{\second\squared\per\meter\squared}$ for both humanoids, and the nonzero inequality-constraint weights are $\rho_{\mathrm g,\mathrm{contact}}=\SI{5e4}{\per\meter\squared}$ for contact-pattern preservation and $\rho_{\mathrm g,\mathrm{self}}=\SI{3e4}{\per\meter\squared}$ for self-collision. The penetration and max-violation terms are evaluated for reporting but have zero loss weight. Contact-pattern preservation uses majority-vote contact schedule inference with $\delta_{\mathrm{clear}}=0$; for \ac{SMPL} we set $\delta_{\mathrm{vote}}=\SI{0.008}{\meter}$ and $\delta_{\mathrm{pen}}=\SI{0.005}{\meter}$, while for the bipedal robot we use $\delta_{\mathrm{vote}}=\SI{0.004}{\meter}$ and $\delta_{\mathrm{pen}}=\SI{0.002}{\meter}$. 

\paragraph{Dynamic Feasibility Optimization}
Unless specified otherwise for an individual experiment, we use attention-bias weights $b_{\mathrm{syl}}=1.5$ and $b_{\mathrm{pst}}=-0.5$. The loss is a weighted sum without a loss-weight curriculum. The nonzero objective weights are $\rho_{\mathrm{track}}=1$ and $\rho_{\mathrm{cons}}=300$. The nonzero equality-constraint weights are $\rho_{\mathrm h,\mathrm{slip}}=\SI{10}{\second\squared\per\meter\squared}$ and $\rho_{\mathrm h,\mathrm{acc}}=50$, and the nonzero inequality-constraint weight is $\rho_{\mathrm g,\mathrm{self}}=\SI{3e4}{\per\meter\squared}$.
Contact-pattern preservation uses the same settings for the bipedal robot as in the kinematic setting, and dynamics evaluation uses a static friction coefficient $\mu_{\mathrm c,p}=0.5$ at every contact with acceleration-space residuals. For the two-foot-jump benchmark, motion derivatives are estimated using \ac{WE} smoothing with $\kappa_{\mathrm{WE}}=1$.

\begin{table}[t]
\centering
\caption{Default optimization hyperparameters and objective weights for each embodiment and feasibility optimization setting. Experiment-specific settings are given in the corresponding subsections. Zero weights mean that a term does not contribute to the optimized loss.}
\label{tab:appendix_optimization_settings}
\small
\begin{tabular}{lccc}
\toprule
Parameter & SMPL (kinematic) & \shortstack{bipedal robot\\(kinematic)} & \shortstack{bipedal robot\\(dynamic)}\\
\midrule
$P, F$ & $10,20$ & $15,30$ & $15,30$\\
$F_{\mathrm{keep}}, F_{\mathrm{ov}}$ & $15,5$ & $25,5$ & $25,5$\\
$K$ & $100$ & $100$ & $2500$\\
$N_{\mathrm{pop}}$ & $64$ & $64$ & $64$\\
$D$ & $8$ & $8$ & $8$\\
$b_{\mathrm{syl}},b_{\mathrm{pst}}$ & $1,-0.5$ & $2.5,-1$ & $1.5,-0.5$\\
$\rho_{\mathrm{track}},\rho_{\mathrm{cons}}$ & $5,300$ & $1,300$ & $1,300$\\
$\rho_{\mathrm h,\mathrm{slip}}$ & $10$ & $10$ & $10$\\
$\rho_{\mathrm h,\mathrm{acc}}$ & $0$ & $0$ & $50$\\
$\rho_{\mathrm g,\mathrm{contact}}$ & $5\!\times\!10^4$ & $5\!\times\!10^4$ & $0$\\
$\rho_{\mathrm g,\mathrm{self}}$ & $3\!\times\!10^4$ & $3\!\times\!10^4$ & $3\!\times\!10^4$\\
$\delta_{\mathrm{vote}}$ (mm) & $8$ & $4$ & $4$\\
$\delta_{\mathrm{pen}}$ (mm) & $5$ & $2$ & $2$\\
$\delta_{\mathrm{clear}}$ (mm) & $0$ & $0$ & $0$\\
\bottomrule
\end{tabular}
\end{table}

\subsection{Evaluation Metrics}
\label{apx:sub:metrics}

\paragraph{Physical Feasibility Metrics}
We quantify violations of physical feasibility at each motion frame using the metrics $E_{j,\ell}=v_{j,\ell}$ and also report the task-objective values defined above. Ground penetration and ground slip are averaged over all candidate contact points, and self-collision penetration is averaged over all checked body pairs. Acceleration reconstruction is the mean absolute acceleration residual over the floating base and joints. The inverse-dynamics residual is the Euclidean norm of the generalized force imbalance:
\begin{equation}
E_{\mathrm{acc},\ell}
=\frac{1}{n_\nu}\|\mathbf e_{\mathrm{acc},\ell}\|_1,
\qquad
E_{\mathrm{ID},\ell}=\|\mathbf e_{\mathrm{dyn},\ell}\|_2.
\label{eq:feasibility_diagnostics}
\end{equation}

\paragraph{Metric Variability across Seeds}
For each evaluation metric, let $a_i$ denote the score obtained with optimization seed $i$ for a fixed evaluation setting. For a single sequence, $a_i$ is its temporal mean. When evaluating a collection of $S$ sequences, we use $a_i=S^{-1}\sum_{s=1}^{S}e_{i,s}$, where $e_{i,s}$ is the temporal mean for sequence $s$ under seed $i$, so each sequence receives equal weight. The same sequences and commands are evaluated for every seed. For $N\geq2$ seeds, the mean and sample standard deviation are
\begin{equation}
\bar a=\frac{1}{N}\sum_{i=1}^{N}a_i,\qquad
s_a=\sqrt{\frac{1}{N-1}\sum_{i=1}^{N}(a_i-\bar a)^2}.
\label{eq:optimizer_seed_statistics}
\end{equation}
We express seed variability as $\bar a\pm s_a$. The standard deviation measures seed sensitivity of the chosen score; when scores are averaged over multiple sequences, this aggregation can conceal greater variability within individual sequences. The \emph{best weighted loss} is the loss in \eqnref{eq:fitness-unconstrained} of the best candidate selected at each autoregressive step, averaged over prediction horizons within each sequence. The remaining objective and constraint metrics use the saved best-motion values averaged over retained frames. For \ac{LDJ}, we average finite values over retained frames within each sequence.

\paragraph{Motion Repeatability across Seeds}
We compare motions at matching timestamps within each sequence $s$. For each seed and sequence, we remove the initial planar root position and yaw from the root trajectory, preserving absolute height; the same initial yaw rotation aligns the root orientation while preserving tilt and subsequent heading changes. Denote the aligned root positions and unit quaternions by $\widetilde{\mathbf x}_{i,s,\ell}$ and $\widetilde{\mathbf q}^{\mathrm{root}}_{i,s,\ell}$. For body positions, we subtract the root position and remove its yaw at every frame, giving $\widetilde{\mathbf X}_{i,s,\ell,b}$ for each non-root body $b$. Joint quaternions $\mathbf q^{\mathrm{joint}}_{i,s,\ell,j}$ are expressed relative to their parents. The shortest rotation angle between two unit quaternions is $\theta(\mathbf q,\mathbf q')=2\arccos(\min\{1,|\mathbf q^\top\mathbf q'|\})$.
For seeds $i<k$, the distances over $L_s$ frames are
\begin{equation}
\begin{aligned}
d^{\mathrm{root}}_{ik,s}
&=\sqrt{\frac{1}{L_s}\sum_{\ell=1}^{L_s}\|\widetilde{\mathbf x}_{i,s,\ell}-\widetilde{\mathbf x}_{k,s,\ell}\|_2^2},\\
d^{\mathrm{ori}}_{ik,s}
&=\frac{1}{L_s}\sum_{\ell=1}^{L_s}\theta(\widetilde{\mathbf q}^{\mathrm{root}}_{i,s,\ell},\widetilde{\mathbf q}^{\mathrm{root}}_{k,s,\ell}),\\
d^{\mathrm{joint}}_{ik,s}
&=\frac{1}{L_s J}\sum_{\ell=1}^{L_s}\sum_{j=1}^{J}\theta(\mathbf q^{\mathrm{joint}}_{i,s,\ell,j},\mathbf q^{\mathrm{joint}}_{k,s,\ell,j}),\\
d^{\mathrm{body}}_{ik,s}
&=\frac{1}{L_s B}\sum_{\ell=1}^{L_s}\sum_{b=1}^{B}\|\widetilde{\mathbf X}_{i,s,\ell,b}-\widetilde{\mathbf X}_{k,s,\ell,b}\|_2,
\end{aligned}
\label{eq:seed_motion_distances}
\end{equation}
where $J$ is the number of joints and $B=n_{\mathrm b}-1$ excludes the root body. For each sequence, the average pairwise distance is $\bar d_s=\binom{N}{2}^{-1}\sum_{i<k}d_{ik,s}$. For a collection of $S$ sequences, we average these values equally, $\bar d=S^{-1}\sum_{s=1}^{S}\bar d_s$; for a single sequence, $\bar d=\bar d_s$. Root-trajectory \ac{RMSE} and body-position distance are in meters; root-orientation and joint-angle differences are in radians. Smaller values indicate more repeatable motion under the same commands.

\paragraph{Log Dimensionless Jerk}
\label{apx:par:ldj}
We report group-weighted \ac{LDJ}~\citep{balasubramanian2015smoothness} $E_{\mathrm{LDJ},t}$ as a smoothness diagnostic for an evaluated window at autoregressive step $t$. Let $F_{\mathrm{eval}}\geq5$ be its number of frames and $T_{\mathrm{eval}}=(F_{\mathrm{eval}}-1)\Delta t$ its duration, using the sampling interval $\Delta t$ from Appendix~\ref{par:whittaker_eilers}. Following the generalized-velocity decomposition $\bm\nu_\ell=(\mathbf v_\ell,\bm\omega_\ell,\mathbf v_{\mathrm{joint},\ell})$, we denote the corresponding unsmoothed groups by $\bm\nu^{\mathrm{raw}}_{g,\ell}$, with $g\in\mathcal G=\{\mathrm{lin},\mathrm{ang},\mathrm{joint}\}$, and their jerk vectors by $\mathbf j^{\mathrm{raw}}_{g,\ell}$. The group dimensions are $3$, $3$, and $n_{\mathrm{joint}}$; each spherical joint contributes three angular components.

Velocities and jerks use finite differences of the generated trajectories, without the \ac{WE} smoothing used for dynamics. Angular velocities use quaternion differences for root and spherical-joint orientations; angular jerk uses their cumulative incremental rotation vectors. Revolute-joint angles are unwrapped before differentiation. We retain the estimator's endpoint samples and integrate squared jerk with the trapezoidal rule:
\begin{equation}
\begin{aligned}
V_g^2 &= \max_{1\leq f\leq F_{\mathrm{eval}}}\|\bm\nu^{\mathrm{raw}}_{g,t+f}\|_2^2,\\
I_g &= \Delta t\!\left(\tfrac12\|\mathbf j^{\mathrm{raw}}_{g,t+1}\|_2^2
+\sum_{f=2}^{F_{\mathrm{eval}}-1}\|\mathbf j^{\mathrm{raw}}_{g,t+f}\|_2^2
+\tfrac12\|\mathbf j^{\mathrm{raw}}_{g,t+F_{\mathrm{eval}}}\|_2^2\right),
\qquad Q_g=\frac{T_{\mathrm{eval}}^3 I_g}{V_g^2}.
\end{aligned}
\label{eq:dimensionless_jerk_energy}
\end{equation}
Each $Q_g$ is dimensionless. The active set $\mathcal G_{\mathrm{act}}\subseteq\mathcal G$ contains groups with peak speed $V_g>10^{-8}$ in the corresponding velocity units. We give each active group equal weight and apply the negative natural logarithm after averaging its normalized energy:
\begin{equation}
E_{\mathrm{LDJ},t}=-\ln\!\left[\frac{1}{|\mathcal G_{\mathrm{act}}|}\sum_{g\in\mathcal G_{\mathrm{act}}}Q_g\right].
\label{eq:group_weighted_ldj}
\end{equation}
Thus, the joint group has the same weight as each active root group, irrespective of its number of degrees of freedom. Higher $E_{\mathrm{LDJ},t}$ indicates smoother motion. 

\subsection{Style Classifier and FID Evaluation}
Following \citet{chen2024taming}, we evaluate whether generated motions preserve the commanded style using a style classifier trained on $15$-timestep motion chunks (\SI{0.5}{s}) from the 100STYLE dataset retargeted to the SMPL humanoid~\citep{loper2015smpl}. We segment each generated sequence (with a fixed style command) into overlapping $15$-timestep windows, run the classifier, and report: (i) window-level style accuracy and (ii) overall \ac{FID} pooled across styles and per-style \ac{FID} computed from the classifier activations against the corresponding reference data.

The style classifier embeds each frame's complete encoded motion into 128 features and prepends a learned register token. After positional encoding, a four-layer transformer encoder with four heads, feed-forward width 1024, GELU activation, and dropout 0.2 produces the register representation. This representation, before the final GELU and linear classification head, supplies the features for FID. The head predicts one of the 100 styles.

Classifier inputs are expressed relative to the root position and heading of the window's first frame. Its training script uses 15-frame windows at stride ten, cross-entropy loss, and Adam with learning rate $10^{-3}$. Evaluation uses every sliding 15-frame window at stride one. For a 1500-frame sequence of one style this gives 1486 evaluated windows. The generated features are compared with the stored per-style reference means and covariances; the final style accuracy and FID are evaluated over the full committed sequence.

\subsection{Computational Environment and Runtime Measurement}
\label{apx:sub:runtime_measurement}
We benchmark the runtime of each optimization method in two ways: (i) time per iteration (e.g., a single \ac{GD} update, rolling out and scoring a population of noise samples, or one guided diffusion step) and (ii) total wall-clock time to optimize one motion horizon. For the total-time comparison, we use the same number of iterations for all sampling-based methods (e.g., best-of-n, Sep-CMA-ES) and \ac{GD}-based methods. The optimization experiments and runtime measurements are run on a desktop workstation with a 12th Gen 16-core Intel i9-12900K CPU, an NVIDIA RTX 4090 GPU, and \SI{64}{GB} of RAM.

The per-iteration timers bracket the optimizer's sampling/generation, objective and constraint evaluation, and update. They exclude optimizer initialization and the final regeneration or reevaluation after the last update. Horizon timing brackets the complete optimization call and therefore includes that setup and final evaluation, including CPU transfers and cone-layer construction when first required. For unoptimized generation, the horizon timer brackets sampling; the subsequent feasibility scoring is outside that timer. 



\subsection{Downstream RL Tracking Protocol}
\label{apx:sub:rl_tracking_protocol}
Following prior work~\citep{yang2026omniretarget,mueller2026reactor}, we assess whether improved physical feasibility translates into improved downstream control by training motion-tracking policies on the generated robot motions. We use tracking performance as a proxy for motion quality. We train a separate DeepMimic-style policy~\citep{peng2018deepmimic} for each of the unoptimized and Physible-optimized bipedal two-foot-jump motions from \tabref{tab:dynoptim:bipedal_robot_two_foot_jump_dynoptim}. Each trajectory contains 250 frames at \SI{50}{Hz} (\SI{5}{s}) and is converted directly into a robot reference without further motion correction. The reference contains the root state and 20 joint states.

\paragraph{Simulation and policy.}
We simulate the robot in NVIDIA Isaac Sim 4.2 using $4{,}096$ parallel environments on a single NVIDIA RTX 5090 GPU. The policy runs at \SI{50}{Hz}. The action $\mathbf a_t\in\mathbb R^{20}$ specifies joint-position targets for joint-level PD controllers. The leg actuators use $k_p=15.0$ and $k_d=0.6$, while the arm and neck actuators use $k_p=5.0$ and $k_d=0.2$.

The actor and critic are three-layer MLPs with 512 ELU units per layer. The Gaussian actor is initialized with standard deviation 1.0.

Following VMP~\citep{serifi2024vmp}, let $\mathcal H_t$ denote the robot's motion frame. We write $\mathbf p^{\mathcal H}_t$, $\mathbf R^{\mathcal H}_t$, $\mathbf v^{\mathcal H}_t$, and $\boldsymbol\omega^{\mathcal H}_t$ for the root position, orientation, linear velocity, and angular velocity expressed in this frame, respectively, and let $\rho_6(\cdot)$ denote the 6D rotation representation. The actor observation is
\begin{equation}
\mathbf o_t=
\left(
\mathbf p^{\mathcal H}_t,
\rho_6(\mathbf R^{\mathcal H}_t),
\mathbf v^{\mathcal H}_t,
\boldsymbol\omega^{\mathcal H}_t,
\mathbf q_t,
\dot{\mathbf q}_t,
\mathbf a_{t-1},
\mathbf a_{t-2},
\hat{\mathbf g}_t
\right),
\label{eq:downstream_rl_observation}
\end{equation}
where the motion reference is
\begin{equation}
\hat{\mathbf g}_t=
\left(
\hat{\mathbf p}^{\mathcal H}_t,
\rho_6(\hat{\mathbf R}^{\mathcal H}_t),
\hat{\mathbf v}^{\mathcal H}_t,
\hat{\boldsymbol\omega}^{\mathcal H}_t,
\hat{\mathbf q}_t,
\hat{\dot{\mathbf q}}_t
\right).
\label{eq:downstream_rl_reference}
\end{equation}
The critic additionally receives the privileged root pose, noise-free root and joint velocities, and joint positions.

\paragraph{Reward.}
The reward terms and their weights are summarized in \tabref{tab:downstream_rl_rewards}. Here, $\mathbf p^{\mathcal H}_{b,t}$ denotes the root-relative position of rigid body $b$, and hats denote reference quantities.

\begin{table}[t]
\centering
\caption{\textbf{Downstream RL reward.} Regularization terms are nonnegative costs with negative weights.}
\label{tab:downstream_rl_rewards}
\small
\begin{tabularx}{\textwidth}{l X r}
\toprule
\textbf{Name} & \textbf{Term $r_{i,t}$} & \textbf{Weight $w_i$} \\
\midrule
\multicolumn{3}{c}{\textit{Motion tracking}} \\
\midrule
Root position
& $1-\lVert\mathbf p^{\mathcal H}_t-\hat{\mathbf p}^{\mathcal H}_t\rVert_2^2$ & $1.0$ \\
Root orientation
& $1-\lVert\rho_6(\mathbf R^{\mathcal H}_t)-\rho_6(\hat{\mathbf R}^{\mathcal H}_t)\rVert_2^2$ & $1.0$ \\
Root linear velocity
& $1-\lVert\mathbf v^{\mathcal H}_t-\hat{\mathbf v}^{\mathcal H}_t\rVert_2^2$ & $1.0$ \\
Root angular velocity
& $1-\lVert\boldsymbol\omega^{\mathcal H}_t-\hat{\boldsymbol\omega}^{\mathcal H}_t\rVert_2^2$ & $1.0$ \\
Rigid-body positions
& $1-\sum_b\lVert\mathbf p^{\mathcal H}_{b,t}-\hat{\mathbf p}^{\mathcal H}_{b,t}\rVert_2^2$ & $1.0$ \\
Rigid-body orientations
& $1-\sum_b\lVert\rho_6(\mathbf R^{\mathcal H}_{b,t})-\rho_6(\hat{\mathbf R}^{\mathcal H}_{b,t})\rVert_2^2$ & $1.0$ \\
Joint positions
& $1-\lVert\mathbf q_t-\hat{\mathbf q}_t\rVert_2^2$ & $1.0$ \\
Joint velocities
& $1-\lVert\dot{\mathbf q}_t-\hat{\dot{\mathbf q}}_t\rVert_2^2$ & $0$ \\
Survival
& $1$ & $10.0$ \\
\midrule
\multicolumn{3}{c}{\textit{Regularization}} \\
\midrule
Joint torques
& $\lVert\boldsymbol\tau_t\rVert_2^2$ & $-1.0\mathbin{\cdot}10^{-3}$ \\
Joint accelerations
& $\lVert\ddot{\mathbf q}_t\rVert_2^2$ & $-2.5\mathbin{\cdot}10^{-6}$ \\
Action rate
& $\lVert\mathbf a_t-\mathbf a_{t-1}\rVert_2^2$ & $-1.5$ \\
Action acceleration
& $\lVert\mathbf a_t-2\mathbf a_{t-1}+\mathbf a_{t-2}\rVert_2^2$ & $-0.45$ \\
\bottomrule
\end{tabularx}
\end{table}

\paragraph{Training.}
We optimize both policies with PPO using \SI{10}{s} episodes. Each iteration collects 24 policy steps from every environment, and training runs for at most $100{,}000$ iterations. PPO uses five epochs, four minibatches, an adaptive learning rate initialized at $10^{-3}$, clipping parameter $0.2$, discount factor $0.99$, and GAE parameter $0.95$.

During training, the default bipedal motion-tracking task randomizes body masses by a factor in $[0.9,1.1]$ and foot friction in $[0.4,0.8]$. We additionally apply uniform measurement noise of $\pm0.02\,\mathrm{rad}$ to root roll and pitch, $\pm\SI{0.1}{\meter\per\second}$ to root linear velocity, and $\pm\SI{0.1}{\radian\per\second}$ to root angular velocity.

\paragraph{Evaluation.}
For evaluation, we run the mean action of the Gaussian policy in a single environment using the same observation model and flat terrain. External disturbances are disabled, masses are fixed to their nominal values, and foot friction is fixed at $0.5$. We evaluate both policies under identical conditions and report root and rigid-body position and orientation errors, together with joint-position tracking error, over the complete reference trajectory.

\clearpage
\section{Supplementary Kinematic Feasibility Results}
\label{apx:sec:kinematic_feasibility_optimization_results}
This appendix presents quantitative baseline comparisons, optimization convergence, component ablations, and motion case studies. Experimental settings and metric definitions are given in Appendix~\ref{apx:sec:experimental_protocol}. The case studies pair motion visualizations with the corresponding trajectory and contact diagnostics.

\subsection{Quantitative Comparison with Baselines}
\label{apx:sub:kinematic_baseline_results}
\tabref{tab:kinoptim:smpl_humanoid} and \tabref{tab:kinoptim:bipedal_robot} compare the proposed diffusion-noise optimization approach against the baseline methods for the \ac{SMPL} humanoid and the bipedal robot, respectively.

Overall, the proposed method best addresses the multi-objective problem, consistently reducing kinematic constraint violations and objective costs relative to the unoptimized motion while preserving the motion style. In contrast, the baselines---despite using identical loss/fitness weights---tend to improve only a subset of constraints and objectives while degrading others. Gradient-based methods (e.g., \ac{GD}, \emph{Guided Diffusion}) particularly struggle with the \emph{contact pattern preservation} constraint due to its discontinuous gradients. Sampling-based methods handle this constraint more robustly, with Sep-CMA-ES performing best; however, simple \emph{Best-of-$N$} sampling is insufficient in our high-dimensional search space ($78{,}840$ variables for \ac{SMPL} and $103{,}680$ for the bipedal robot).

While direct motion optimization can be effective, it often introduces visible jitter, which is reflected in lower \ac{LDJ} values. \emph{Guided Diffusion} is computationally efficient because it avoids iterative inference through (or backpropagation through) the full \ac{MDM} and instead only differentiates the loss terms, but its effectiveness is limited---likely due to discontinuous gradients and the small number of denoising steps. Somewhat surprisingly, \ac{GD}-based noise optimization---despite its popularity~\citep{karunratanakul2024optimizing, eyring2025noise}---performs poorly in our setting: it is computationally expensive (requiring backpropagation through both the objectives and the entire \ac{MDM} at each iteration), is sensitive to discontinuous gradient information, and often yields severe ground penetration and degraded motion quality and style.

Under the same physical-feasibility constraint formulation and loss weights, Physible achieves lower mean ground penetration (0.676 vs. \SI{0.939}{mm}), self-collision penetration (0.0152 vs. \SI{0.0455}{mm}), and ground slip (0.0484 vs. \SI{0.0792}{\meter\per\second}) than Noise ZeNO on SMPL, although ZeNO retains slightly higher style accuracy (91.71\% vs. 87.73\%; \tabref{tab:kinoptim:smpl_humanoid}). On the bipedal robot, Physible reduces mean ground penetration, self-collision penetration, and slip by 40\%, 85\%, and 61\% relative to Noise ZeNO, respectively (\tabref{tab:kinoptim:bipedal_robot}).

\subsection{Convergence and Variability}
\label{apx:sub:kinematic_convergence_variability}
\subsubsection{Optimization Convergence}
\label{apx:ssub:kinematic_convergence}
\begin{figure}
    \centering
    \subfigure[Fitness]{\includegraphics[width=0.49\textwidth, trim={5 5 5 5}]{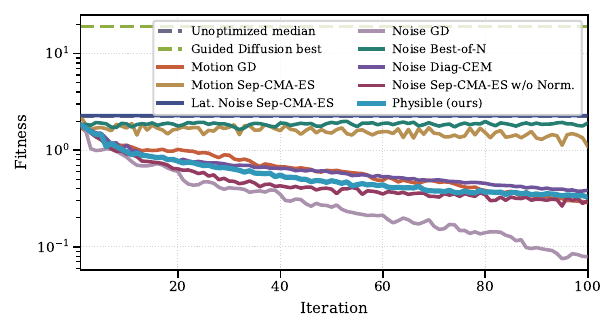}}
    \subfigure[Contact Pattern Preservation]{\includegraphics[width=0.49\textwidth, trim={5 5 5 5}]{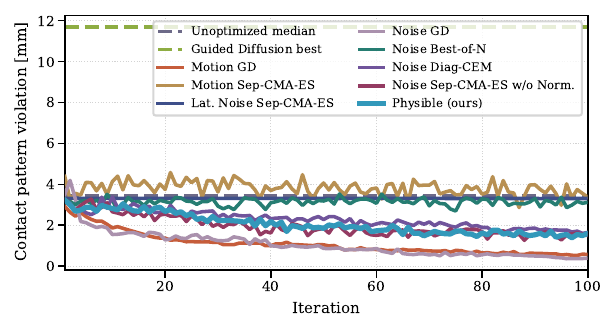}}\\
    \vspace{-0.2cm}
    \subfigure[Ground Penetration]{\includegraphics[width=0.49\textwidth, trim={5 5 5 5}]{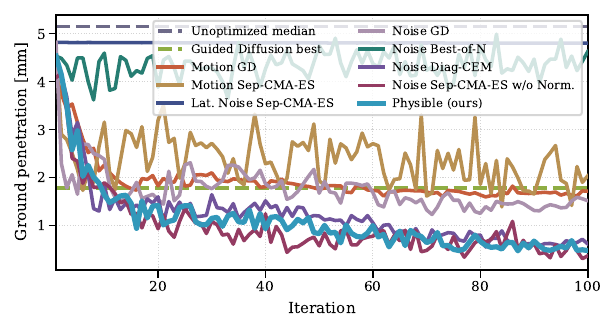}}
    \subfigure[Ground Slip]{\includegraphics[width=0.49\textwidth, trim={5 5 5 5}]{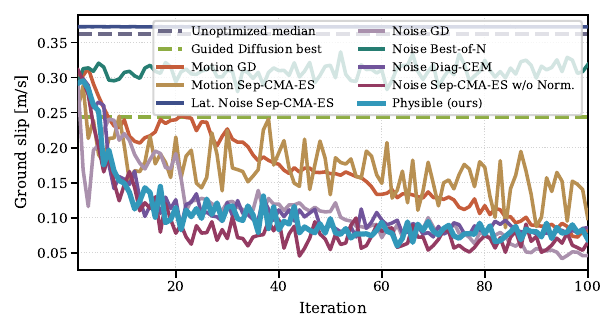}}\\
     \vspace{-0.2cm}
    \subfigure[Self-Collision Penetration]{\includegraphics[width=0.49\textwidth, trim={5 5 5 5}]{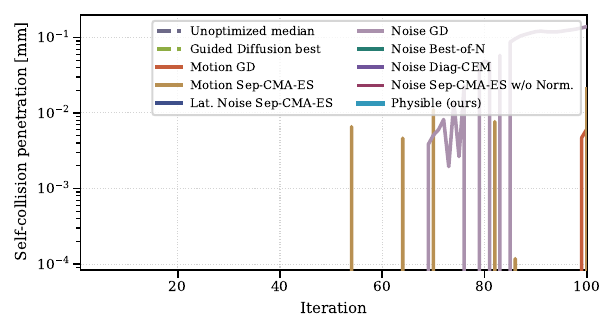}}
    \subfigure[Joint Jerk]{\includegraphics[width=0.49\textwidth, trim={5 5 5 5}]{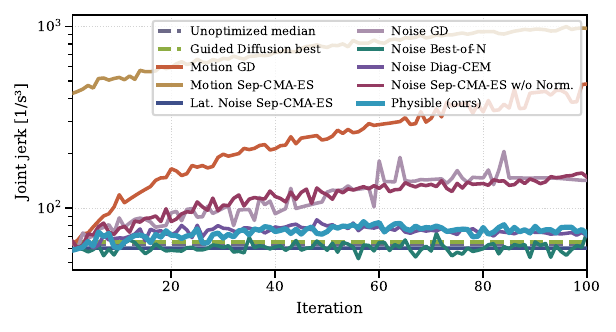}}\\
     \vspace{-0.2cm}
    \subfigure[Linear Root Jerk]{\includegraphics[width=0.49\textwidth, trim={5 5 5 5}]{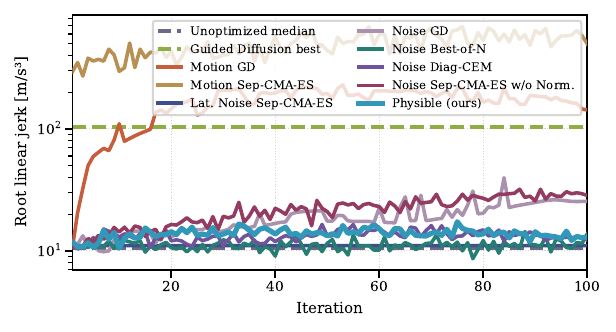}}
    \subfigure[Angular Root Jerk]{\includegraphics[width=0.49\textwidth, trim={5 5 5 5}]{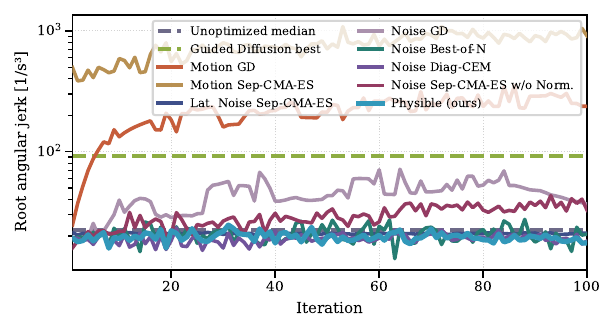}}\\
     \vspace{-0.2cm}
    \subfigure[Velocity Tracking]{\includegraphics[width=0.49\textwidth, trim={5 5 5 5}]{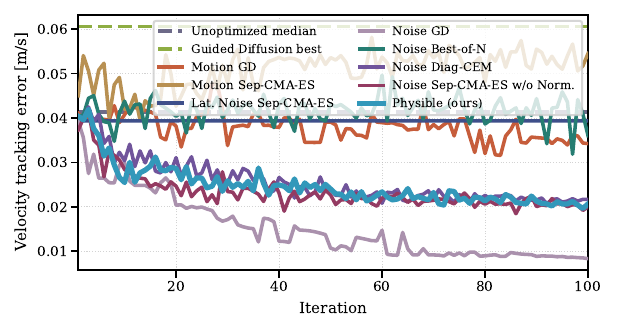}}
    \subfigure[Style Classification Accuracy]{\includegraphics[width=0.49\textwidth, trim={5 5 5 5}]{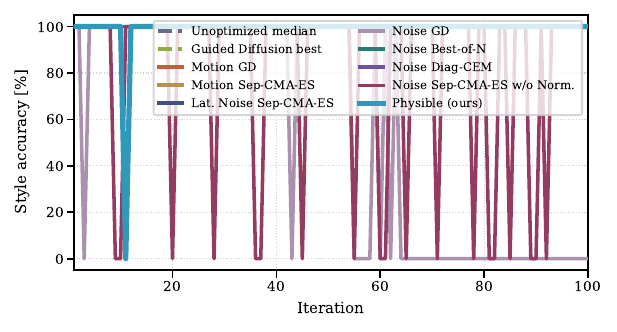}}\\
    \caption{Convergence of kinematic feasibility optimization methods for a single motion horizon and constant forward reference velocity on the SMPL humanoid with \emph{neutral} walking style with velocity command $\mathbf v^{\mathrm{cmd}}=(1,0,0)$.}
    \label{fig:kinoptim_convergence_smpl}
\end{figure}

In \figref{fig:kinoptim_convergence_smpl}, we compare the convergence behavior of the different methods for kinematic feasibility optimization. We optimize a \emph{single} horizon (with the autoregressive motion-consistency objective disabled) for the \ac{SMPL} humanoid under a velocity command $\mathbf v^{\mathrm{cmd}}=(1,0,1)$ (forward walking) and the \emph{neutral} style. All methods in this comparison use \ac{WE} smoothing with $\kappa_{\mathrm{WE}}=1$. We report metrics for the median sample (by loss/fitness) from a batch of $64$ unoptimized motions, the best sample from $64$ motions optimized in parallel with \emph{Guided Diffusion}, and the best sample per iteration for all remaining methods.

The results show that \emph{Noise GD} reduces the fitness fastest, but it does not decrease ground penetration as effectively as the other approaches and the motion style degrades as iterations increase. Notably, despite identical loss weights, \emph{Noise GD} appears to emphasize velocity tracking. While it performs well on contact pattern preservation, it still falls short of the best sampling-based noise optimization methods in terms of reducing ground penetration. \emph{Diag-CEM}~\citep{pinneri2021sample} (\ac{CEM}~\citep{rubinstein1999cross} with diagonal Gaussian covariance) achieves performance comparable to \emph{Sep-CMA-ES}, albeit with slightly slower convergence. Direct motion optimization, in contrast, quickly increases jerk, whereas jerk grows more gradually for the noise optimization methods. Overall, the proposed \emph{Physible} approach shows strong convergence across objectives and constraints while only modestly increasing jerk relative to the unoptimized motion.

\clearpage
\subsubsection{Variability across Optimization Seeds}
\label{apx:ssub:kinematic_seed_variability}
We evaluate the sensitivity of Physible to the random seed using ten seeds ($0$--$9$), all $100$ motion styles, and $150$ generated frames per style on the \ac{SMPL} humanoid. The velocity-command sampling seed is fixed independently at $0$, and the saved commands and optimization configurations match across runs. Table~\ref{tab:kinoptim:smpl_humanoid:seeds} reports average motion differences over all $45$ seed pairs and the mean $\pm$ sample standard deviation of the ten aggregate objective and constraint scores. Each seed score is averaged equally over the $100$ style-specific sequences, using the definitions in Appendix~\ref{apx:sub:metrics}.

The variability across seeds is small: the standard deviation of the best weighted loss is approximately $0.85\%$ of its mean, and the corresponding ratios for velocity tracking and motion consistency are $0.90\%$ and $1.84\%$. Contact-pattern violation and ground slip also vary little, with relative standard deviations of approximately $0.68\%$ and $1.55\%$, respectively. Self-collision penetration has a larger relative variation ($8.36\%$), but its absolute standard deviation is only \SI{0.00310}{mm}. The generated motions are also repeatable, with average pairwise root-trajectory \ac{RMSE} of \SI{0.0351}{m}, body-position distance of \SI{0.0246}{m}, and root-orientation and joint-angle differences of \SI{0.0199}{rad} and \SI{0.0500}{rad}. These results indicate that the optimizer consistently finds solutions of similar quality across the tested seeds, supporting the use of population-based search to mitigate sensitivity to poor local minima, a common concern in gradient-based optimization. These observations provide empirical evidence of low seed sensitivity within the tested evaluation setting.

\begin{table}[htbp]
\centering
\caption{Optimizer-seed sensitivity of Physible on the SMPL humanoid: ten seeds, $100$ styles, and $150$ frames per style, with identical velocity commands. Motion rows report average pairwise differences; objectives and constraints report mean $\pm$ sample standard deviation across seeds after averaging over styles. Alignment and metric definitions are given in Appendix~\ref{apx:sub:metrics}.}
\label{tab:kinoptim:smpl_humanoid:seeds}
\begingroup
\setlength{\tabcolsep}{7pt}
\renewcommand{\arraystretch}{1.1}
\small
\begin{tabular}{lr}
\toprule
Metric & Result \\
\midrule
Pairwise root trajectory RMSE [m] & $3.51\times10^{-2}$ \\
Pairwise root orientation difference [rad] & $1.99\times10^{-2}$ \\
Pairwise joint angle difference [rad] & $5.00\times10^{-2}$ \\
Pairwise body position distance [m] & $2.46\times10^{-2}$ \\
\midrule
Best weighted loss & $3.81\times10^{-1}\pm3.24\times10^{-3}$ \\
Velocity tracking & $3.78\times10^{-2}\pm3.41\times10^{-4}$ \\
Motion consistency & $1.82\times10^{-4}\pm3.35\times10^{-6}$ \\
\addlinespace
Contact-pattern violation [mm] & $1.01\pm6.84\times10^{-3}$ \\
Ground penetration [mm] & $0.68\pm0.01$ \\
Ground slip [m/s] & $5.43\times10^{-2}\pm8.44\times10^{-4}$ \\
Self-collision penetration [mm] & $3.71\times10^{-2}\pm3.10\times10^{-3}$ \\
\bottomrule
\end{tabular}
\endgroup
\end{table}

\clearpage
\subsection{Ablation Studies}
\label{apx:sub:kinematic_ablations}
\subsubsection{Initial and Posterior Noise Optimization}
\label{apx:ssub:kinematic_noise_groups}
The \emph{Lat. Noise Sep-CMA-ES} ablation in Tabs.~\ref{tab:kinoptim:smpl_humanoid} and~\ref{tab:kinoptim:bipedal_robot} optimizes only the initial/latent noise, holding posterior noise fixed. This restriction yields only modest gains over the unoptimized baseline and falls short of joint initial and posterior noise optimization.

\subsubsection{Noise Standardization}
\label{apx:ssub:kinematic_noise_standardization}
As shown in \tabref{tab:kinoptim:smpl_humanoid}, omitting \eqnref{eq:noise_norm} yields a style accuracy of \SI{76.01}{\percent}, overall FID of $22.00$, and mean per-style FID of $74.21$. With standardization, these values improve to \SI{87.73}{\percent}, $14.41$, and $43.95$, respectively: an accuracy gain of $11.72$ percentage points and a $40.8\%$ reduction in mean per-style FID. Standardization also improves \ac{LDJ} from $-7.29$ to $-6.59$ and decreases mean motion-consistency error from $1.92\times10^{-4}$ to $1.52\times10^{-4}$.

These gains come with higher mean ground penetration (0.676 vs. \SI{0.541}{mm}), self-collision penetration (0.0152 vs. \SI{0.00922}{mm}), slip (0.0484 vs. \SI{0.0401}{\meter\per\second}), and velocity-tracking error ($0.0392$ vs. $0.0334$). Thus, the corrected NNS results support a tradeoff between feasibility and style preservation, with residual violations in both methods.

\subsubsection{Contact Pattern Preservation}
\label{apx:ssub:kinematic_contact_ablation}
One aspect of kinematic feasibility is avoiding (excessive) ground penetrations, which are physically infeasible. A natural question is why we favor the \emph{Contact Pattern Preservation} constraint (\secref{sub:kinematic_feasibility_constraints}) over directly minimizing the ground penetration distance. The key reason is to prevent \emph{floating} characters, i.e., motions that exhibit extended time intervals without ground contact.\footnote{Short no-contact phases can be feasible or even desired (e.g., jumping or running). However, if a motion lacks ground contact for long periods, it will generally not be dynamically feasible.} When we co-optimize dynamical feasibility, floating is implicitly discouraged because, without contacts, the accelerations encoded in the motion cannot be supported due to the absence of contact forces. In contrast, under purely kinematic feasibility optimization, this implicit safeguard is absent: optimizing only a ground penetration constraint can yield ``seemingly'' valid solutions that simply avoid penetration by lifting the character off the ground for long durations. We therefore employ the contact pattern preservation constraint, which enforces (i) contact---while limiting penetration---during phases where contact should occur (thereby preventing floating), and (ii) separation during phases where contact should not occur.

To substantiate this reasoning, we perform an ablation study that compares the contact behavior obtained when using the \emph{Ground Penetration} constraint versus the proposed \emph{Contact Pattern Preservation} constraint. The three conditions use the settings in Appendix~\ref{apx:par:kinematic_case_settings}.

\begin{figure}[h!]
    \centering
    \includegraphics[page=1,width=0.88\linewidth, trim={5, 5, 5, 5}]{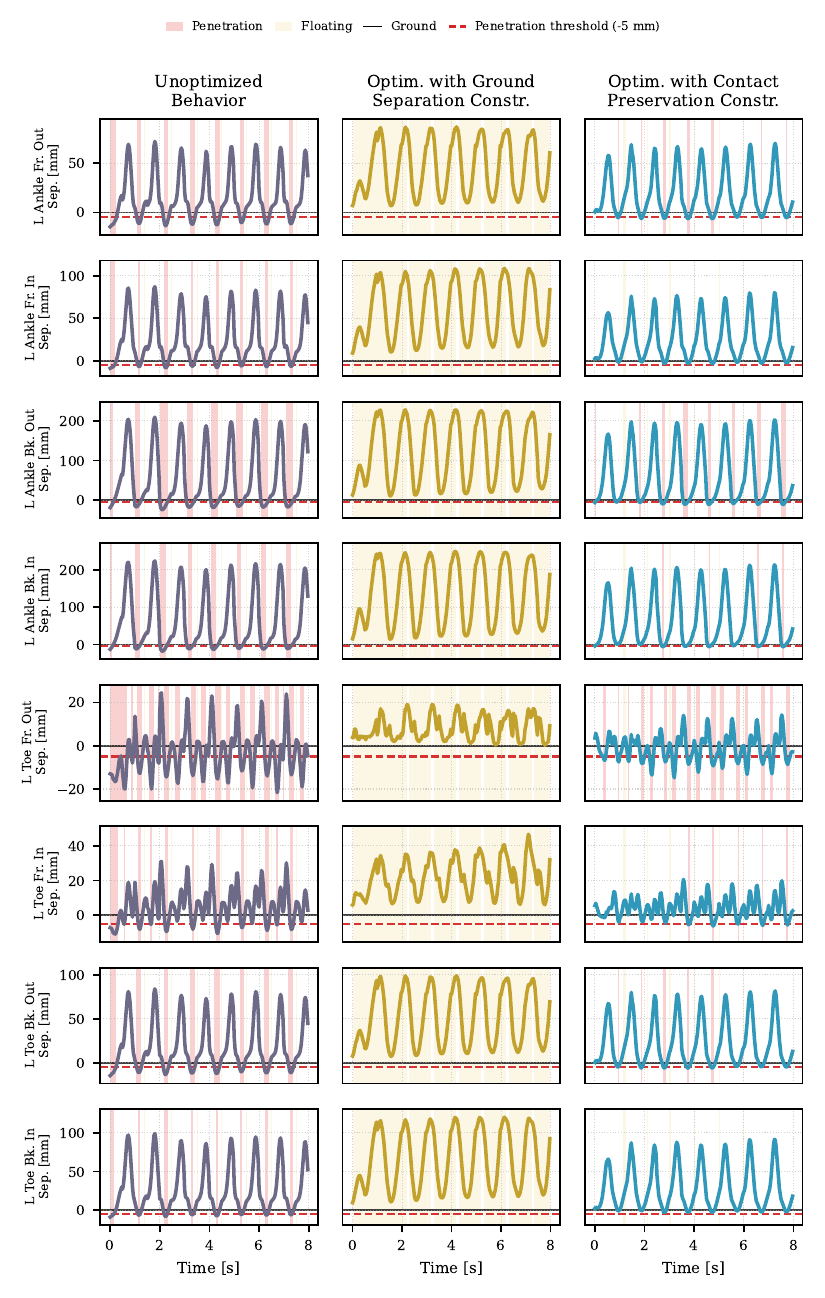}
    \caption{Part~1 of the ablation study compares two constraints for limiting (excessive) ground penetration during kinematic feasibility optimization on the SMPL humanoid for the \emph{neutral} motion style: the baseline \emph{ground penetration} constraint versus the proposed \emph{contact pattern preservation} constraint. For each contact point, the left column shows the ground separation distance of the median unoptimized motion, the center column shows the best motion optimized with the \emph{ground penetration} constraint, and the right column shows the best motion optimized with the \emph{contact pattern preservation} constraint. Red shaded intervals indicate excessive-contact periods where penetration exceeds the threshold (as used by the \emph{contact pattern preservation} constraint), while yellow shaded intervals indicate phases where none of the robot’s contact points are in contact with the ground.}
    \label{fig:kinoptim_results:ground_separation_ablation:smpl_neutral:part1}
\end{figure}

\begin{figure}[h!]
    \centering
    \includegraphics[page=2,width=0.88\linewidth, trim={5, 5, 5, 5}]{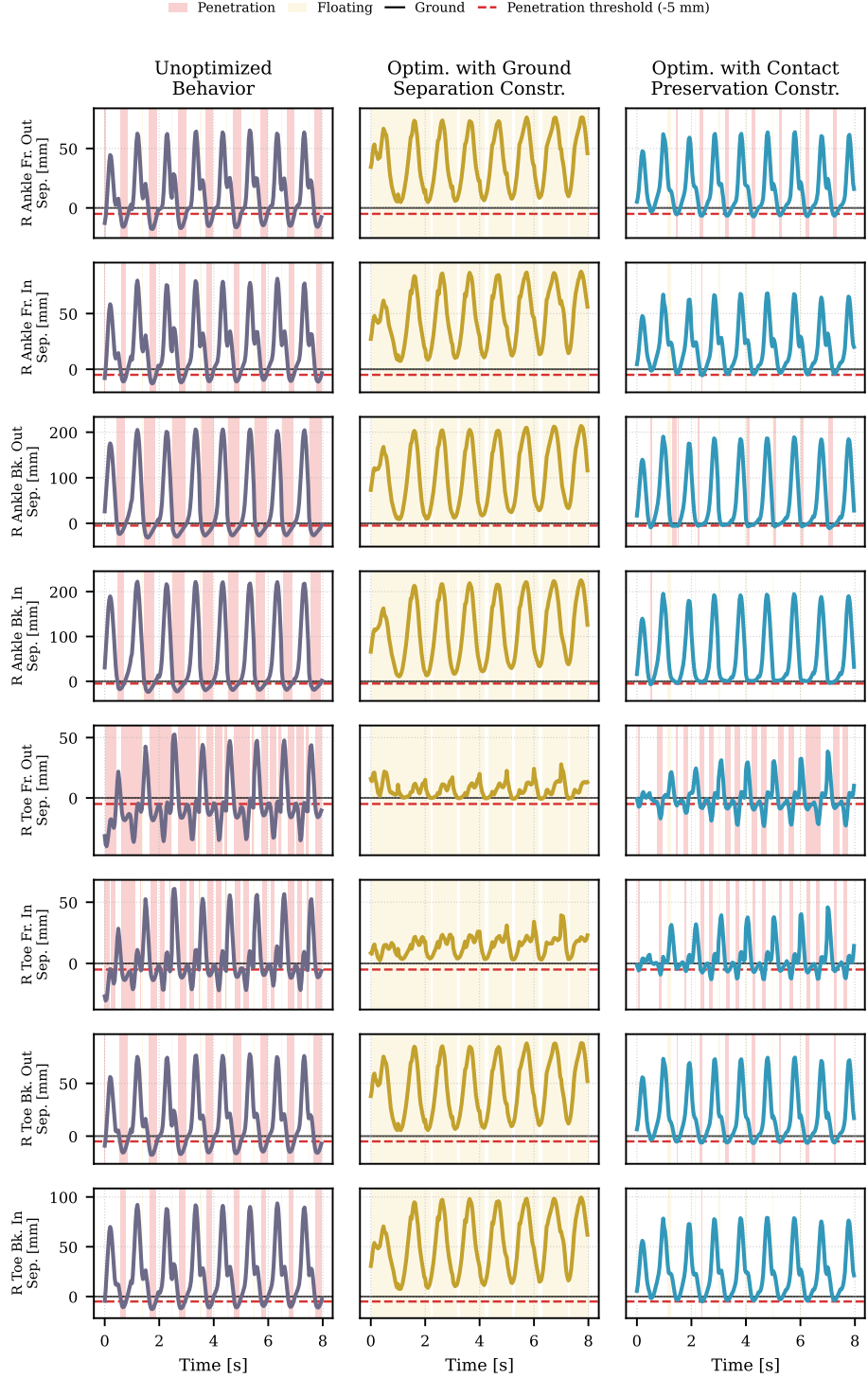}
    \caption{Part~2 of the ablation study compares two constraints for limiting (excessive) ground penetration during kinematic feasibility optimization on the SMPL humanoid for the \emph{neutral} motion style: the baseline \emph{ground penetration} constraint versus the proposed \emph{contact pattern preservation} constraint. For each contact point, the left column shows the ground separation distance of the median unoptimized motion, the center column shows the best motion optimized with the \emph{ground penetration} constraint, and the right column shows the best motion optimized with the \emph{contact pattern preservation} constraint. Red shaded intervals indicate excessive-contact periods where penetration exceeds the threshold (as used by the \emph{contact pattern preservation} constraint), while yellow shaded intervals indicate phases where none of the robot’s contact points are in contact with the ground.}
    \label{fig:kinoptim_results:ground_separation_ablation:smpl_neutral:part2}
\end{figure}

Figure~\ref{fig:kinoptim_results:ground_separation_ablation:smpl_neutral:comparison} compares unoptimized motion and the two losses for the \ac{SMPL} humanoid in the \emph{neutral} walking style, using matched foot close-ups, signed ground clearance, and intended versus observed foot contacts. Over the $240$-frame (\SI{8}{s}) clips, the ground penetration distance loss reduces mean penetration from \SI{3.826}{mm} to \SI{0.002}{mm}, but leaves all checked contact points above ground in 91.7\% of frames, with a longest contact loss of \SI{2.20}{s}. Contact pattern preservation reduces mean penetration to \SI{1.166}{mm}, while limiting frames without contact to 4.2\% and the longest contact loss to \SI{0.13}{s} (unoptimized: 3.3\% and \SI{0.07}{s}). Thus, it reduces penetration while avoiding the prolonged floating induced by the ground penetration distance loss, although some penetrations still exceed the \SI{5}{mm} allowance. Figs.~\ref{fig:kinoptim_results:ground_separation_ablation:smpl_neutral:part1} and~\ref{fig:kinoptim_results:ground_separation_ablation:smpl_neutral:part2} provide the corresponding contact-point ground separation plots.

\begin{figure}[p]
    \centering
    \includegraphics[width=\linewidth]{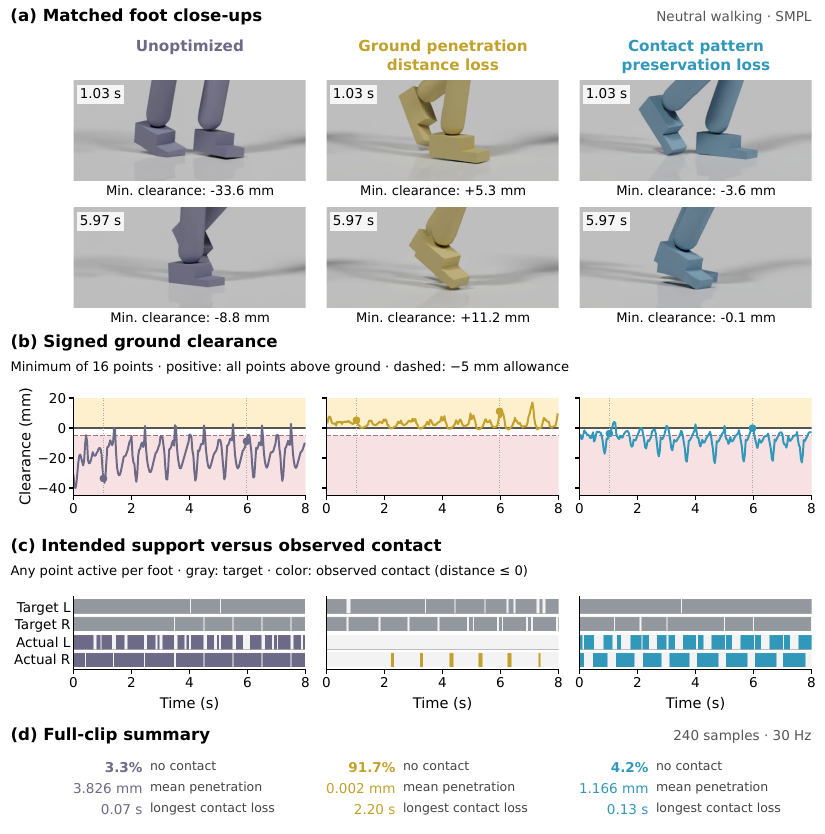}
    \caption{Ablation of the contact pattern preservation loss for \emph{neutral} walking on the SMPL humanoid. Columns compare unoptimized motion, ground penetration distance loss, and contact pattern preservation loss. (a) Matched foot close-ups at $1.03$ and $5.97\,\mathrm{s}$, with minimum signed ground clearance. (b) Minimum clearance over 16 checked points; positive values indicate that all points are above ground, and the dashed line marks the $-5\,\mathrm{mm}$ penetration allowance. (c) Target and observed left/right foot contacts; a foot is in contact when any checked point has nonpositive separation. (d) Summary over 240 samples at \SI{30}{Hz}, reporting the fraction of frames without contact, mean penetration, and longest continuous contact loss.}
    \label{fig:kinoptim_results:ground_separation_ablation:smpl_neutral:comparison}
\end{figure}

\clearpage
\subsection{Motion Case Studies for Qualitative Analysis}
\label{apx:sub:kinematic_case_studies}
\subsubsection{Neutral Walking: Comparison of Optimization Methods}
\label{apx:ssub:kinematic_neutral}
\begin{figure}[t]
    \centering
    \includegraphics[width=\linewidth]{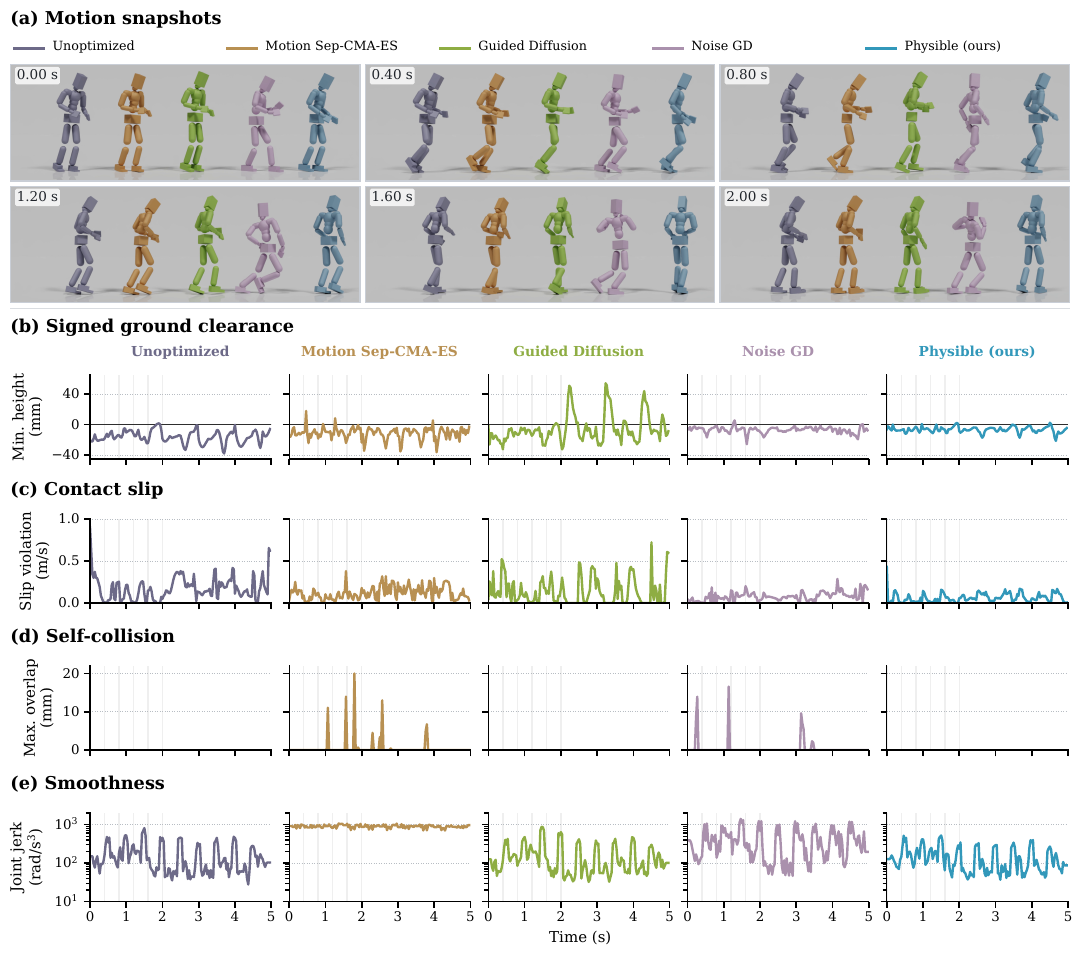}
    \caption{
        Kinematic feasibility optimization methods comparison for \emph{neutral} walking on the \ac{SMPL} humanoid.
        \textbf{(a)} Matched-time snapshots of unoptimized motion, Motion Sep-CMA-ES, Guided Diffusion, Noise GD, and Physible.
        \textbf{(b)} Minimum signed height across contact points; negative values indicate ground penetration, while positive values indicate that all contact points are above ground.
        \textbf{(c)} Contact slip violation, averaged over all contact points after contact masking and velocity-threshold subtraction.
        \textbf{(d)} Maximum self-collision overlap across evaluated body pairs.
        \textbf{(e)} Root-mean-square joint jerk on a logarithmic scale.
        Diagnostics cover the full \SI{5}{s} clip, with a shared scale within each row; gray bands mark snapshot times. Physible reduces penetration and slip while retaining jerk closer to the unoptimized motion, although some ground penetration remains.
    }
    \label{fig:kinematic-feasibility-methods}
\end{figure}

\begin{figure}[h!]
    \centering
    \includegraphics[width=1.0\linewidth, trim={5, 5, 5, 5}]{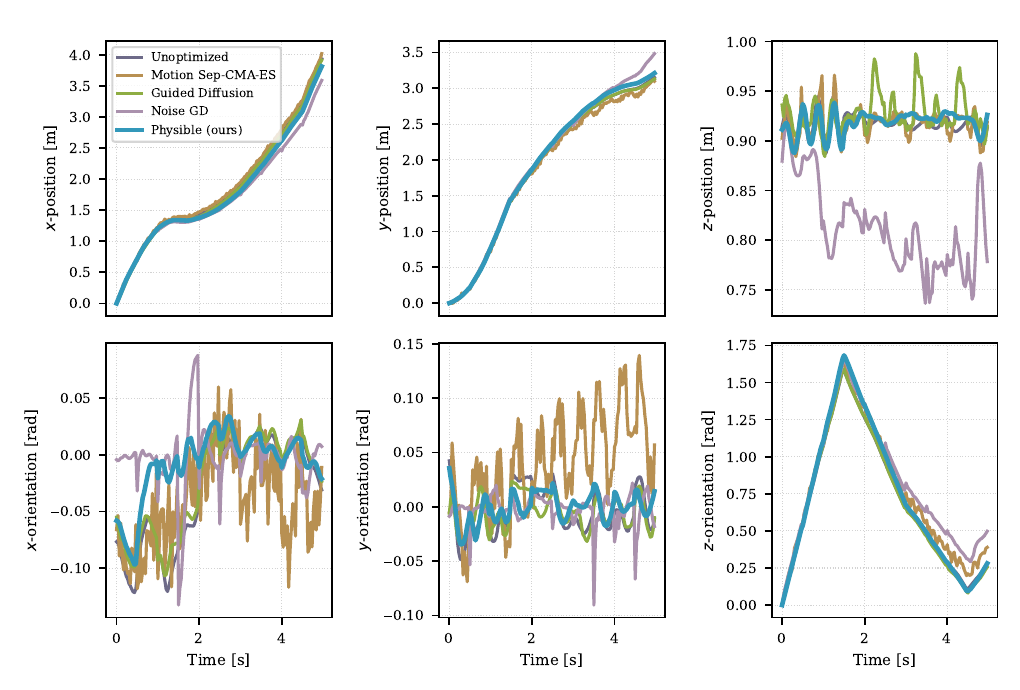}
    \caption{Behavior comparison of the \emph{root motion} for the kinematic feasibility optimization for $150$ time steps of the \emph{neutral} walking style on the SMPL humanoid character.}
    \label{fig:kinoptim_results:behavior_comparison:smpl_neutral:root_motion}
\end{figure}

\begin{figure}[h!]
    \centering
    \includegraphics[page=1, width=1.0\linewidth, trim={5, 5, 5, 5}]{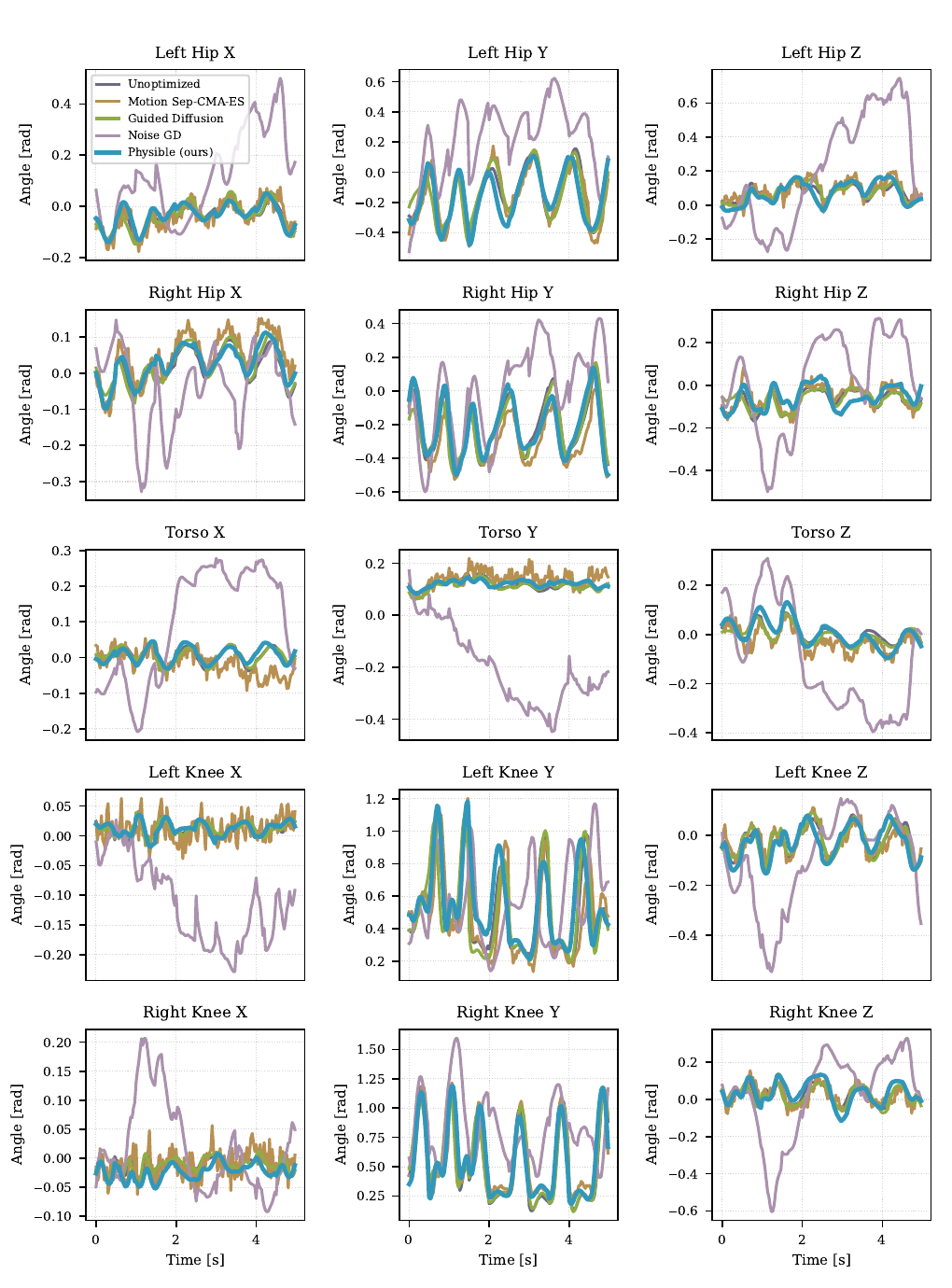}
    \caption{Behavior comparison of the \emph{joint motion} for the kinematic feasibility optimization for $150$ time steps of the \emph{neutral} walking style on the SMPL humanoid character - part 1.}
    \label{fig:kinoptim_results:behavior_comparison:smpl_neutral:joint_motion:part1}
\end{figure}

\begin{figure}[h!]
    \centering
    \includegraphics[page=2, width=1.0\linewidth, trim={5, 5, 5, 5}]{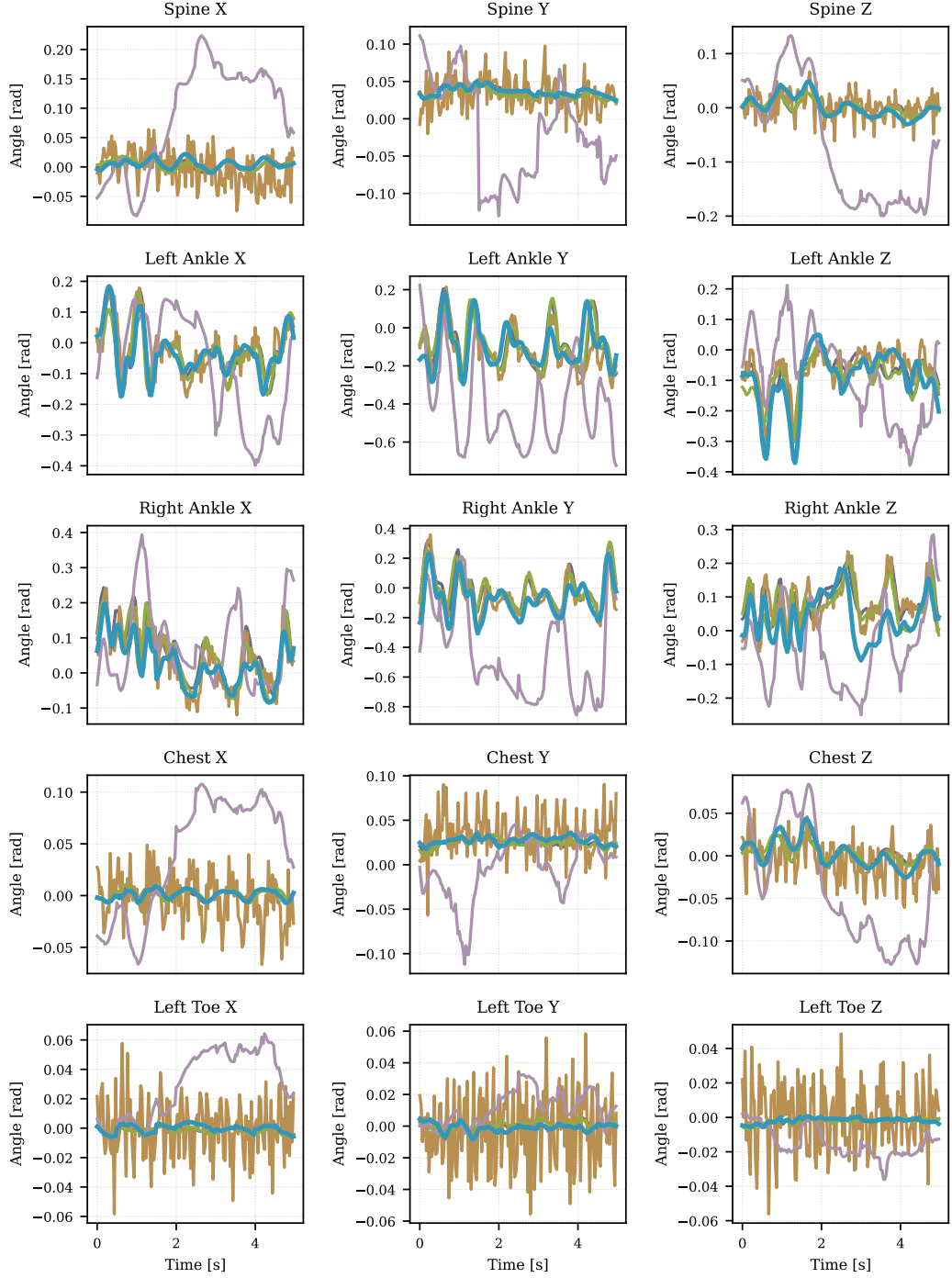}
    \caption{Behavior comparison of the \emph{joint motion} for the kinematic feasibility optimization for $150$ time steps of the \emph{neutral} walking style on the SMPL humanoid character - part 2.}
    \label{fig:kinoptim_results:behavior_comparison:smpl_neutral:joint_motion:part2}
\end{figure}

\begin{figure}[h!]
    \centering
    \includegraphics[page=3, width=1.0\linewidth, trim={5, 5, 5, 5}]{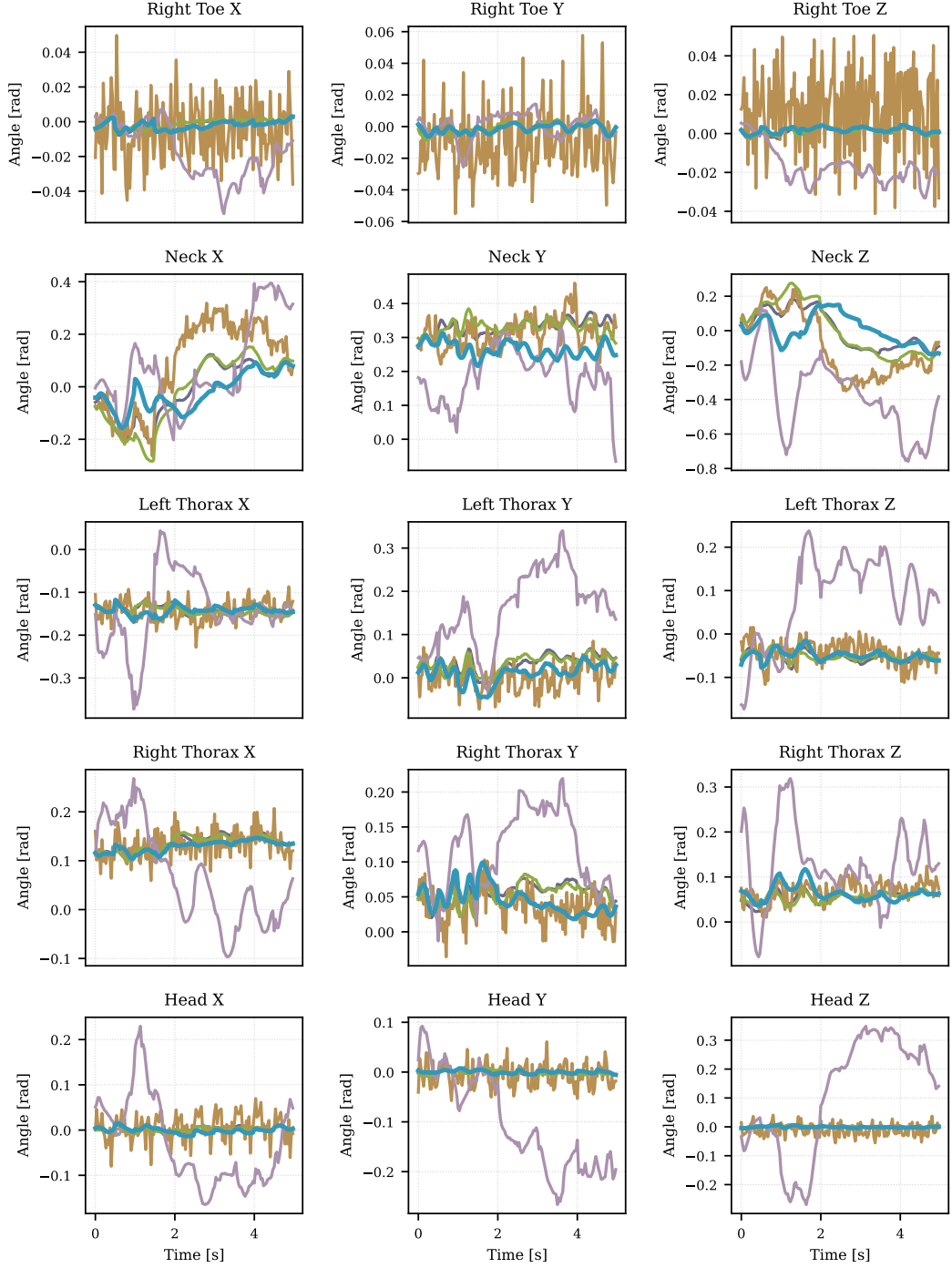}
    \caption{Behavior comparison of the \emph{joint motion} for the kinematic feasibility optimization for $150$ time steps of the \emph{neutral} walking style on the SMPL humanoid character - part 3.}
    \label{fig:kinoptim_results:behavior_comparison:smpl_neutral:joint_motion:part3}
\end{figure}

\begin{figure}[h!]
    \centering
    \includegraphics[page=4, width=1.0\linewidth, trim={5, 5, 5, 5}]{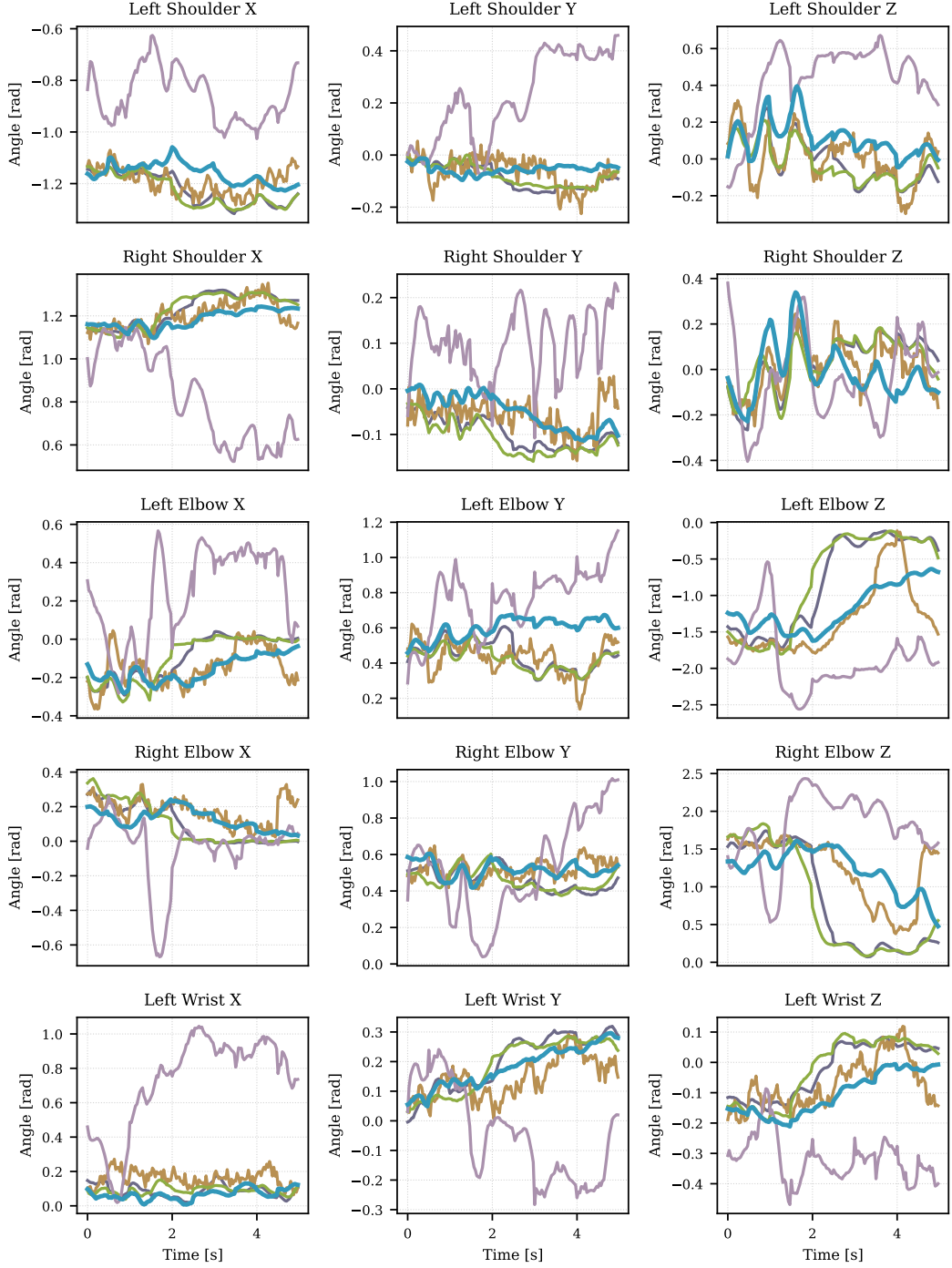}
    \caption{Behavior comparison of the \emph{joint motion} for the kinematic feasibility optimization for $150$ time steps of the \emph{neutral} walking style on the SMPL humanoid character - part 4.}
    \label{fig:kinoptim_results:behavior_comparison:smpl_neutral:joint_motion:part4}
\end{figure}

\begin{figure}[h!]
    \centering
    \includegraphics[width=1.0\linewidth, trim={5, 5, 5, 5}]{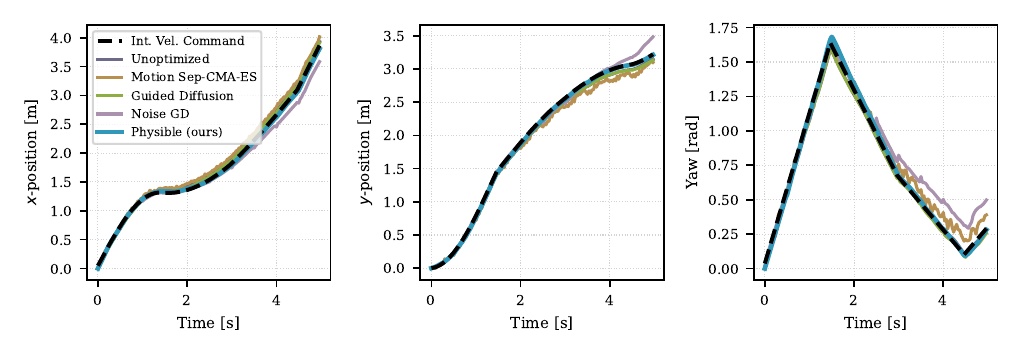}
    \caption{Behavior comparison of the \emph{root motion} in the x-y plane with integrated velocity command-based reference for the kinematic feasibility optimization for $150$ time steps of the \emph{neutral} walking style on the SMPL humanoid character.}
    \label{fig:kinoptim_results:behavior_comparison:smpl_neutral:root_planar_with_reference}
\end{figure}

\begin{figure}[h!]
    \centering
    \includegraphics[width=0.7\linewidth, trim={5, 5, 5, 5}]{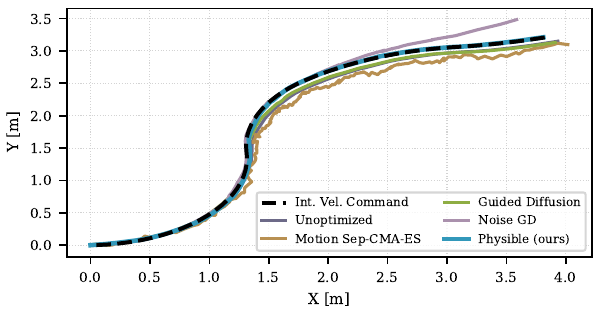}
    \caption{Behavior comparison of the \emph{positional root motion} in the x-y plane with integrated velocity command-based reference for the kinematic feasibility optimization for $150$ time steps of the \emph{neutral} walking style on the SMPL humanoid character.}
    \label{fig:kinoptim_results:behavior_comparison:smpl_neutral:root_position_planar_with_reference}
\end{figure}

\begin{figure}[h!]
    \centering
    \includegraphics[width=1.0\linewidth, trim={5, 5, 5, 5}]{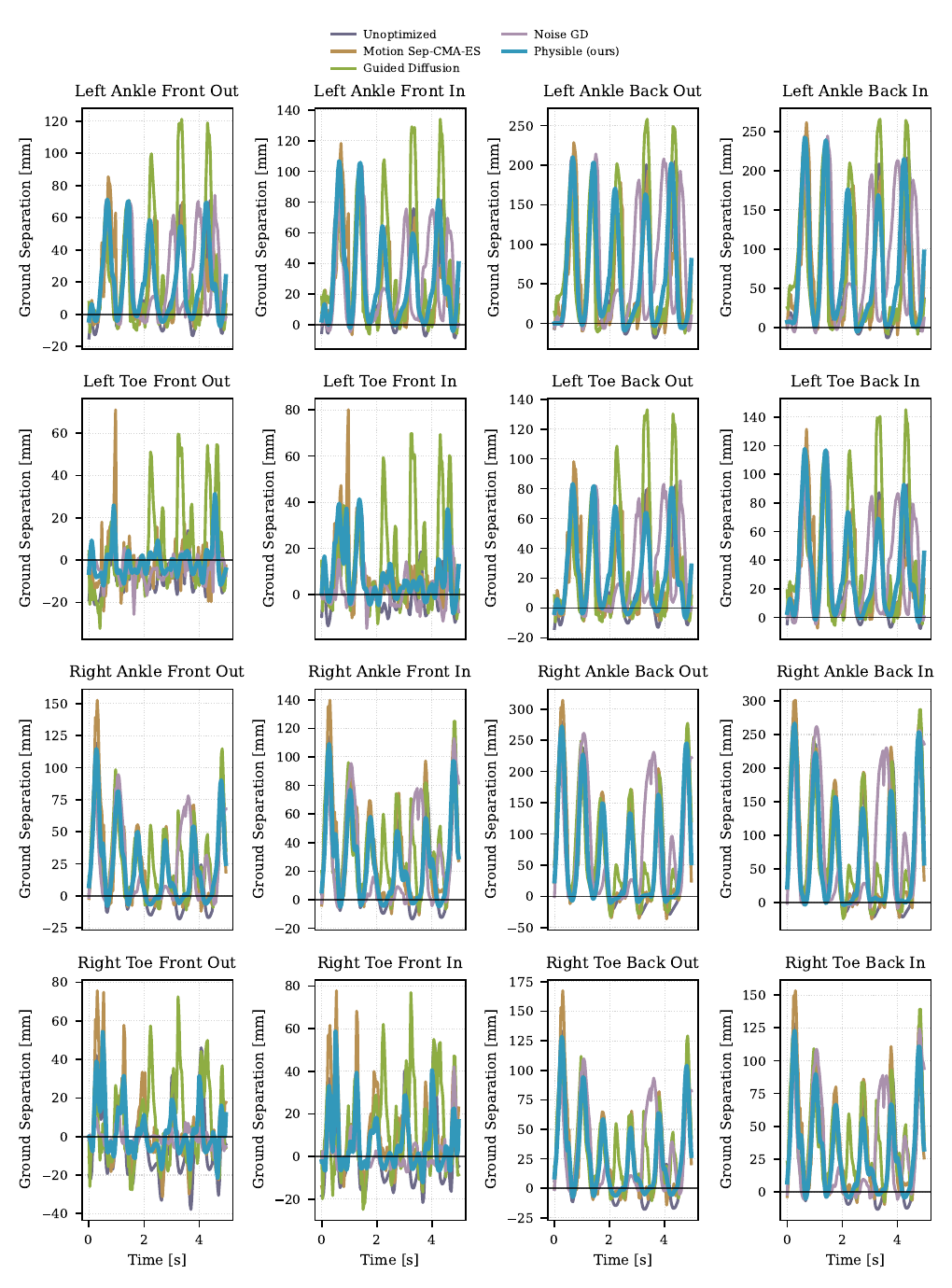}
    \caption{Behavior comparison of the \emph{ground separation} for the kinematic feasibility optimization for $150$ time steps of the \emph{neutral} walking style on the SMPL humanoid character.}
    \label{fig:kinoptim_results:behavior_comparison:smpl_neutral:ground_separation}
\end{figure}

\begin{figure}[h!]
    \centering
    \includegraphics[width=1.0\linewidth, trim={5, 5, 5, 5}]{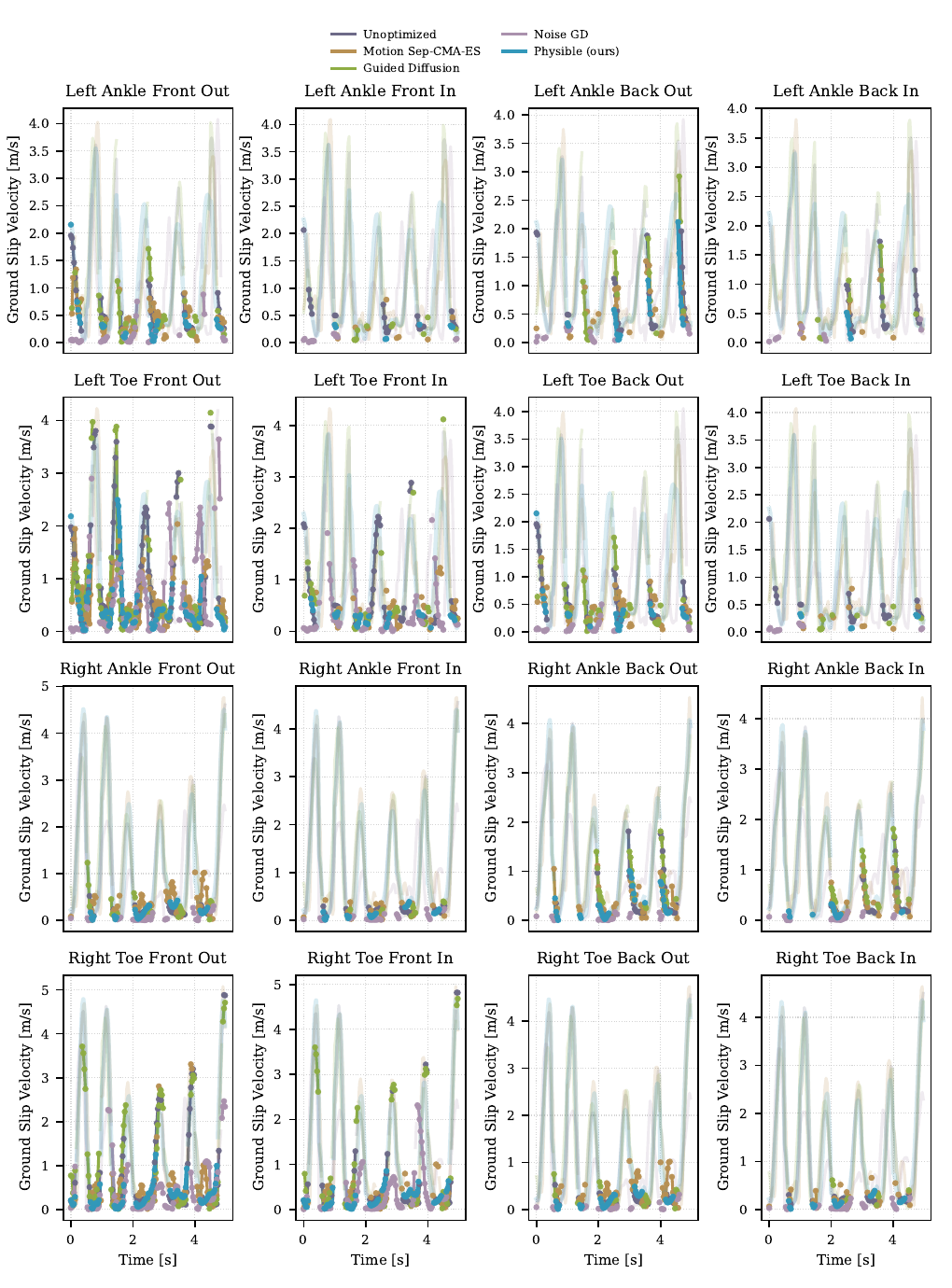}
    \caption{Behavior comparison of the \emph{ground slip velocity} for the kinematic feasibility optimization for $150$ time steps of the \emph{neutral} walking style on the SMPL humanoid character. The full-opacity and markers denoted active contacts (i.e., with a semi-negative ground separation distance). The transparent lines denote the tangential velocities of the contact points when they are not in contact and separated from the ground.}
    \label{fig:kinoptim_results:behavior_comparison:smpl_neutral:ground_slip}
\end{figure}

\begin{figure}[h!]
    \centering
    \includegraphics[page=1, width=1.0\linewidth, trim={5, 5, 5, 5}]{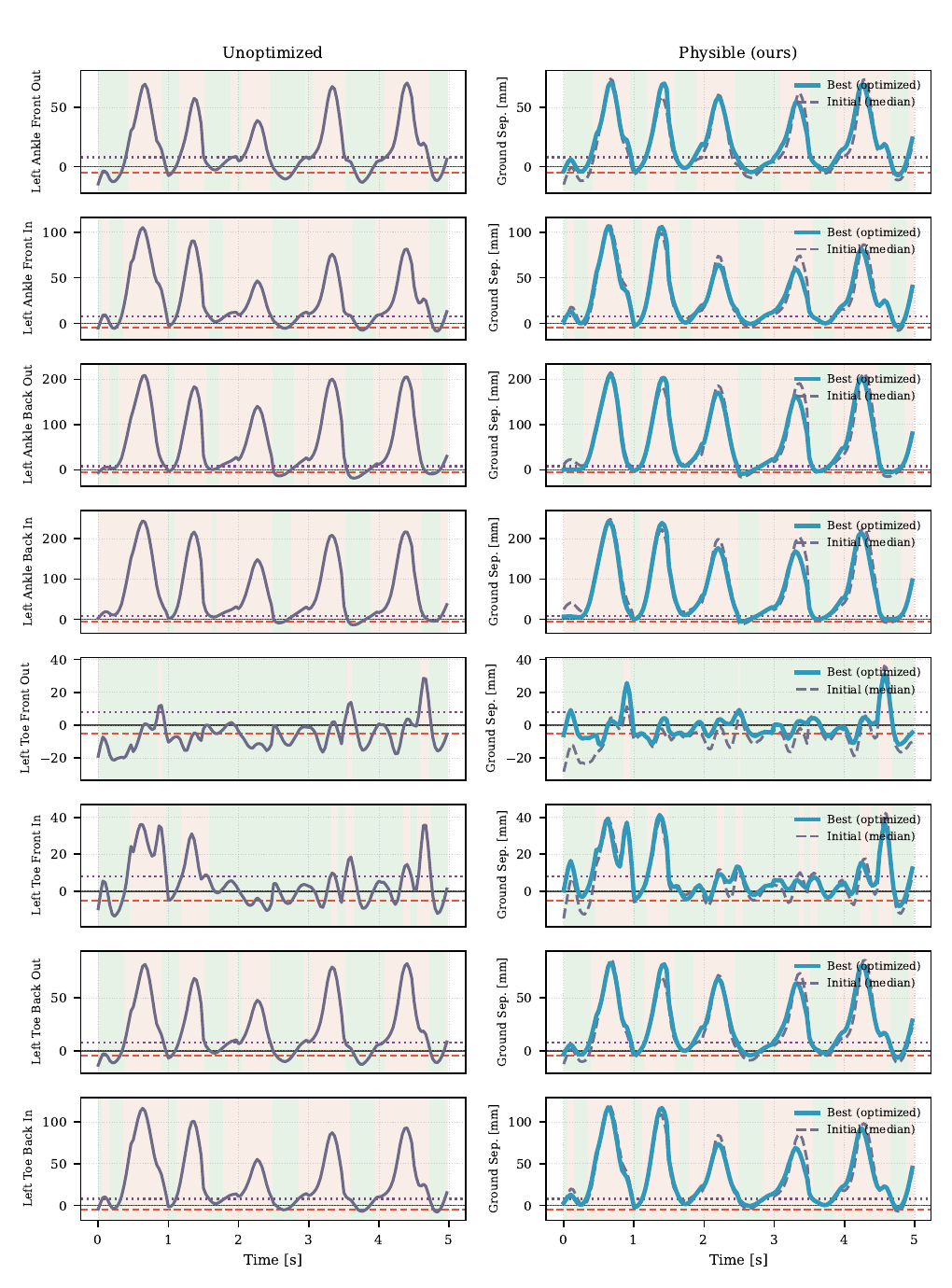}
    \caption{Behavior comparison of the \emph{contact pattern preservation} for the kinematic feasibility optimization for $150$ time steps of the \emph{neutral} walking style on the SMPL humanoid character. The soft green and warm tints visualize time periods where the respective contact point should be in contact and \emph{not} in contact with the ground, based on the contact pattern identified via a majority vote using the height threshold (purple-dashed line) using a batch of diffused, unoptimized motions. The red-dashed lines denotes the maximum-allowed ground penetration - Part 1.}
    \label{fig:kinoptim_results:behavior_comparison:smpl_neutral:contact_pattern_preservation:part1}
\end{figure}

\begin{figure}[h!]
    \centering
    \includegraphics[page=2, width=1.0\linewidth, trim={5, 5, 5, 5}]{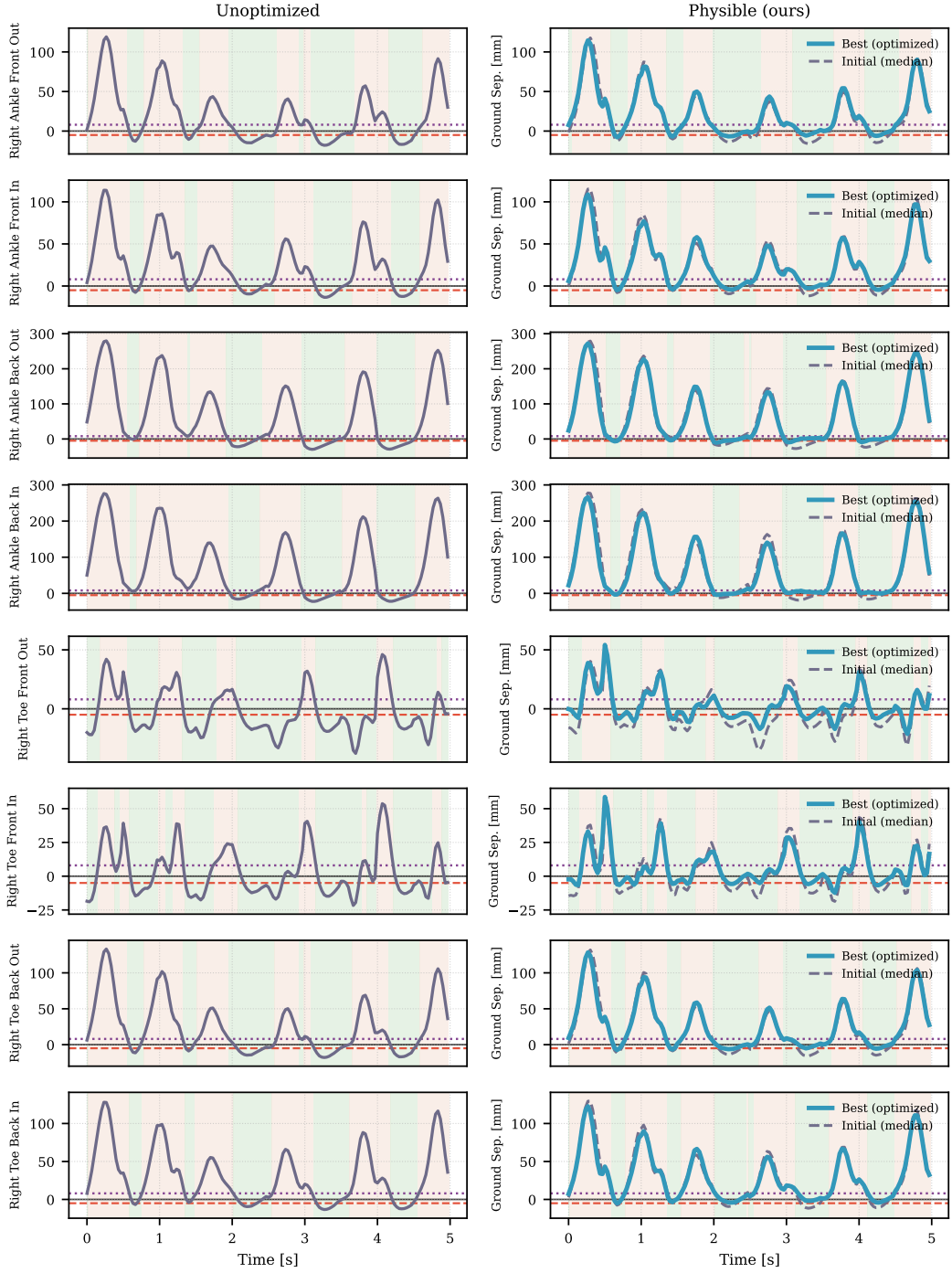}
    \caption{Behavior comparison of the \emph{contact pattern preservation} for the kinematic feasibility optimization for $150$ time steps of the \emph{neutral} walking style on the SMPL humanoid character. The soft green and warm tints visualize time periods where the respective contact point should be in contact and \emph{not} in contact with the ground, based on the contact pattern identified via a majority vote using the height threshold (purple-dashed line) using a batch of diffused, unoptimized motions. The red-dashed lines denote the maximum-allowed ground penetration - Part 2.}
    \label{fig:kinoptim_results:behavior_comparison:smpl_neutral:contact_pattern_preservation:part2}
\end{figure}

\begin{figure}[h!]
    \centering
    \includegraphics[width=1.0\linewidth, trim={5, 5, 5, 5}]{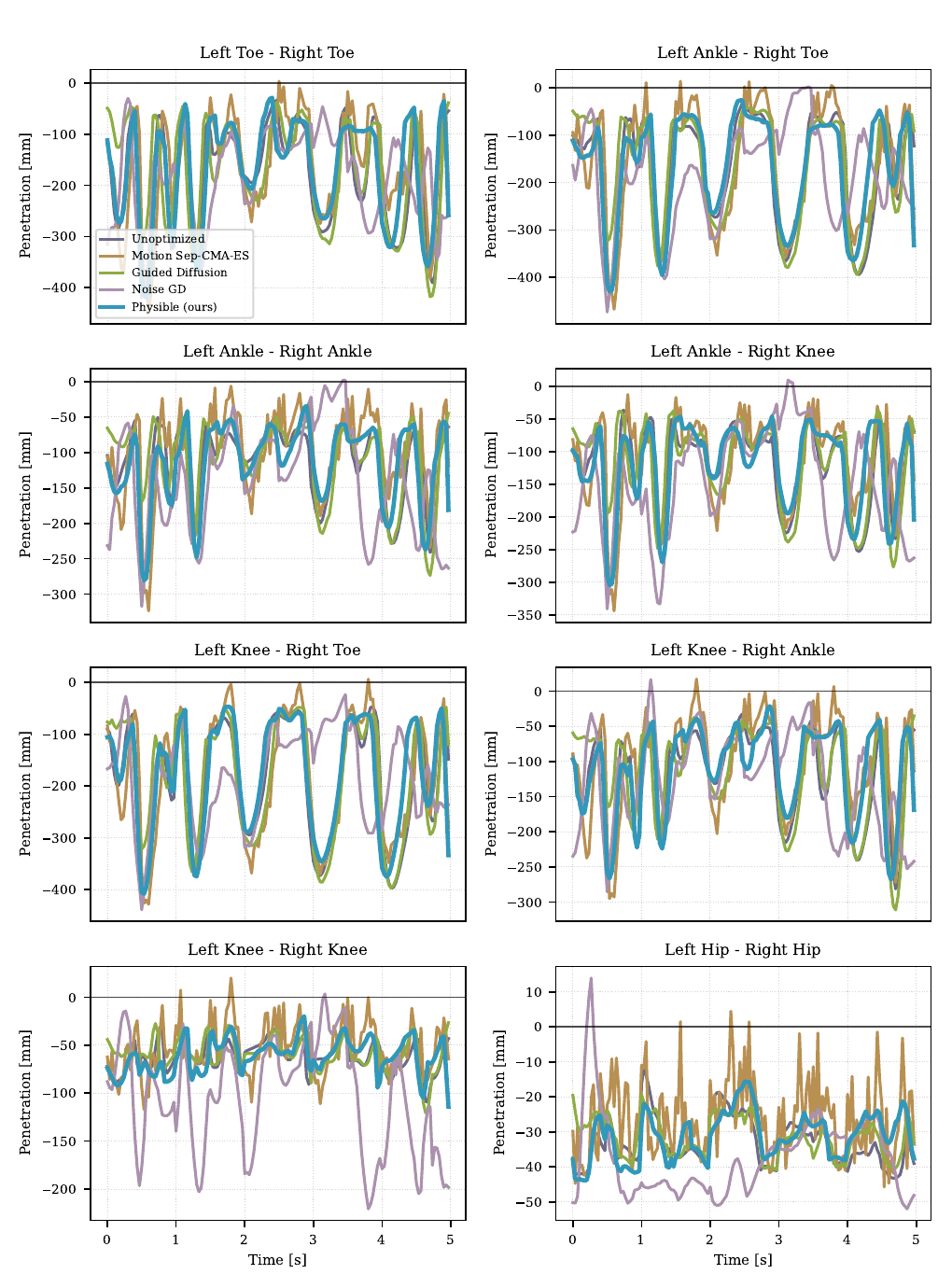}
    \caption{Behavior comparison of the \emph{self-collision penetration} for the kinematic feasibility optimization for $150$ time steps of the \emph{neutral} walking style on the SMPL humanoid character.}
    \label{fig:kinoptim_results:behavior_comparison:smpl_neutral:self_collision}
\end{figure}

In Figs.~\ref{fig:kinoptim_results:behavior_comparison:smpl_neutral:root_motion}--\ref{fig:kinoptim_results:behavior_comparison:smpl_neutral:self_collision}, we compare the kinematic-feasibility-optimized behavior over $150$ time steps for the \ac{SMPL} humanoid in the \emph{neutral} style. We include root and joint trajectories, the resulting SE(2) motion relative to the integrated velocity reference, ground separation, and slip velocity per contact point, contact-pattern preservation, and penetration distances for each self-collision pair. For clarity, we show a subset of methods: the unoptimized baseline, direct motion optimization with Sep-CMA-ES, \emph{Guided Diffusion}, noise \ac{GD}, and the proposed \emph{Physible} noise optimization with \ac{Sep-CMA-ES}. Figure~\ref{fig:kinematic-feasibility-methods} summarizes this comparison through matched-time snapshots and ground-clearance, contact-slip, self-collision, and joint-jerk diagnostics.

The plots again highlight the \emph{jittery} artifacts introduced by direct motion optimization (\emph{Motion Sep-CMA-ES}). For velocity tracking, the proposed method is the only approach that follows the integrated command almost perfectly, with negligible drift over the full \SI{5}{s} horizon; the other methods (including the unoptimized motion) begin to drift from the reference around \SI{1.5}{s}. Figs.~\ref{fig:kinoptim_results:behavior_comparison:smpl_neutral:contact_pattern_preservation:part1} and \ref{fig:kinoptim_results:behavior_comparison:smpl_neutral:contact_pattern_preservation:part2} further show that \emph{Physible} largely preserves the contact pattern, maintaining contact whenever the unoptimized motion is in contact. Crucially, it also substantially reduces ground penetration---especially the deepest events---so that the penetration threshold is exceeded only rarely and by a small margin. On a related note, the ground slip velocities are with \emph{Physible} also significantly reduced compared to the unoptimized motion and to a larger extent for the other optimization methods. Only for a single contact point (\emph{Left Ankle Back Out}), the slip velocity remains slightly higher at certain time steps. For the \emph{neutral} style, self-collision is not a major issue in the unoptimized motion; however, methods that noticeably degrade motion quality and deviate from the style, such as \emph{Motion Sep-CMA-ES} and \emph{Noise GD}, can introduce leg--leg self-collisions.

\clearpage
\subsubsection{Cat Motion: Ground-Penetration Reduction}
\label{apx:ssub:kinematic_cat}
\begin{figure}[!htbp]
    \centering
    \includegraphics[width=\columnwidth]{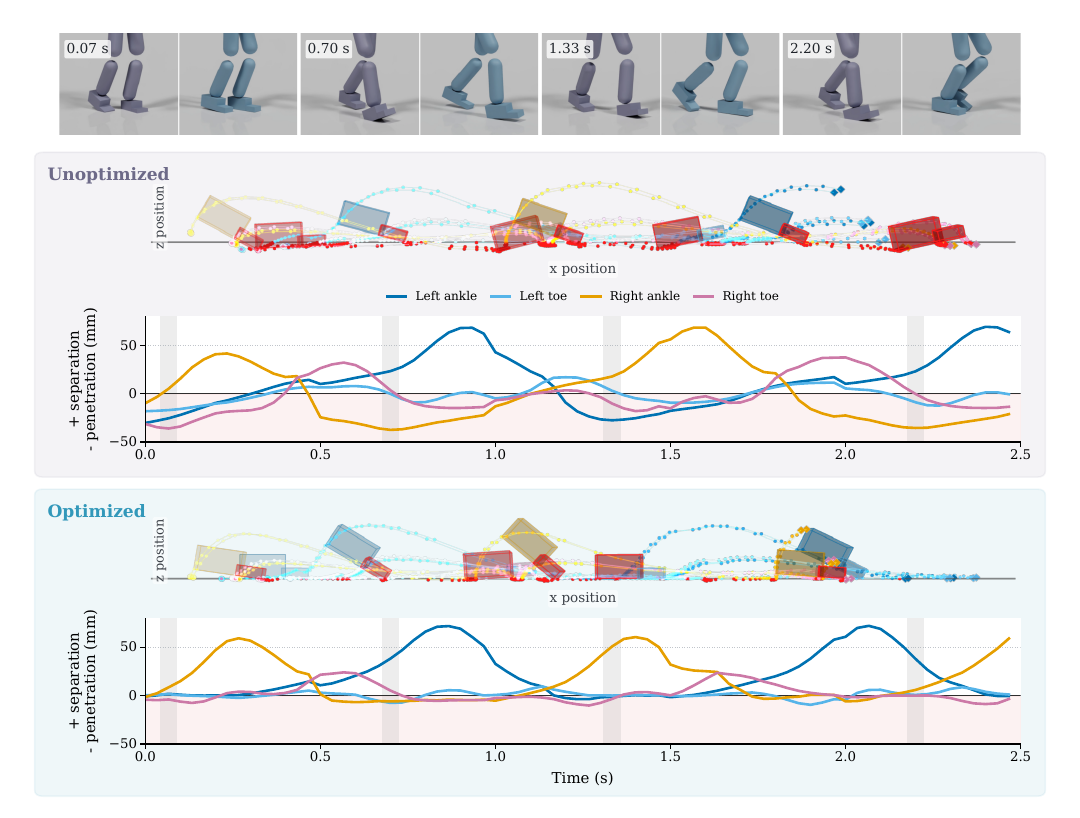}
    \caption{
        Ground-contact behavior for the SMPL humanoid during the \emph{cat} motion. The top row shows rendered snapshots at time instants with visible foot-ground interaction for the unoptimized gray-purple character and optimized cyan-blue character. The lower panels compare unoptimized and optimized motion with matching side-view contact trajectories in the $x$-$z$ plane and grouped foot-body ground distances for the ankles and toes. Positive values indicate separation above the ground plane, while negative values indicate penetration below the ground plane. The optimized motion significantly reduces foot-ground penetration while preserving the overall motion style.
    }
    \label{fig:groundpen-cat}
\end{figure}

Figure~\ref{fig:groundpen-cat} compares unoptimized and Physible-optimized \emph{cat} motion on the \ac{SMPL} humanoid. The matched snapshots and foot trajectories show reduced ground penetration while retaining the characteristic poses. The ground-distance traces connect the visible foot-ground interaction to the remaining penetration events.

\clearpage
\subsubsection{Lawn-Mower Motion: Self-Collision Avoidance}
\label{apx:ssub:kinematic_lawnmower}
We evaluate self-collision avoidance on a $150$-frame \emph{lawn-mower} sequence of the \ac{SMPL} humanoid, whose unoptimized motion exhibits arm interpenetration. In this targeted experiment, Physible optimizes only the self-collision constraint for $100$ iterations. The experiment settings are given in Appendix~\ref{apx:par:kinematic_case_settings}.

\begin{figure}[!htbp]
    \centering
    \includegraphics[width=\columnwidth]{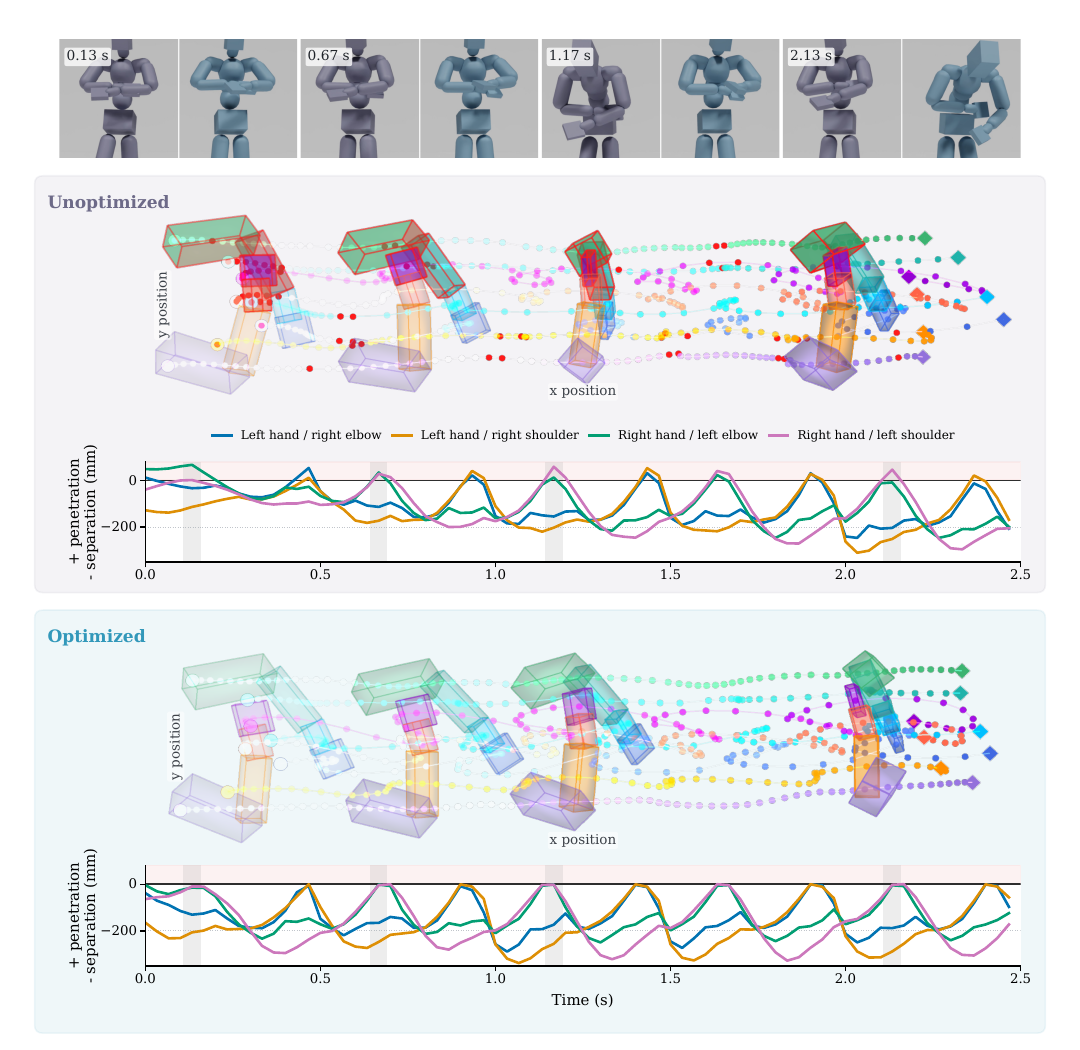}
    \caption{
        Arm self-collision behavior for the SMPL humanoid during a \emph{lawn-mower} motion. The top row shows rendered snapshots at collision-relevant time instants for the unoptimized gray-purple character and optimized cyan-blue character. The lower panels compare unoptimized and optimized motion with matching top-down arm trajectory views and selected hand-arm collision distances. Red-shaded and red-bordered collision geometries and red trajectory markers emphasize time steps for which collisions for the respective bodies take place. Positive signed distances denote penetration, while negative values denote separation. The optimized motion reduces arm interpenetration while preserving the overall motion style.
    }
    \label{fig:selfcol-lawnmower}
\end{figure}

\begin{figure}[h!]
    \centering
    \includegraphics[page=1, width=1.0\linewidth, trim={5, 5, 5, 5}]{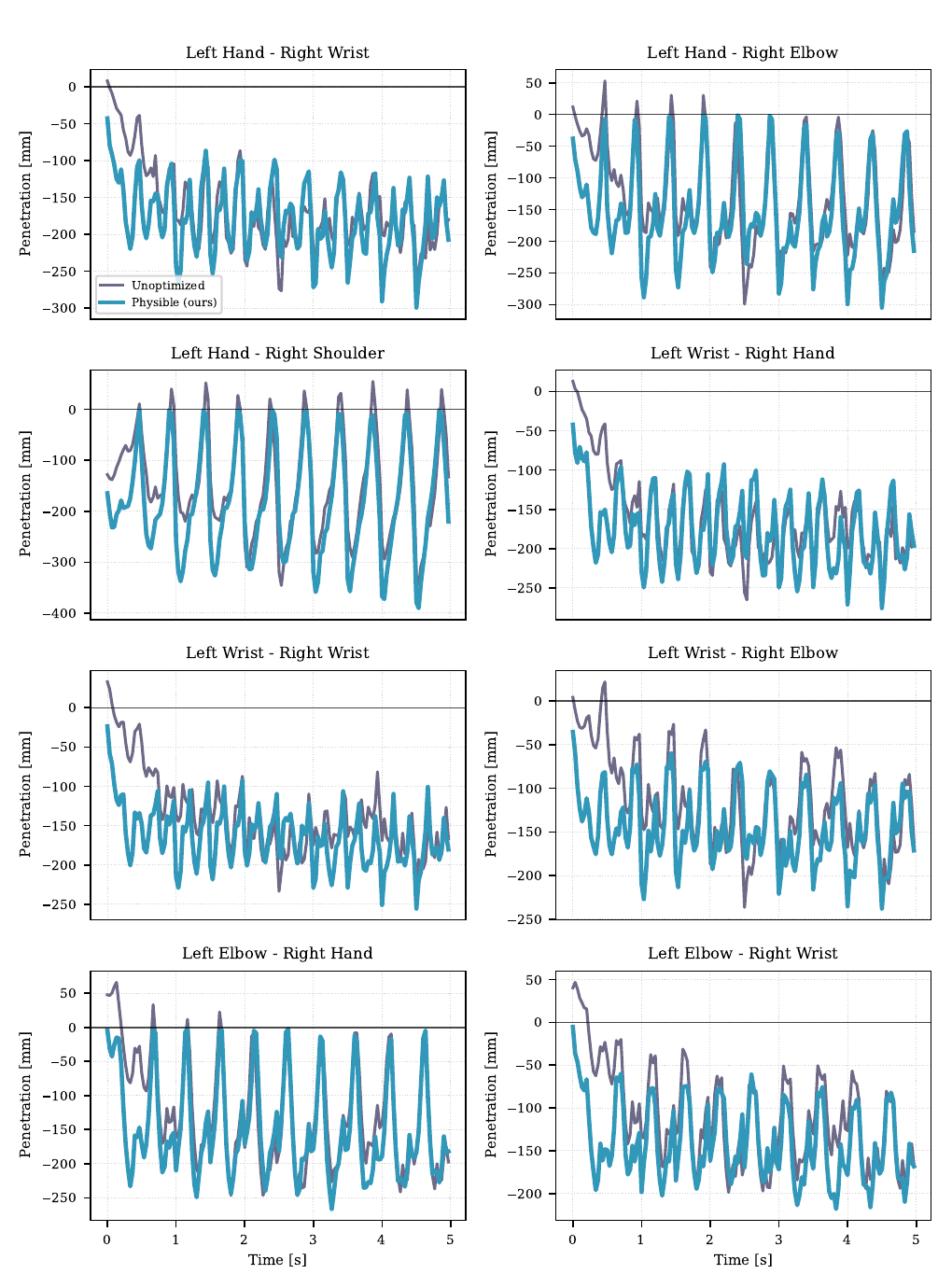}
    \caption{Part 1 of the comparison of the unoptimized vs. optimized self-collision penetration distance when used as a constraint during the kinematic feasibility optimization for $150$ time steps of the \emph{lawn mower} walking style on the SMPL humanoid character. The penetration/separation distances are determined via an \ac{SAT} algorithm applied to pre-defined collision pairs for bodies approximated as cuboids along separate leaves of the kinematic tree.}
    \label{fig:kinoptim_results:selfcol_comparison:smpl_lawnmower:penetration_distance:part1}
\end{figure}

\begin{figure}[h!]
    \centering
    \includegraphics[page=2, width=1.0\linewidth, trim={5, 5, 5, 5}]{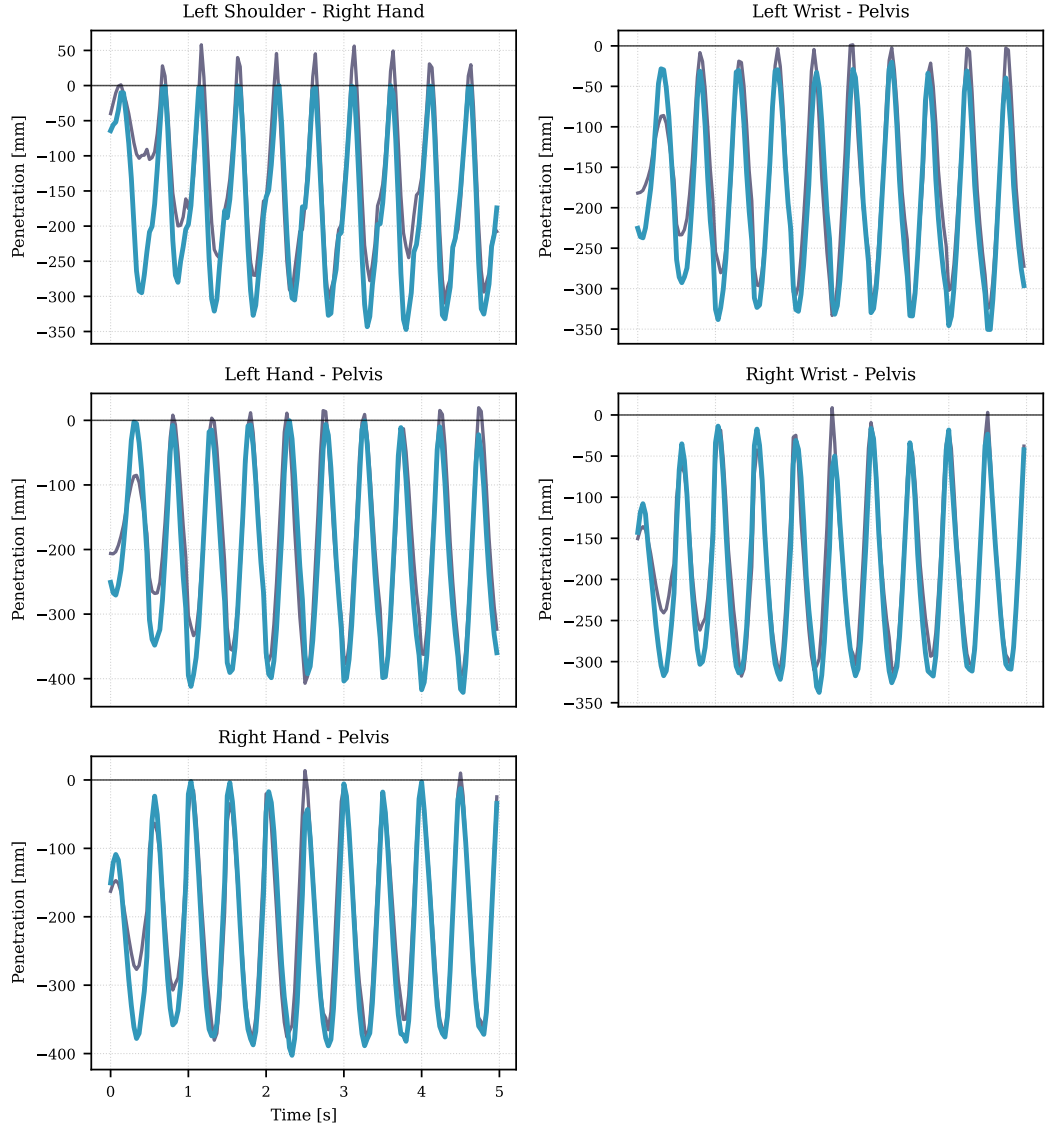}
    \caption{Part 2 of the comparison of the unoptimized vs. optimized self-collision penetration distance when used as a constraint during the kinematic feasibility optimization for $150$ time steps of the \emph{lawn mower} walking style on the SMPL humanoid character. The penetration/separation distances are determined via an \ac{SAT} algorithm applied to pre-defined collision pairs for bodies approximated as cuboids along separate leaves of the kinematic tree.}
    \label{fig:kinoptim_results:selfcol_comparison:smpl_lawnmower:penetration_distance:part2}
\end{figure}

The rendered snapshots and arm trajectories in \figref{fig:selfcol-lawnmower} show how optimization separates the arms at collision-relevant times while retaining the characteristic lawn-mower motion. The corresponding collision-pair traces in Figs.~\ref{fig:kinoptim_results:selfcol_comparison:smpl_lawnmower:penetration_distance:part1} and~\ref{fig:kinoptim_results:selfcol_comparison:smpl_lawnmower:penetration_distance:part2} show that penetration is reduced to negligible levels for the evaluated body pairs. Together, these results demonstrate effective self-collision avoidance in this sequence while preserving the overall arm-motion style.

\clearpage
\subsubsection{Style Preservation across Selected Motions}
\label{apx:ssub:kinematic_style_preservation}
\begin{figure}[!htbp]
    \centering
    \includegraphics[width=\linewidth]{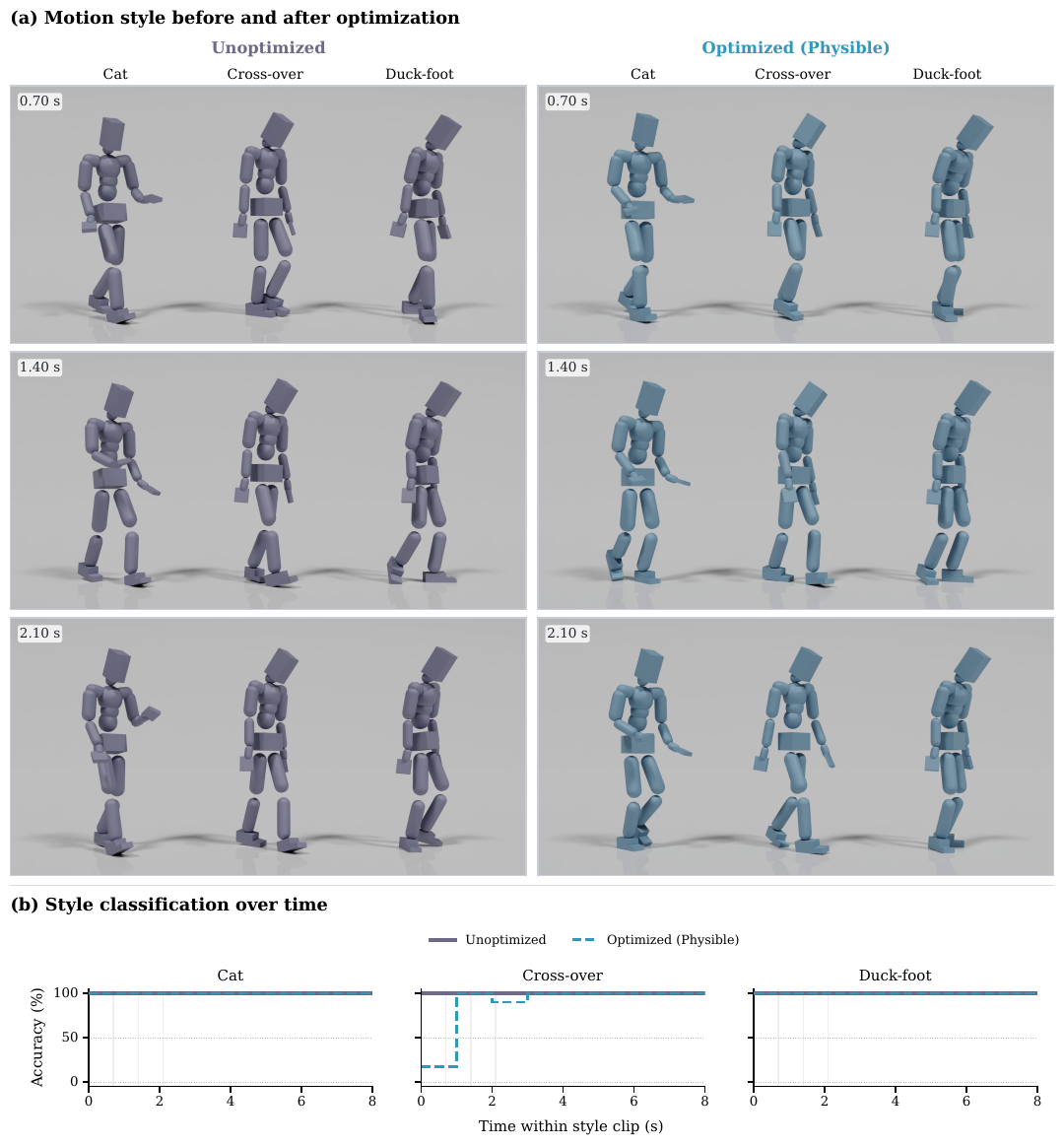}
    \caption{
        Motion-style preservation after kinematic feasibility optimization with Physible on the \ac{SMPL} humanoid.
        \textbf{(a)} Selected, matched-time snapshots of the \emph{cat}, \emph{cross-over}, and \emph{duck-foot} styles at \SI{0.70}{s}, \SI{1.40}{s}, and \SI{2.10}{s}, comparing unoptimized motion (left) with Physible (right).
        \textbf{(b)} Style classification accuracy over each full \SI{8}{s} clip, computed from $15$-frame sliding windows and averaged in \SI{1}{s} bins; vertical markers indicate snapshot times. Cat and duck-foot retain $100\%$ accuracy, while cross-over exhibits a temporary decrease after optimization.
    }
    \label{fig:smpl-style-preservation}
\end{figure}

Figure~\ref{fig:smpl-style-preservation} illustrates style preservation for \emph{cat}, \emph{cross-over}, and \emph{duck-foot}. Matched-time snapshots compare characteristic poses, while sliding-window classification over each full \SI{8}{s} clip shows how style preservation varies over time. These examples complement the aggregate style accuracy and FID results in \tabref{tab:kinoptim:smpl_humanoid}.

\clearpage
\section{Supplementary Dynamic Feasibility Results}
\label{apx:sec:dynamic_feasibility_optimization_results}
This appendix compares methods, evaluates results across motion styles and contact conditions, and measures variability across optimization seeds. It then analyzes selected motions and downstream RL tracking. Experimental settings and metric definitions are given in Appendix~\ref{apx:sec:experimental_protocol}.

\subsection{Quantitative Evaluation}
\label{apx:sub:dynamic_quantitative_results}
\subsubsection{Comparison with Baselines for Two-Foot-Jump}
\label{apx:ssub:dynamic_baseline_results}
The quantitative results in \tabref{tab:dynoptim:bipedal_robot_two_foot_jump_dynoptim} show that most optimization methods reduce the dynamical infeasibility metric relative to the unoptimized motion, but with different trade-offs. \emph{Noise GD} obtains the lowest acceleration-reconstruction error and \ac{ID} residual ($0.013$ and $1.94$, respectively), but requires by far the longest wall-clock time per horizon and reduces \ac{LDJ} from $-7.71$ to $-10.03$. Direct \emph{Motion Sep-CMA-ES} also strongly reduces the two dynamical metrics, but has the lowest \ac{LDJ} ($-11.67$) among the compared methods. In contrast, Physible reduces acceleration reconstruction from $0.454$ to $0.081$ and the \ac{ID} residual from $84.4$ to $26.5$, while attaining the highest \ac{LDJ} among the optimized methods ($-8.51$). \emph{Guided Diffusion} is comparatively fast in total horizon time, but fails in this setting, increasing both dynamical errors and reducing \ac{LDJ} to $-9.91$. Physible trades larger remaining residuals for smoother motion and lower computational cost than \emph{Noise GD}. Appendix~\ref{apx:ssub:dynamic_twofootjump} examines the associated changes in motion style and contact behavior.

\subsubsection{Results across Motion Styles and Contact Conditions}
\label{apx:sub:dynamic_sequence_results}
Table~\ref{tab:dynoptim:sequence_comparison} compares Physible with unoptimized generation on 11 bipedal robot sequences: six flat-ground motion styles, neutral walking on uphill and downhill slopes of $25^\circ$ with $\mu=1.0$, and neutral walking on flat ground with $\mu\in\{0.50,0.10,0.05\}$. Each condition contributes one sequence per method. The two-foot jump spans \SI{5}{s}; all other sequences span \SI{3}{s}. We report temporal means and percentiles of the metrics defined in Appendix~\ref{apx:sub:metrics}; these characterize variation within each sequence. For each dynamical metric, we summarize the relative improvement as the median of $100(1-\bar E^{\mathrm{Physible}}/\bar E^{\mathrm{Unoptimized}})$ across the 11 sequences, where $\bar E$ denotes the sequence mean. This gives each sequence equal weight regardless of duration.

Physible lowers both mean dynamical residuals and their 90th percentiles in every evaluated sequence. The median reductions in mean acceleration-reconstruction error and mean \ac{ID} residual are approximately 89.5\% and 79.5\%, respectively. Improvements vary by motion: left hop and kick reduce both means by about 98\%, whereas bouncy right reduces acceleration-reconstruction error by 68\% and the \ac{ID} residual by 26\%. The slope results show larger reductions downhill (92\% and 87\%) than uphill (71\% and 57\%). Under reduced friction, Physible continues to improve both metrics, although the remaining mean \ac{ID} residual rises from $7.22$ at $\mu=0.50$ to $17.39$ at $\mu=0.10$ and $30.63$ at $\mu=0.05$. These results support improved dynamic consistency within the evaluated robot model across the tested styles and contact conditions, while leaving appreciable infeasibility in some sequences.

Evaluated over 25-frame retained windows for all sequences, mean \ac{LDJ} is lower after optimization in every sequence, indicating a smoothness tradeoff. The largest decrease occurs at $\mu=0.50$, from $-7.63$ to $-8.97$.

\begin{table*}[!htbp]
\centering
\caption{Dynamical feasibility across bipedal-robot sequences. Each sequence is
reported separately for Unoptimized and Physible. Entries are mean
[p90] for acceleration reconstruction and inverse-dynamics residual, and mean
[p10] for LDJ. Bold marks the better mean within each sequence and metric.}
\label{tab:dynoptim:sequence_comparison}
\begingroup
\setlength{\tabcolsep}{2.7pt}
\renewcommand{\arraystretch}{1.18}
\fontsize{8}{9.6}\selectfont
\begin{tabular}{@{}lr rrrr rr@{}}
\toprule
Sequence & \shortstack{Frames /\\Duration [s]} &
\multicolumn{2}{c}{\shortstack{Acceleration\\reconstruction $\downarrow$}} &
\multicolumn{2}{c}{\shortstack{Inverse-dynamics\\residual $\downarrow$}} &
\multicolumn{2}{c}{LDJ $\uparrow$} \\
\cmidrule(lr){3-4}\cmidrule(lr){5-6}\cmidrule(l){7-8}
& & Unoptimized & \textcolor[HTML]{237A96}{Physible} & Unoptimized & \textcolor[HTML]{237A96}{Physible} & Unoptimized & \textcolor[HTML]{237A96}{Physible} \\
\midrule
\multicolumn{8}{l}{\textit{Flat-ground motion styles}} \\
\addlinespace[2pt]
Two-foot jump & 250 / 5 & 0.454 [0.954] & \textbf{0.081 [0.174]} & 84.40 [186.93] & \textbf{26.48 [62.70]} & \textbf{-7.71 [-8.11]} & -8.51 [-8.83] \\
Left hop & 150 / 3 & 0.511 [0.721] & \textbf{0.009 [0.014]} & 106.43 [167.33] & \textbf{1.91 [3.19]} & \textbf{-8.57 [-8.96]} & -8.83 [-9.42] \\
Bouncy right & 150 / 3 & 0.505 [1.014] & \textbf{0.161 [0.271]} & 74.11 [147.12] & \textbf{54.61 [95.67]} & \textbf{-7.59 [-7.80]} & -8.27 [-8.56] \\
Skip & 150 / 3 & 0.800 [0.927] & \textbf{0.059 [0.175]} & 93.73 [166.87] & \textbf{19.20 [64.97]} & \textbf{-7.69 [-7.95]} & -8.31 [-8.73] \\
Kick & 150 / 3 & 0.893 [1.450] & \textbf{0.011 [0.021]} & 116.91 [187.61] & \textbf{2.70 [6.05]} & \textbf{-7.92 [-8.75]} & -8.31 [-8.62] \\
Morris & 150 / 3 & 0.438 [0.779] & \textbf{0.046 [0.149]} & 78.25 [157.96] & \textbf{15.26 [56.44]} & \textbf{-7.93 [-8.41]} & -8.50 [-8.92] \\
\midrule
\multicolumn{8}{l}{\textit{Terrain transfer: neutral style, $\mu=1.0$}} \\
\addlinespace[2pt]
Uphill $25^{\circ}$ & 150 / 3 & 0.451 [0.952] & \textbf{0.131 [0.275]} & 89.31 [133.85] & \textbf{38.65 [74.45]} & \textbf{-7.79 [-8.13]} & -8.29 [-8.71] \\
Downhill $25^{\circ}$ & 150 / 3 & 0.387 [0.987] & \textbf{0.032 [0.091]} & 66.04 [138.51] & \textbf{8.58 [24.20]} & \textbf{-7.60 [-7.84]} & -8.11 [-8.40] \\
\midrule
\multicolumn{8}{l}{\textit{Friction variation: neutral style, flat ground}} \\
\addlinespace[2pt]
$\mu = 0.50$ & 150 / 3 & 0.330 [0.837] & \textbf{0.025 [0.058]} & 49.98 [132.39] & \textbf{7.22 [15.54]} & \textbf{-7.63 [-7.77]} & -8.97 [-10.16] \\
$\mu = 0.10$ & 150 / 3 & 0.487 [0.932] & \textbf{0.054 [0.161]} & 57.82 [140.16] & \textbf{17.39 [62.26]} & \textbf{-7.69 [-7.97]} & -8.18 [-8.49] \\
$\mu = 0.05$ & 150 / 3 & 0.601 [0.935] & \textbf{0.098 [0.299]} & 70.10 [143.81] & \textbf{30.63 [109.14]} & \textbf{-7.63 [-7.96]} & -8.34 [-8.94] \\
\bottomrule
\end{tabular}
\endgroup
\end{table*}

\clearpage
\subsection{Convergence and Variability}
\label{apx:sub:dynamic_convergence_variability}
\subsubsection{Variability across Optimization Seeds}
\label{apx:ssub:dynamic_seed_variability}
We evaluate the sensitivity of Physible to the random seed during dynamic feasibility optimization using ten seeds ($0$--$9$) for the two-foot-jump style on the bipedal robot. Each run generates $150$ frames (\SI{3}{s} at \SI{50}{Hz}) under the fixed command $(v_x^{\mathrm{cmd}},v_y^{\mathrm{cmd}},\omega_z^{\mathrm{cmd}})=(\SI{1}{\meter\per\second},\SI{0}{\meter\per\second},\SI{0}{\radian\per\second})$. We use \ac{Sep-CMA-ES} with $K=2500$ iterations per horizon and $N_{\mathrm{pop}}=64$, keeping the commands and optimization configuration identical across runs. Table~\ref{tab:dynoptim:bipedal_robot:twofootjump:seeds} reports motion differences averaged over all $45$ seed pairs and mean $\pm$ sample standard deviation across the ten sequence scores. Each seed contributes one temporal mean for each evaluation metric, following Appendix~\ref{apx:sub:metrics}.

The dynamic feasibility metrics show moderate variability across seeds: acceleration reconstruction is $0.0877\pm0.00979$ and the \ac{ID} residual is $28.82\pm3.38$, corresponding to relative standard deviations of approximately $11.2\%$ and $11.7\%$. Motion smoothness is consistent, with \ac{LDJ} of $-9.06\pm0.12$. Average pairwise root-trajectory \ac{RMSE} and body-position distance are \SI{0.0757}{m} and \SI{0.0169}{m}, respectively, while root-orientation and joint-angle differences are \SI{0.0438}{rad} and \SI{0.138}{rad}. Together, these results support consistent optimization quality and motion smoothness across the tested seeds for this dynamic task.

\begin{table}[htbp]
\centering
\caption{Optimizer-seed sensitivity of Physible for two-foot-jump dynamic feasibility optimization on the bipedal robot: ten seeds and $150$ frames (\SI{3}{s}) per seed, with identical velocity commands. Motion rows report average pairwise differences; evaluation metrics report mean $\pm$ sample standard deviation across seed-specific temporal means. Alignment and metric definitions are given in Appendix~\ref{apx:sub:metrics}.}
\label{tab:dynoptim:bipedal_robot:twofootjump:seeds}
\begingroup
\setlength{\tabcolsep}{7pt}
\renewcommand{\arraystretch}{1.1}
\small
\begin{tabular}{lr}
\toprule
Metric & Result \\
\midrule
Pairwise root trajectory RMSE [m] & $7.57\times10^{-2}$ \\
Pairwise root orientation difference [rad] & $4.38\times10^{-2}$ \\
Pairwise joint angle difference [rad] & $1.38\times10^{-1}$ \\
Pairwise body position distance [m] & $1.69\times10^{-2}$ \\
\midrule
Acceleration reconstruction & $8.77\times10^{-2}\pm9.79\times10^{-3}$ \\
Inverse-dynamics residual & $28.82\pm3.38$ \\
LDJ & $-9.06\pm0.12$ \\
\bottomrule
\end{tabular}
\endgroup
\end{table}

\clearpage
\subsection{Motion Case Studies for Qualitative Analysis}
\label{apx:sub:dynamic_case_studies}
The following cases pair motion behavior with time-resolved diagnostics to allow for a qualitative analysis of the optimized motion. Acceleration, torque, and friction diagnostics use the nominal inverse-dynamics solution. Signed acceleration residuals are plotted as planned minus reconstructed acceleration, corresponding to $-\mathbf e_{\mathrm{acc}}$. Virtual-wrench diagnostics use the separate relaxed solve in \eqnref{eq:relaxed_id_socp}. Across the method comparisons, lower dynamic residuals can coincide with degraded style, reduced smoothness, or loss of ground contact; the diagnostics are therefore interpreted together.

\subsubsection{Two-Foot Jump}
\label{apx:ssub:dynamic_twofootjump}
\begin{figure}[p]
    \centering
    \includegraphics[width=\linewidth]{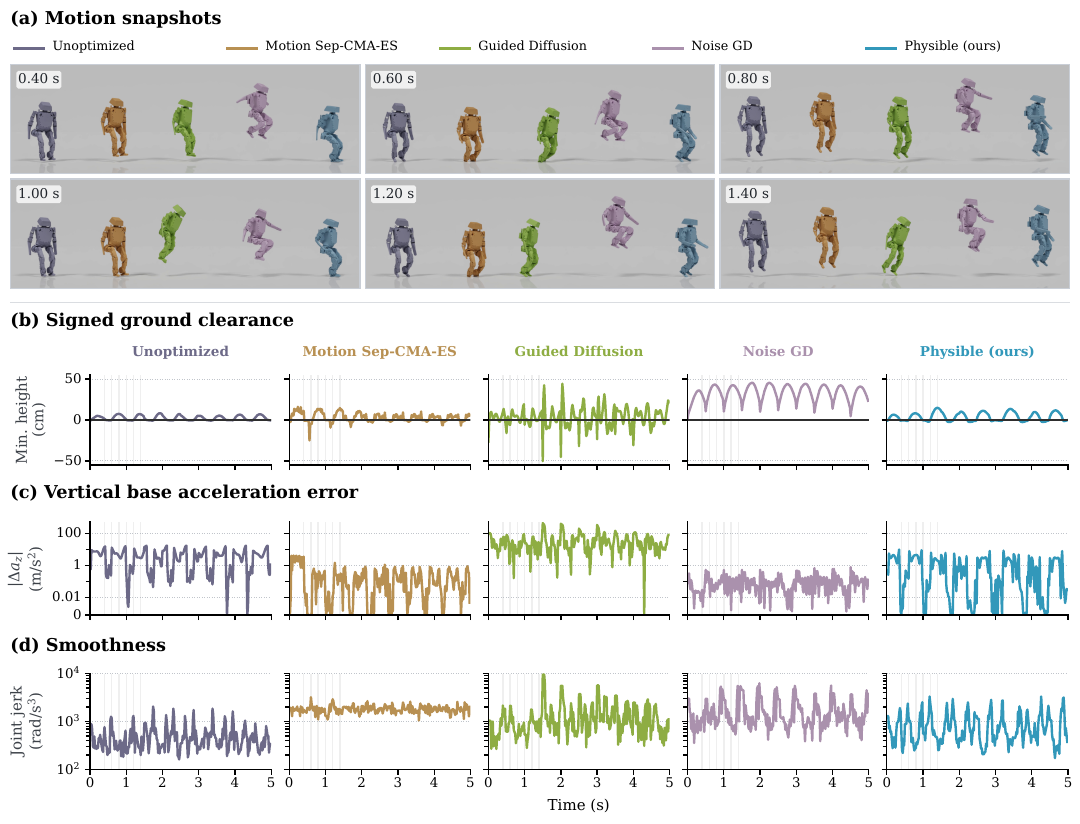}
    \caption{Comparison of dynamic feasibility optimization methods for the bipedal robot's \emph{two-foot-jump} motion: unoptimized generation, Motion Sep-CMA-ES, Guided Diffusion, Noise GD, and Physible. (a) Motion snapshots at the indicated times. The plots show (b) signed ground clearance, with negative values indicating ground penetration; (c) the absolute difference between planned and dynamically feasible vertical base acceleration, $|\Delta a_z|$; and (d) joint jerk over a \SI{5}{s} sequence. Gray vertical bands mark the snapshot times. The comparison illustrates trade-offs between acceleration consistency, ground contact, and motion smoothness: Noise GD has low acceleration error but remains above the ground, whereas Physible retains repeated ground contacts and the characteristic jumping motion.}
    \label{fig:dynoptim-methods}
\end{figure}

\begin{figure}[h!]
    \centering
    \includegraphics[page=1,width=1.0\linewidth, trim={5, 5, 5, 5}]{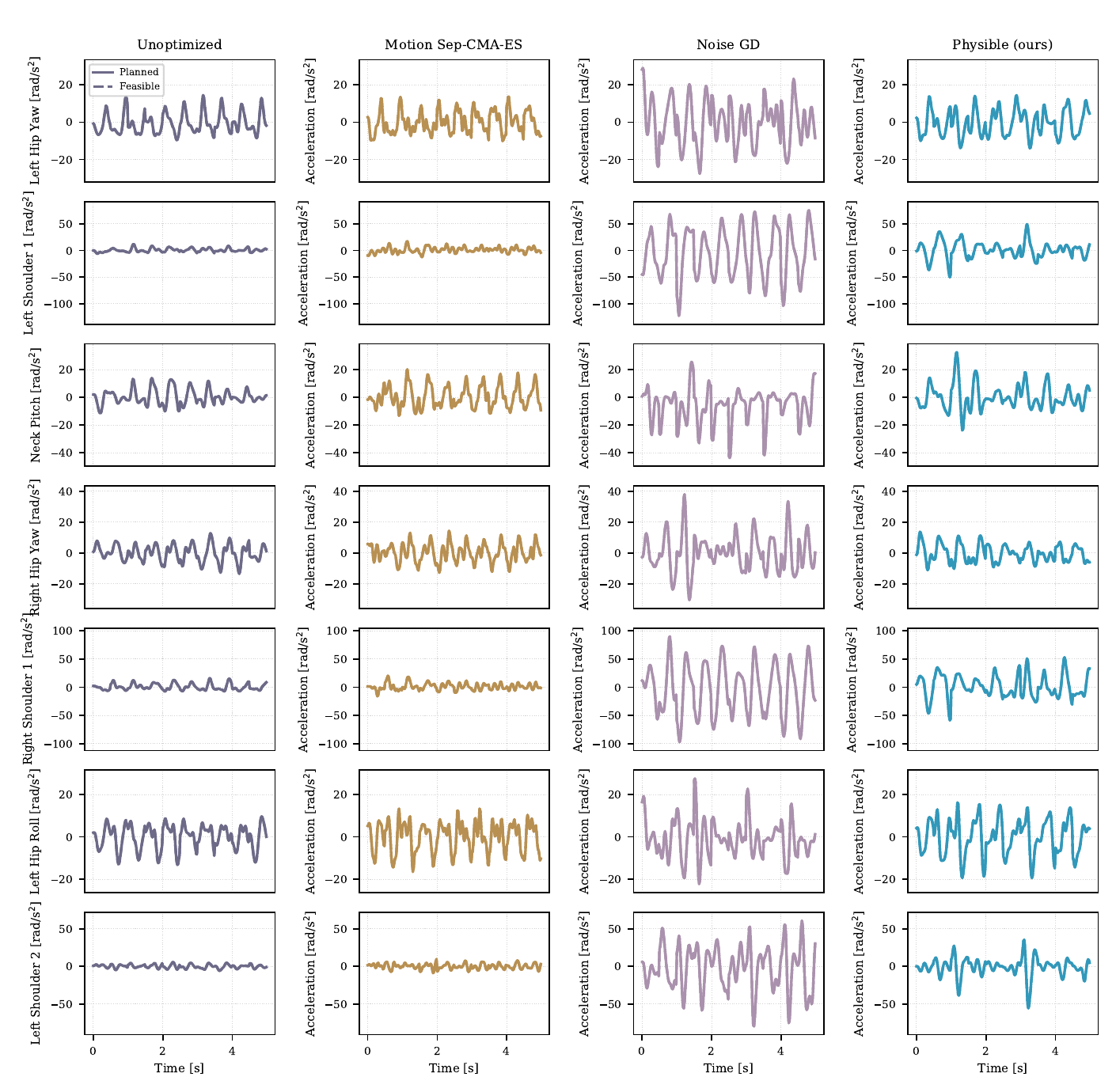}
    \caption{Behavior comparison of the \emph{joint acceleration planned vs. feasible} by method for the dynamic feasibility optimization for $250$ time steps of the \emph{two-foot-jump} motion style on the bipedal robot -- Part 1.}
    \label{fig:dynoptim_results:behavior_comparison:bipedal_robot_twofootjump:joint_acceleration_planned_vs_feasible:part1}
\end{figure}

\begin{figure}[h!]
    \centering
    \includegraphics[page=2,width=1.0\linewidth, trim={5, 5, 5, 5}]{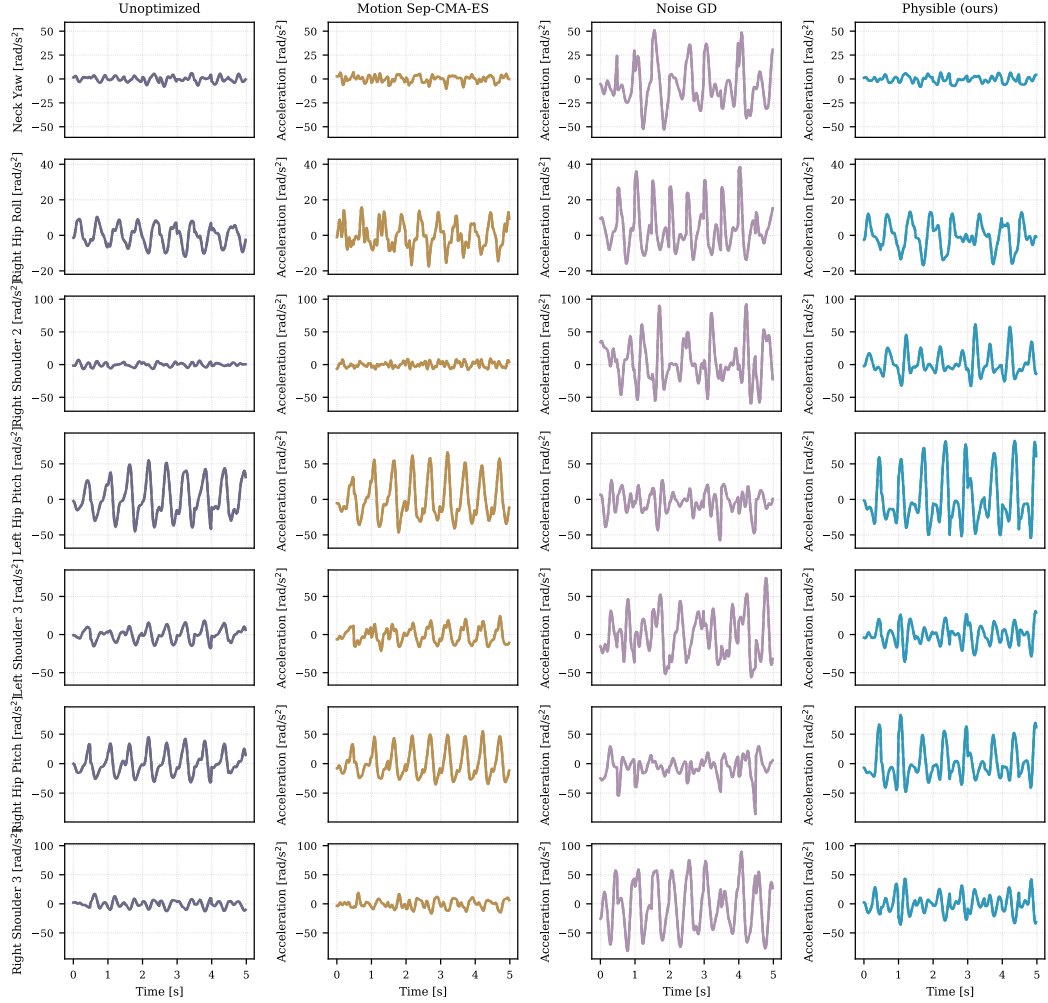}
    \caption{Behavior comparison of the \emph{joint acceleration planned vs. feasible} by method for the dynamic feasibility optimization for $250$ time steps of the \emph{two-foot-jump} motion style on the bipedal robot -- Part 2.}
    \label{fig:dynoptim_results:behavior_comparison:bipedal_robot_twofootjump:joint_acceleration_planned_vs_feasible:part2}
\end{figure}

\begin{figure}[h!]
    \centering
    \includegraphics[page=3,width=1.0\linewidth, trim={5, 5, 5, 5}]{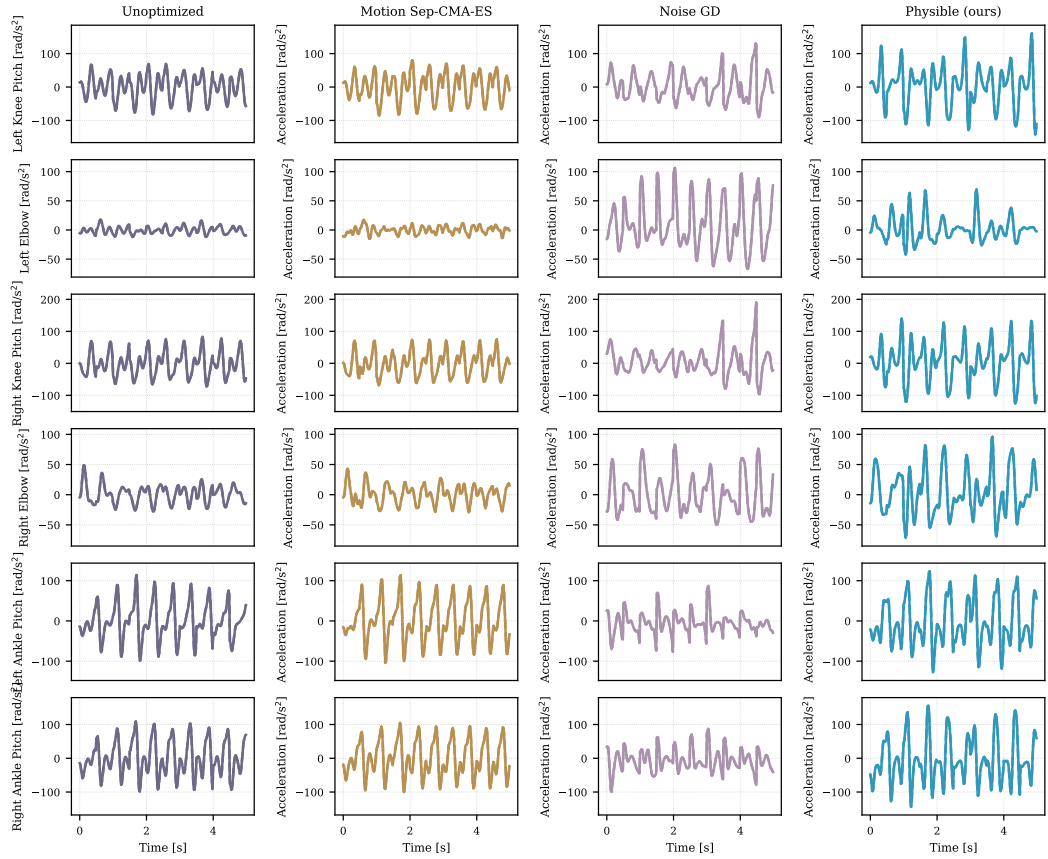}
    \caption{Behavior comparison of the \emph{joint acceleration planned vs. feasible} by method for the dynamic feasibility optimization for $250$ time steps of the \emph{two-foot-jump} motion style on the bipedal robot -- Part 3.}
    \label{fig:dynoptim_results:behavior_comparison:bipedal_robot_twofootjump:joint_acceleration_planned_vs_feasible:part3}
\end{figure}

\begin{figure}[h!]
    \centering
    \includegraphics[page=1,width=1.0\linewidth, trim={5, 5, 5, 5}]{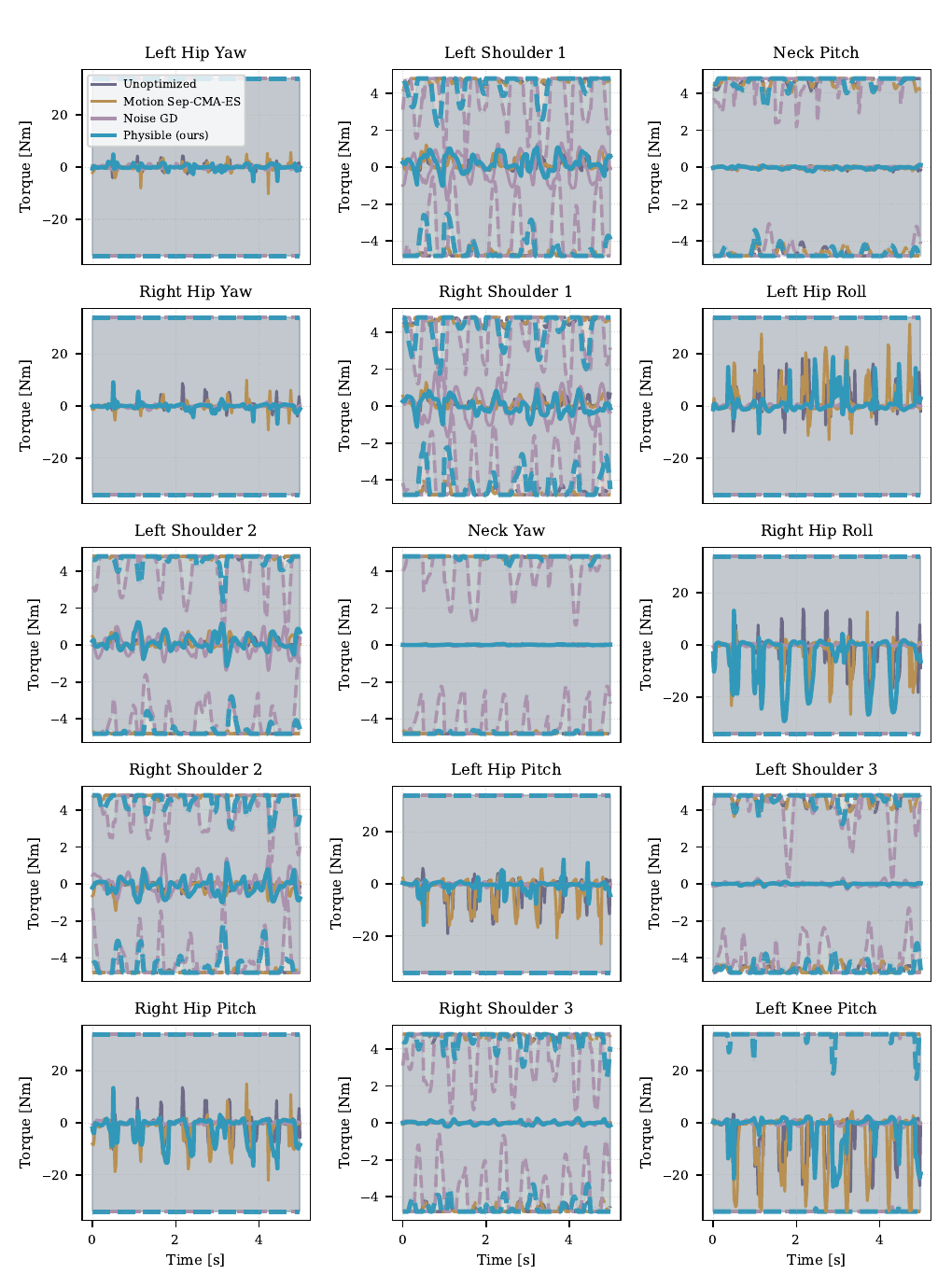}
    \caption{Behavior comparison of the \emph{joint torques (including limits)} for the dynamic feasibility optimization for $250$ time steps of the \emph{two-foot-jump} motion style on the bipedal robot -- Part 1.}
    \label{fig:dynoptim_results:behavior_comparison:bipedal_robot_twofootjump:joint_torques:part1}
\end{figure}

\begin{figure}[h!]
    \centering
    \includegraphics[page=2,width=1.0\linewidth, trim={5, 5, 5, 5}]{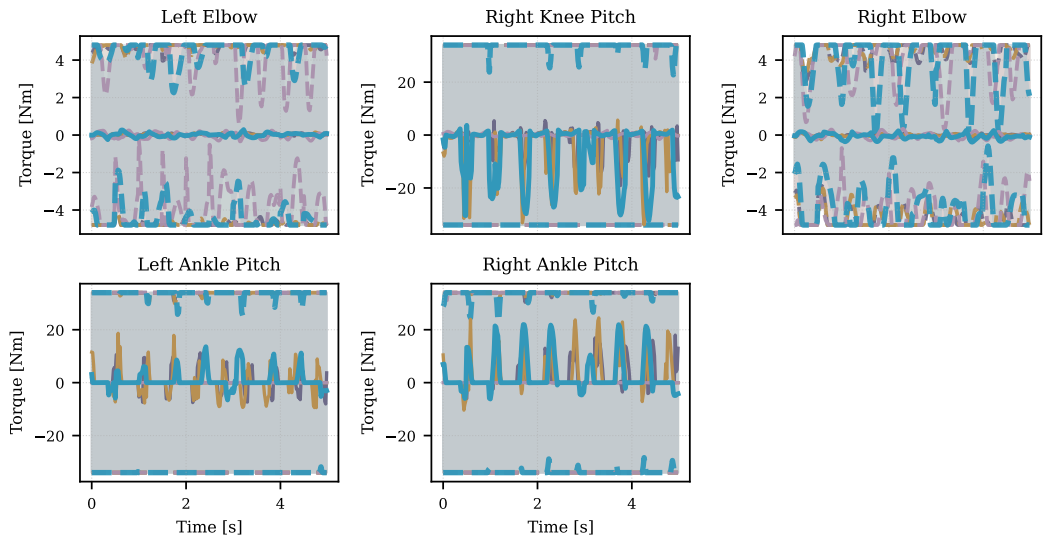}
    \caption{Behavior comparison of the \emph{joint torques (including limits)} for the dynamic feasibility optimization for $250$ time steps of the \emph{two-foot-jump} motion style on the bipedal robot -- Part 2.}
    \label{fig:dynoptim_results:behavior_comparison:bipedal_robot_twofootjump:joint_torques:part2}
\end{figure}

\begin{figure}[h!]
    \centering
    \includegraphics[width=1.0\linewidth, trim={5, 5, 5, 5}]{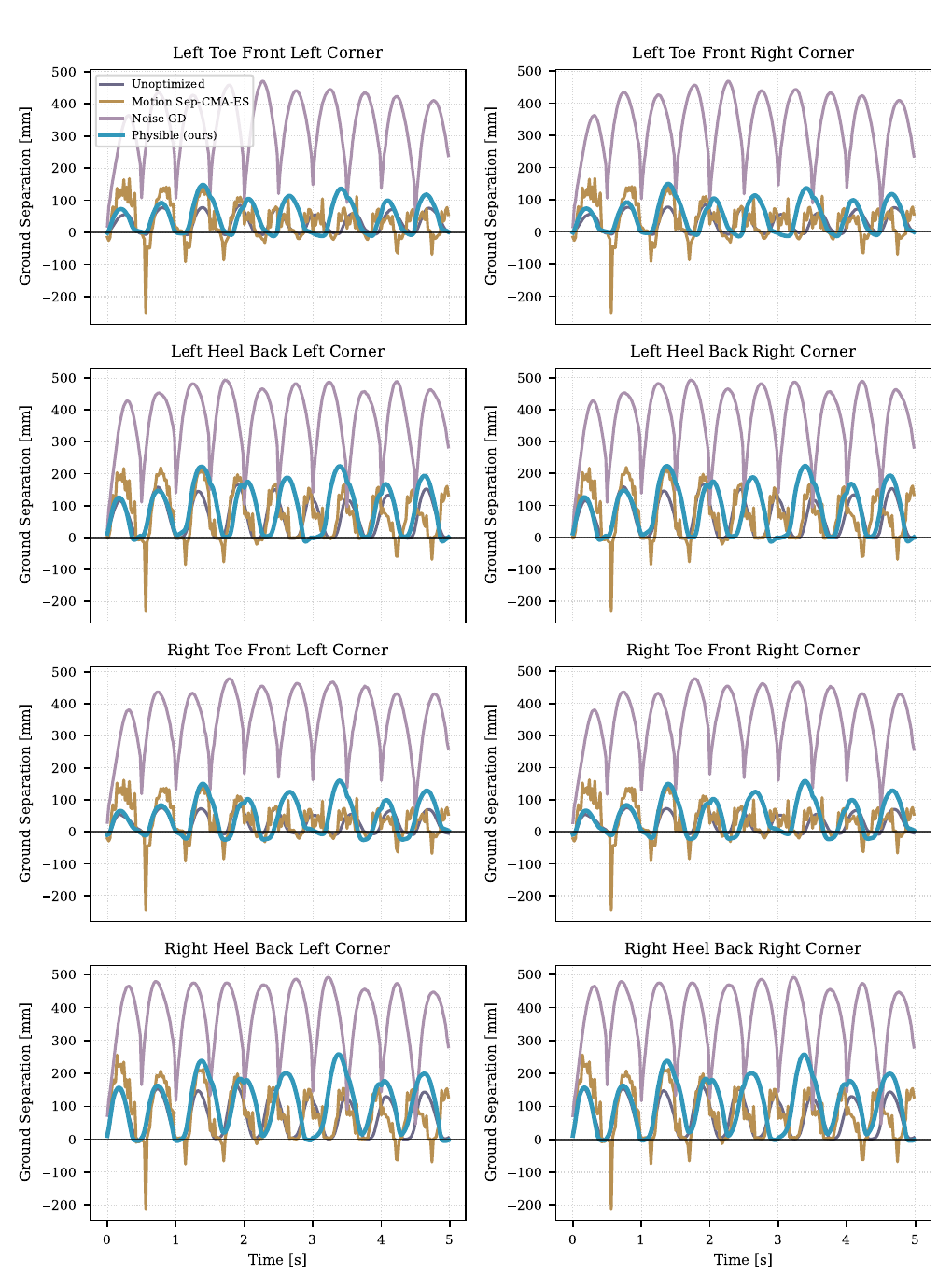}
    \caption{Behavior comparison of the \emph{ground separation} for the dynamic feasibility optimization for $250$ time steps of the \emph{two-foot-jump} motion style on the bipedal robot.}
    \label{fig:dynoptim_results:behavior_comparison:bipedal_robot_twofootjump:ground_separation}
\end{figure}

\begin{figure}[h!]
    \centering
    \includegraphics[width=1.0\linewidth, trim={5, 5, 5, 5}]{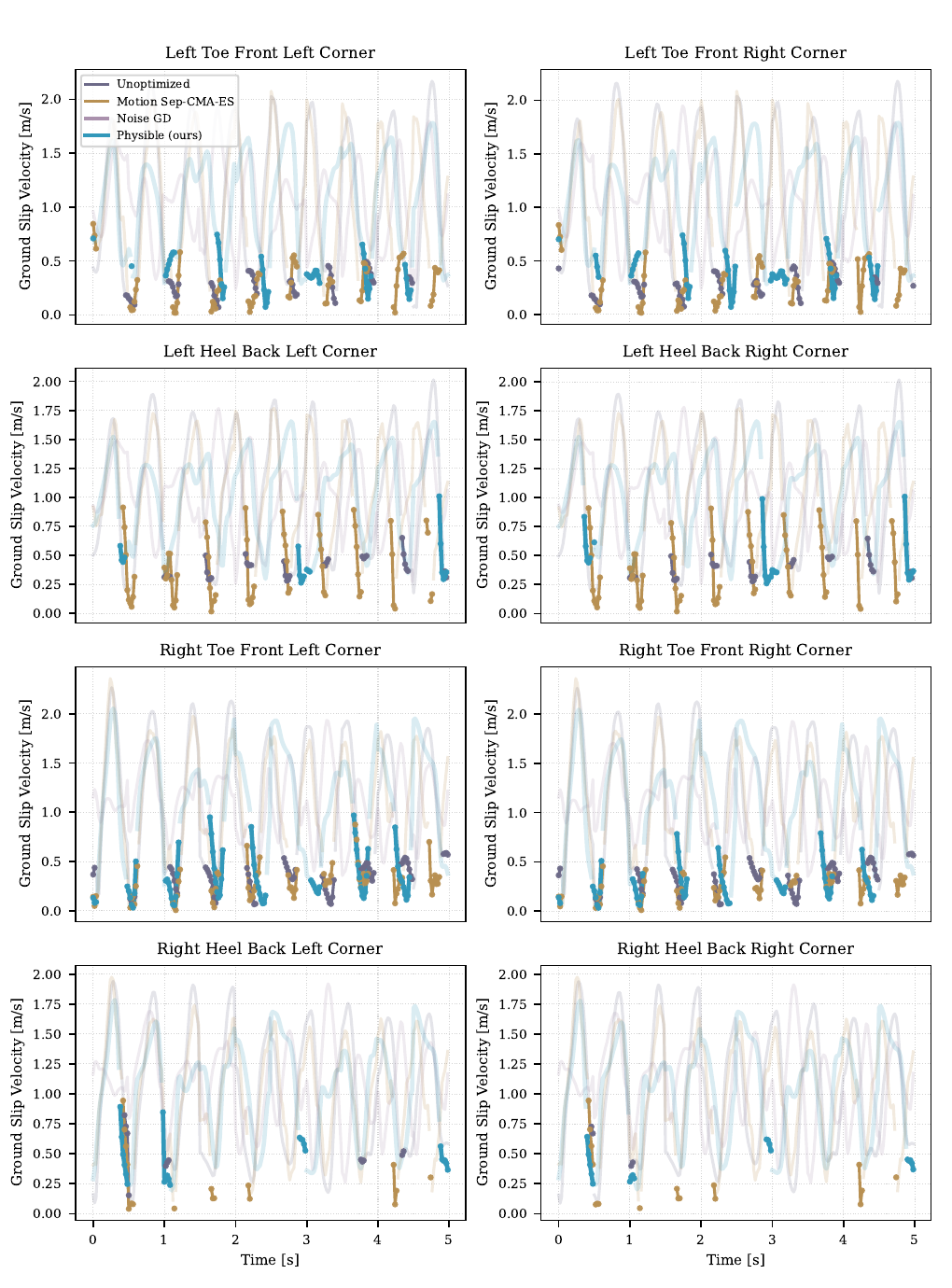}
    \caption{Behavior comparison of the \emph{ground slip velocity} for the dynamic feasibility optimization for $250$ time steps of the \emph{two-foot-jump} motion style on the bipedal robot.}
    \label{fig:dynoptim_results:behavior_comparison:bipedal_robot_twofootjump:ground_slip}
\end{figure}

\paragraph{Comparison against Baselines}
We compare optimization methods on a $250$-frame (\SI{5}{s}) two-foot-jump sequence of the bipedal robot. Figure~\ref{fig:dynoptim-methods} shows unoptimized generation, Motion Sep-CMA-ES, Guided Diffusion, Noise GD, and Physible, corresponding to the methods evaluated in \tabref{tab:dynoptim:bipedal_robot_two_foot_jump_dynoptim}. Motion snapshots and time histories of ground clearance, vertical base acceleration error, and joint jerk illustrate the tradeoffs between dynamic consistency, contact behavior, and motion quality. \emph{Motion Sep-CMA-ES} introduces substantial jitter, reflected in rapidly varying joint accelerations and torques (Figs.~\ref{fig:dynoptim_results:behavior_comparison:bipedal_robot_twofootjump:joint_acceleration_planned_vs_feasible:part1}--\ref{fig:dynoptim_results:behavior_comparison:bipedal_robot_twofootjump:joint_acceleration_planned_vs_feasible:part3} and Figs.~\ref{fig:dynoptim_results:behavior_comparison:bipedal_robot_twofootjump:joint_torques:part1}--\ref{fig:dynoptim_results:behavior_comparison:bipedal_robot_twofootjump:joint_torques:part2}), while \emph{Guided Diffusion} increases vertical base acceleration errors and produces both unrealistic floating and ground penetration. \emph{Noise GD} reduces acceleration error but leaves the character floating throughout the sequence, suggesting that optimization can settle on a physically implausible solution despite achieving a small residual. The ground-separation and slip traces show that its low slip coincides with loss of contact (Figs.~\ref{fig:dynoptim_results:behavior_comparison:bipedal_robot_twofootjump:ground_separation} and~\ref{fig:dynoptim_results:behavior_comparison:bipedal_robot_twofootjump:ground_slip}). \emph{Physible} achieves a more balanced tradeoff: it reduces vertical base acceleration error while retaining the characteristic jumping motion, repeated ground contacts, and comparatively smooth trajectories.

The full-sequence diagnostics compare unoptimized generation, Motion Sep-CMA-ES, Noise GD, and Physible. Guided Diffusion is omitted from these plots because its much larger errors would obscure the differences among the remaining methods.

\begin{figure}[t]
    \centering
    \includegraphics[width=\linewidth]{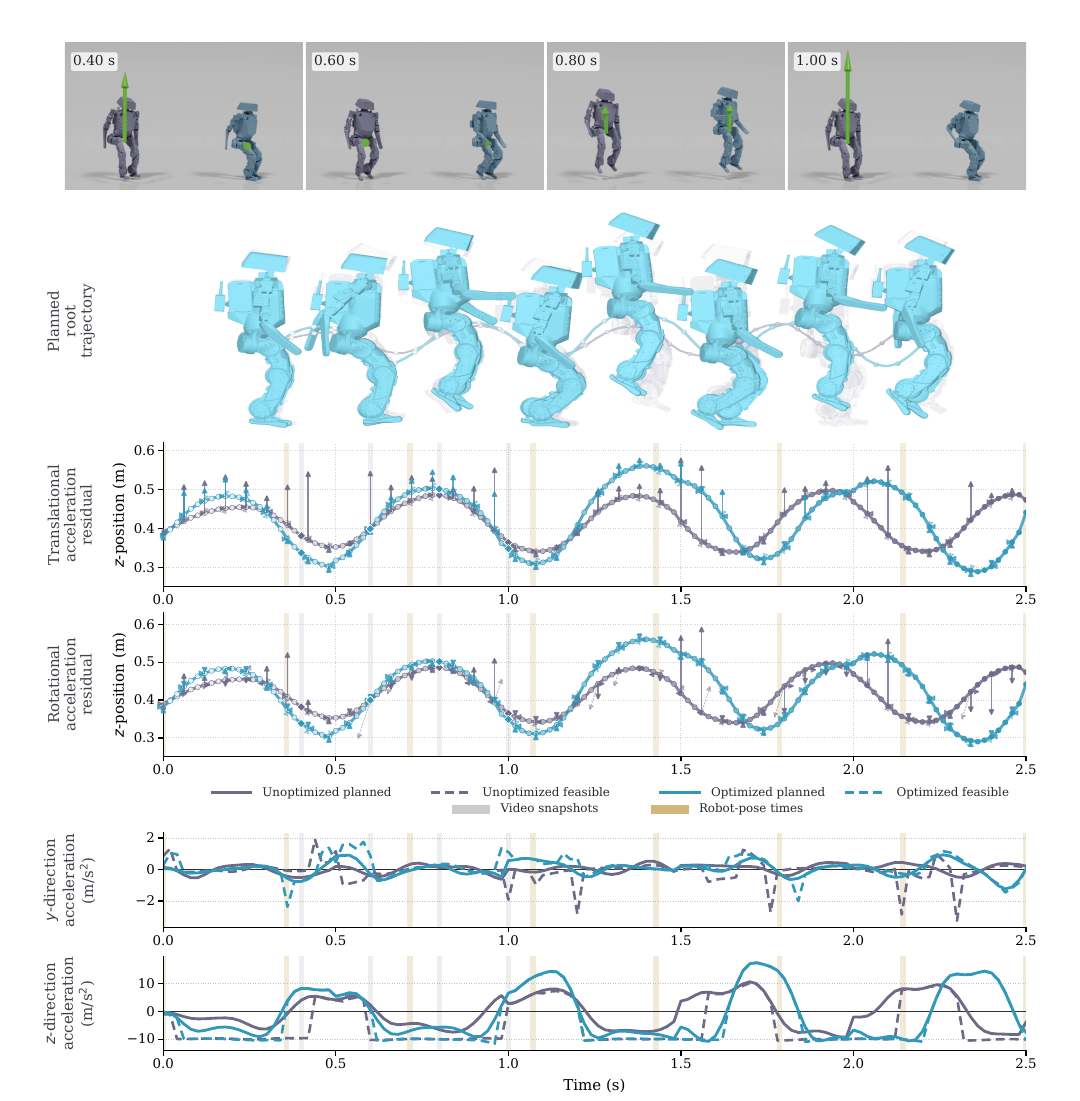}
    \caption{
    Dynamical feasibility optimization of the \emph{two-foot-jump} motion on the bipedal robot over a $2.5\,\mathrm{s}$ horizon.
    The top row shows rendered video snapshots of the unoptimized and optimized motions, including the virtual root force visualization used during dynamical feasibility analysis. The second row shows the planned root trajectories from a side view: the unoptimized motion is rendered transparently in the background, while the optimized motion is rendered in the foreground. Robot configurations are shown at separate robot-pose times distributed across the full horizon, whereas the video snapshots correspond to the selected first-row rendering times. The gray vertical bands in the quantitative plots mark the video snapshot times, and the amber bands mark the robot-pose times used for the second-row configuration renderings. The translational and rotational acceleration residual plots visualize the planned root trajectory together with arrows denoting the residual acceleration, computed as planned minus dynamically feasible acceleration. Smaller residual arrows indicate improved dynamical consistency. The bottom rows compare planned and feasible root accelerations in the $y$- and $z$-directions; solid lines denote planned accelerations and dashed lines denote feasible accelerations, with separate colors for unoptimized and optimized motions.
    }
    \label{fig:dynoptim-twofootjump}
\end{figure}

\begin{figure}[h!]
    \centering
    \includegraphics[width=1.0\linewidth, trim={5, 5, 5, 5}]{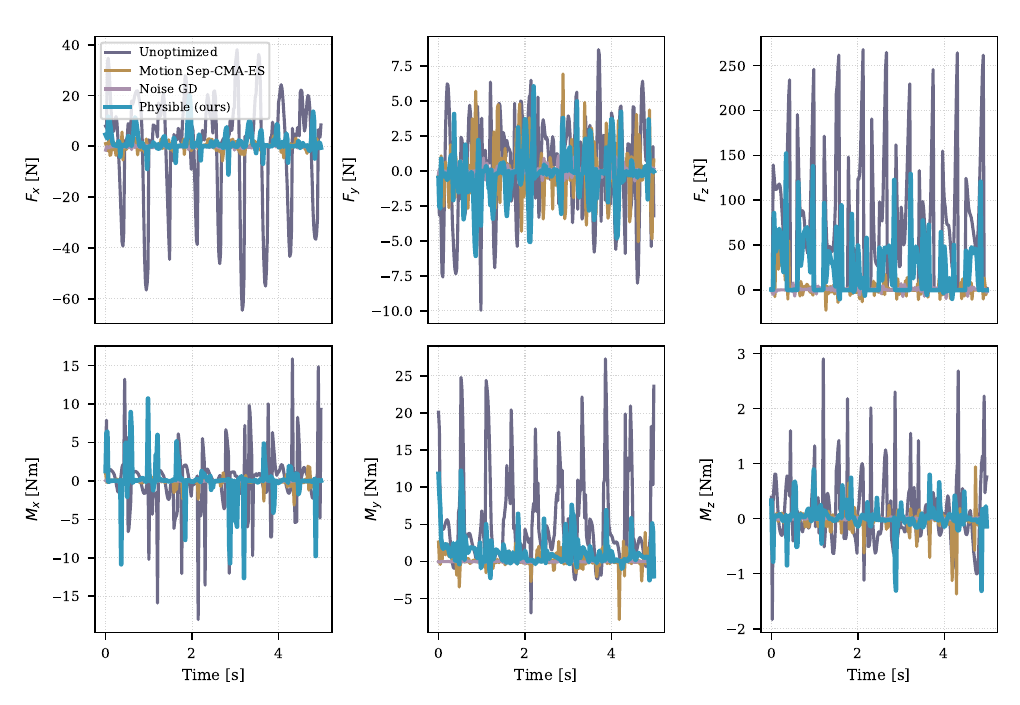}
    \caption{Behavior comparison of the \emph{virtual wrench} resulting from the relaxed \ac{ID} for the dynamic feasibility optimization for $250$ time steps of the \emph{two-foot-jump} motion style on the bipedal robot.}
    \label{fig:dynoptim_results:behavior_comparison:bipedal_robot_twofootjump:virtual_wrench}
\end{figure}

\begin{figure}[h!]
    \centering
    \includegraphics[width=1.0\linewidth, trim={5, 5, 5, 5}]{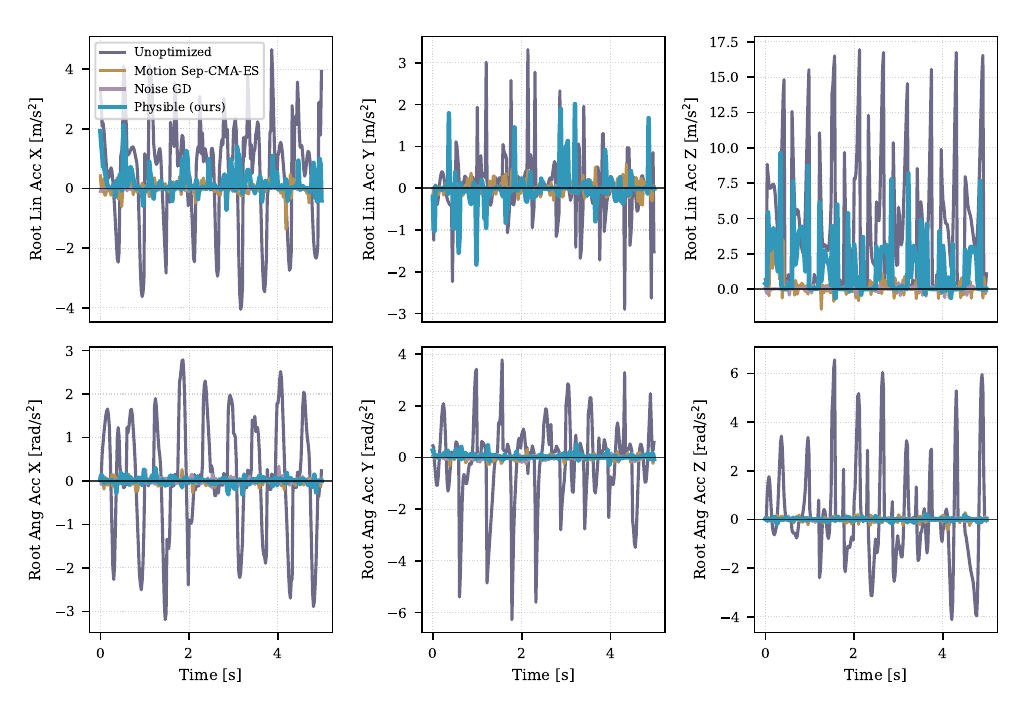}
    \caption{Behavior comparison of the \emph{root acceleration residual} for the dynamic feasibility optimization for $250$ time steps of the \emph{two-foot-jump} motion style on the bipedal robot.}
    \label{fig:dynoptim_results:behavior_comparison:bipedal_robot_twofootjump:root_acceleration_residual}
\end{figure}

\begin{figure}[h!]
    \centering
    \includegraphics[width=1.0\linewidth, trim={5, 5, 5, 5}]{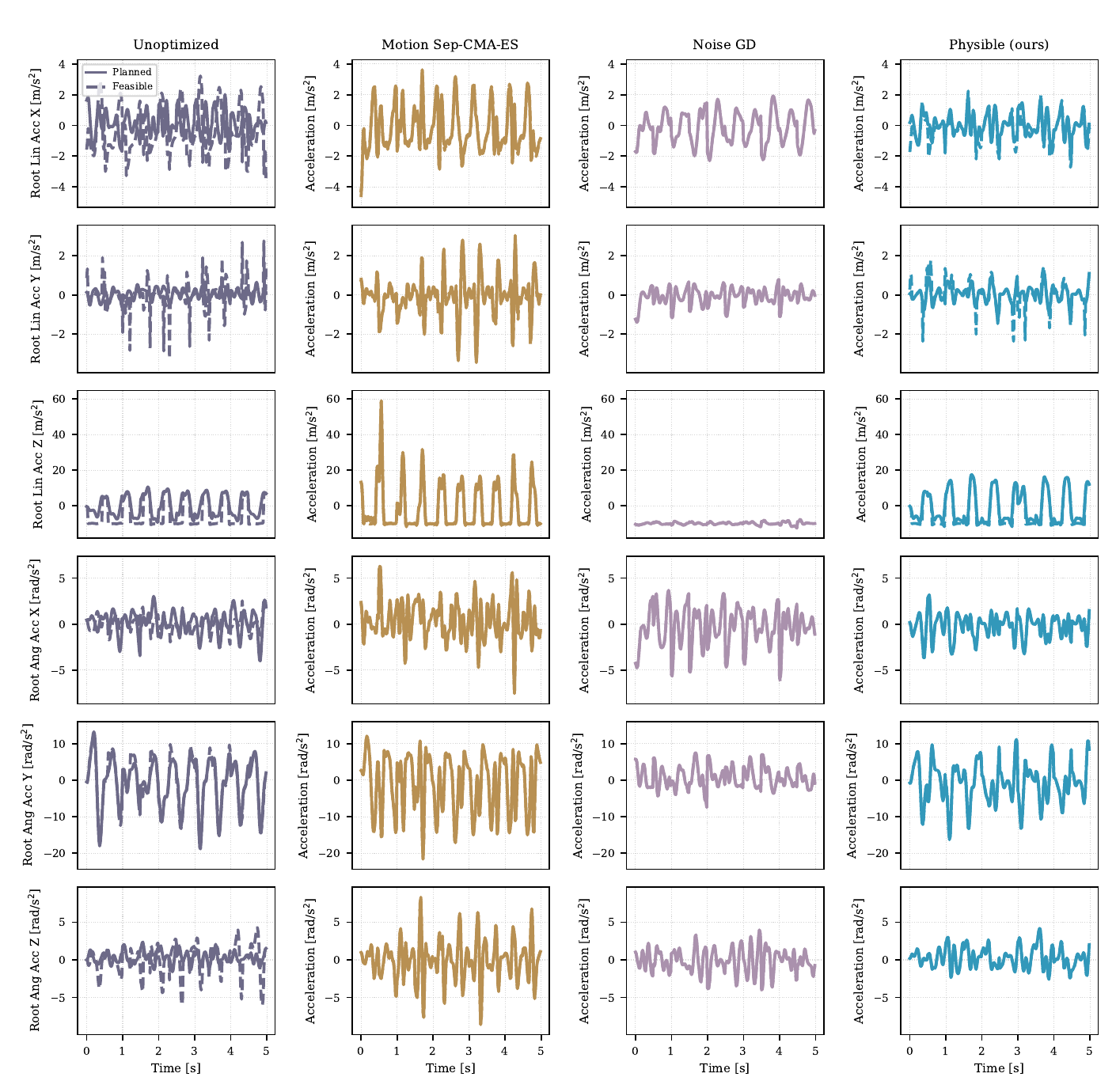}
    \caption{Behavior comparison of the \emph{root acceleration planned vs. feasible} by method for the dynamic feasibility optimization for $250$ time steps of the \emph{two-foot-jump} motion style on the bipedal robot.}
    \label{fig:dynoptim_results:behavior_comparison:bipedal_robot_twofootjump:root_acceleration_planned_vs_feasible}
\end{figure}

\begin{figure}[h!]
    \centering
    \includegraphics[width=1.0\linewidth, trim={5, 5, 5, 5}]{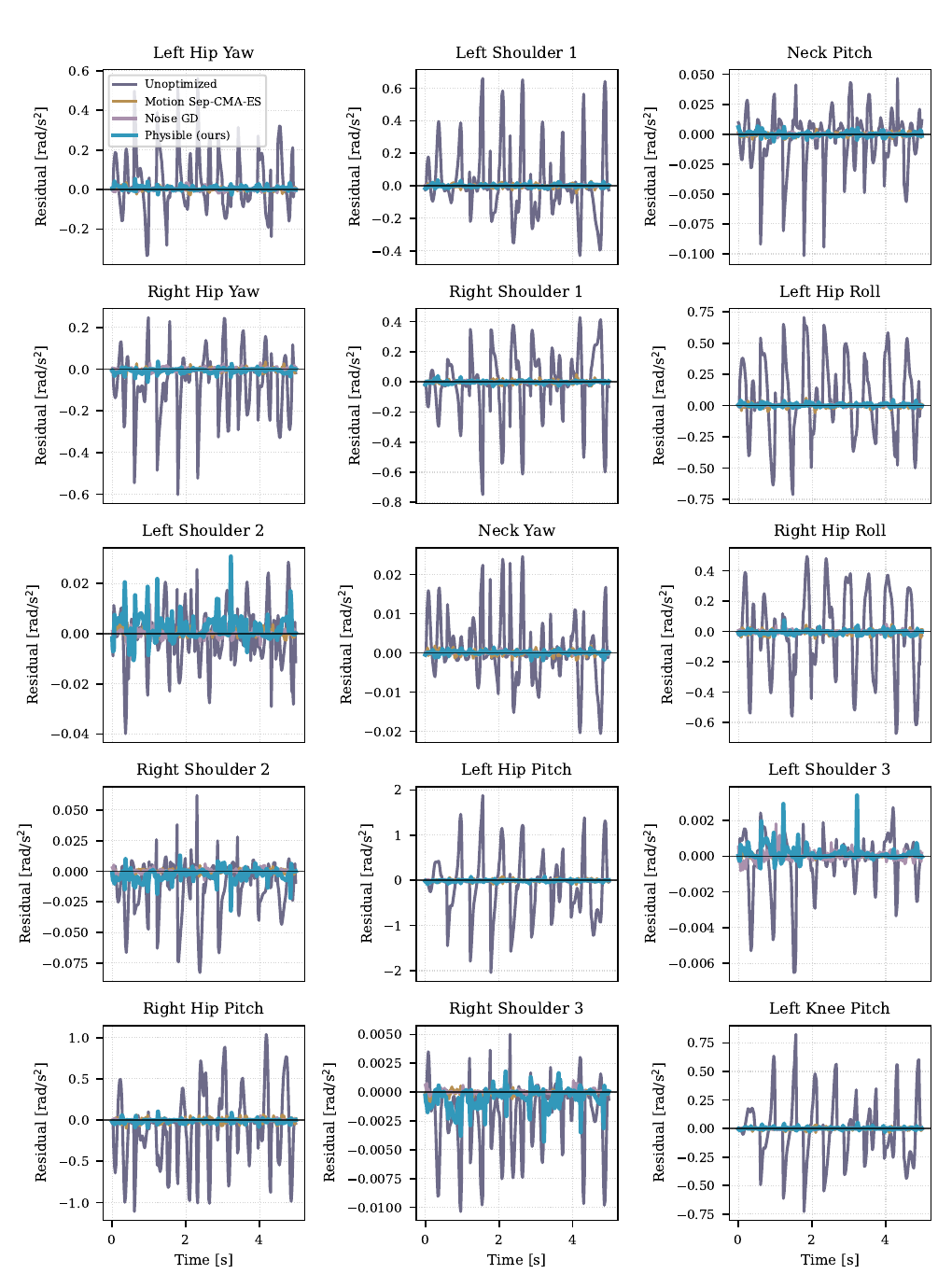}
    \caption{Behavior comparison of the \emph{joint acceleration residual} for the dynamic feasibility optimization for $250$ time steps of the \emph{two-foot-jump} motion style on the bipedal robot.}
    \label{fig:dynoptim_results:behavior_comparison:bipedal_robot_twofootjump:joint_acceleration_residual}
\end{figure}

\paragraph{Unoptimized vs. Optimized}
Physible improves dynamic consistency while retaining the characteristic jump trajectory and support phases. In \figref{fig:dynoptim-twofootjump}, the rendered poses and side-view root trajectories show that the optimized motion follows a similar jump arc to the unoptimized reference. The smaller virtual root forces and acceleration-residual arrows illustrate the reduction in nonphysical residual support and acceleration mismatch, particularly around takeoff and landing. The same improvement is visible over the full sequence in the virtual-wrench and root-acceleration-residual traces (Figs.~\ref{fig:dynoptim_results:behavior_comparison:bipedal_robot_twofootjump:virtual_wrench} and~\ref{fig:dynoptim_results:behavior_comparison:bipedal_robot_twofootjump:root_acceleration_residual}). The planned-versus-feasible acceleration curves in \figref{fig:dynoptim-twofootjump} and Figs.~\ref{fig:dynoptim_results:behavior_comparison:bipedal_robot_twofootjump:root_acceleration_planned_vs_feasible}--\ref{fig:dynoptim_results:behavior_comparison:bipedal_robot_twofootjump:joint_acceleration_planned_vs_feasible:part3} show closer agreement after optimization for the root and joints while preserving the temporal structure of the jump.

\begin{figure}[h!]
    \centering
    \includegraphics[page=1,width=1.0\linewidth, trim={5, 5, 5, 5}]{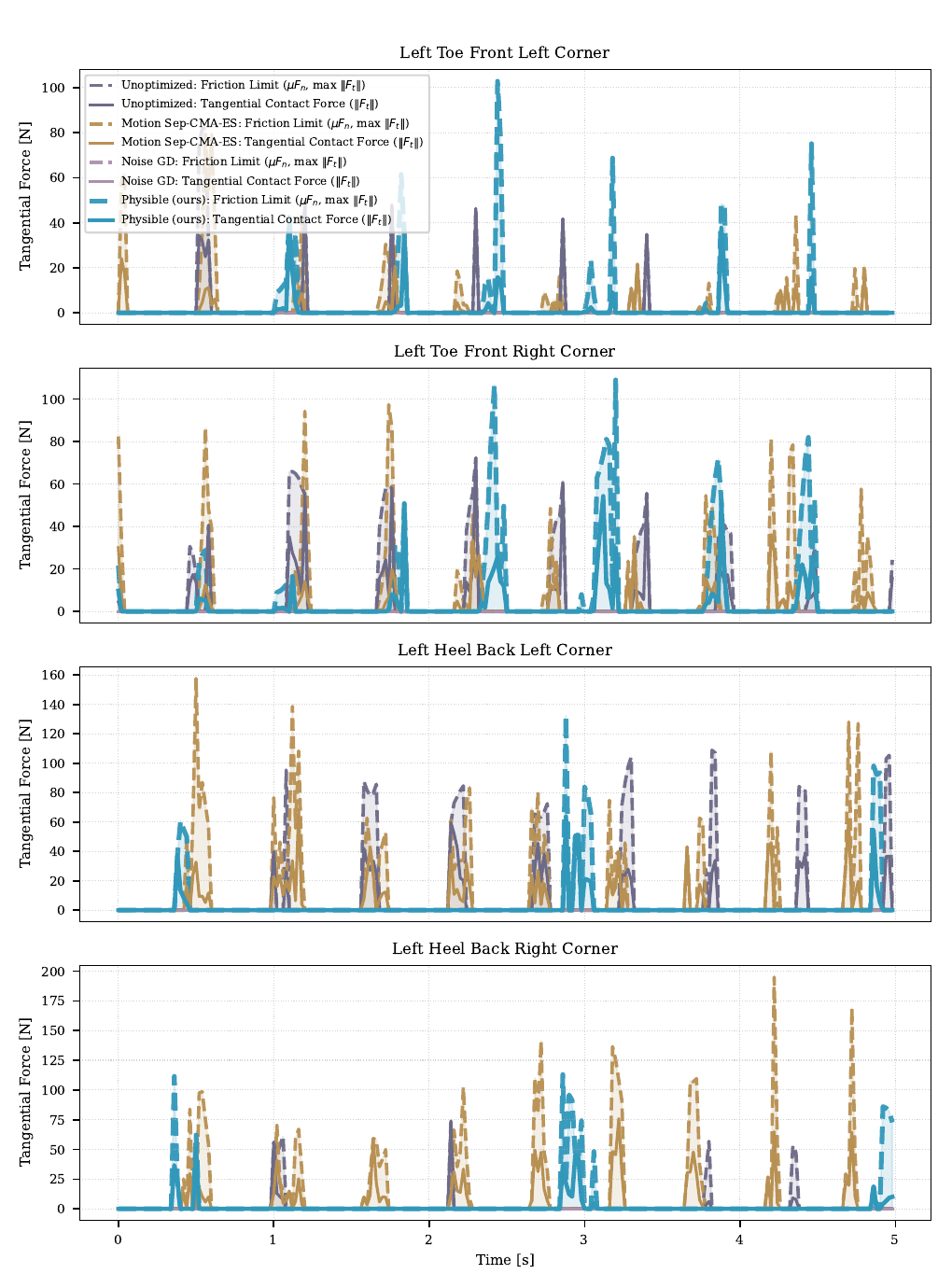}
    \caption{Behavior comparison of the \emph{tangential ground contact forces and the friction cone} for the dynamic feasibility optimization for $250$ time steps of the \emph{two-foot-jump} motion style on the bipedal robot.}
    \label{fig:dynoptim_results:behavior_comparison:bipedal_robot_twofootjump:friction_cone}
\end{figure}

These reductions in dynamic mismatch accompany repeated ground contacts during the support phases, as shown by the motion snapshots in \figref{fig:dynoptim-twofootjump} and the ground-separation traces in \figref{fig:dynoptim_results:behavior_comparison:bipedal_robot_twofootjump:ground_separation}. The slip and contact-force plots provide complementary views of the ground interaction (Figs.~\ref{fig:dynoptim_results:behavior_comparison:bipedal_robot_twofootjump:ground_slip} and~\ref{fig:dynoptim_results:behavior_comparison:bipedal_robot_twofootjump:friction_cone}), with contact forces remaining within the friction cones. Residual acceleration mismatch remains, but Physible retains the intended jumping behavior and support phases without an explicit contact-pattern objective.

\clearpage
\subsubsection{Left Hop: Unoptimized vs. Physible}
\label{apx:ssub:dynamic_lefthop}
The left-hop comparison covers $150$ frames (\SI{3}{s}) on the bipedal robot and compares unoptimized generation with Physible under asymmetric support. 
In \tabref{tab:dynoptim:sequence_comparison}, mean acceleration-reconstruction error decreases from $0.511$ to $0.009$, and the mean \ac{ID} residual decreases from $106.43$ to $1.91$, corresponding to reductions of approximately $98\%$ in both metrics.  This qualitative analysis reveal how these improvements arise and which aspects of the original motion change during refinement.

\begin{figure}[h!]
    \centering
    \includegraphics[width=1.0\linewidth, trim={5, 5, 5, 5}]{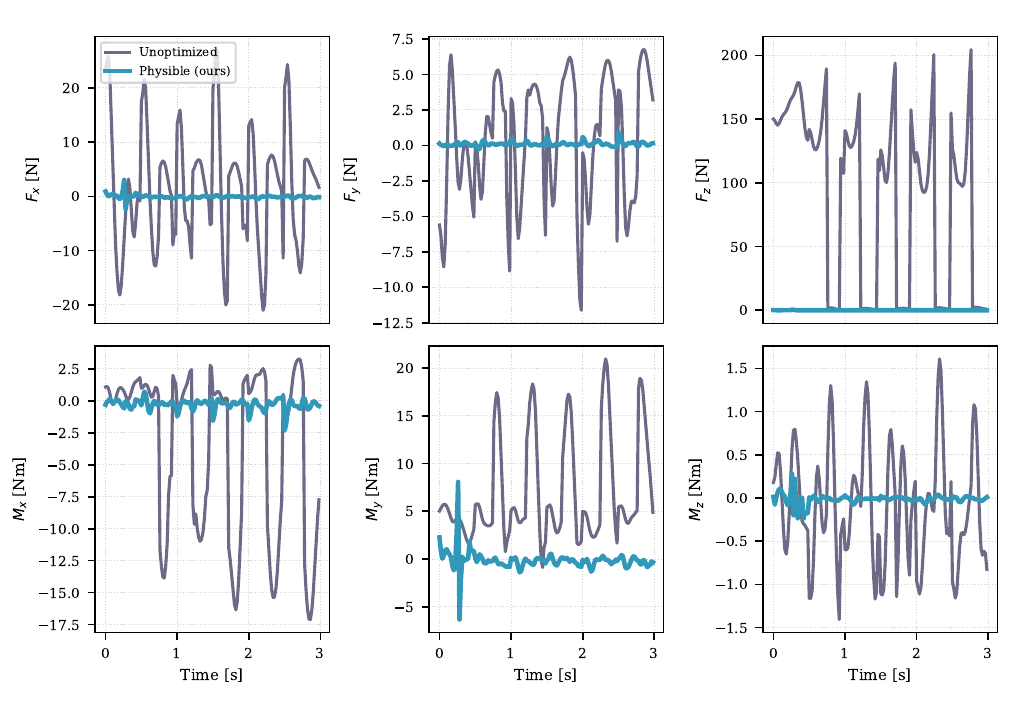}
    \caption{Behavior comparison of the \emph{virtual wrench} resulting from the relaxed \ac{ID} for the dynamic feasibility optimization for $150$ time steps of the \emph{left-hop} walking style on the bipedal robot.}
    \label{fig:dynoptim_results:behavior_comparison:bipedal_robot_lefthop:virtual_wrench}
\end{figure}

\begin{figure}[h!]
    \centering
    \includegraphics[width=1.0\linewidth, trim={5, 5, 5, 5}]{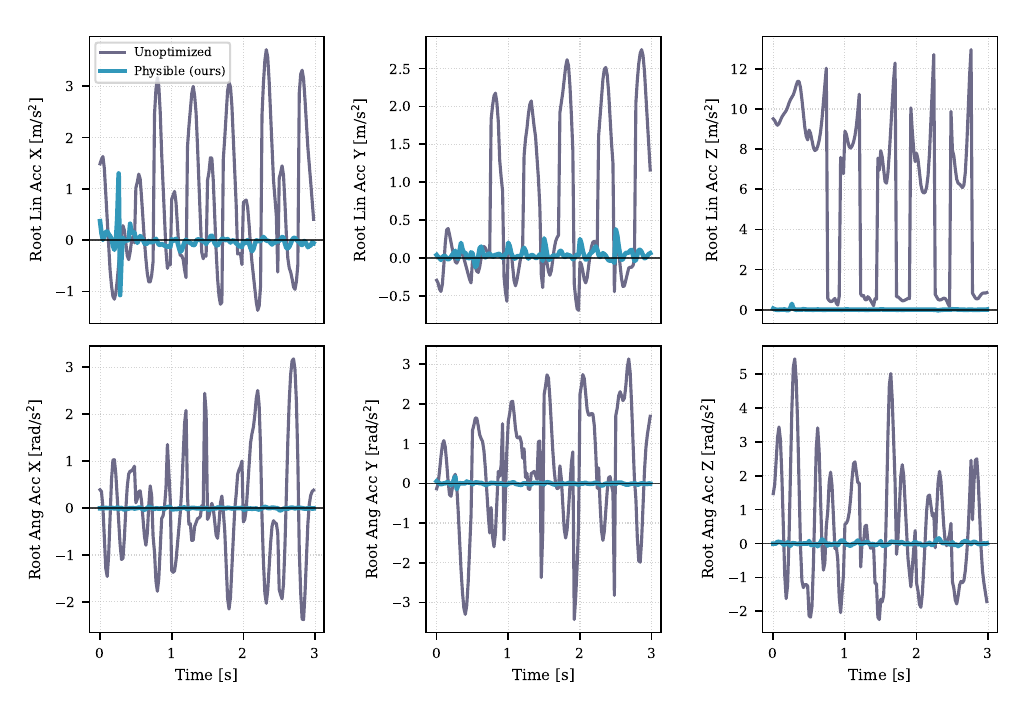}
    \caption{Behavior comparison of the \emph{root acceleration residual} for the dynamic feasibility optimization for $150$ time steps of the \emph{left-hop} walking style on the bipedal robot.}
    \label{fig:dynoptim_results:behavior_comparison:bipedal_robot_lefthop:root_acceleration_residual}
\end{figure}

\begin{figure}[h!]
    \centering
    \includegraphics[width=1.0\linewidth, trim={5, 5, 5, 5}]{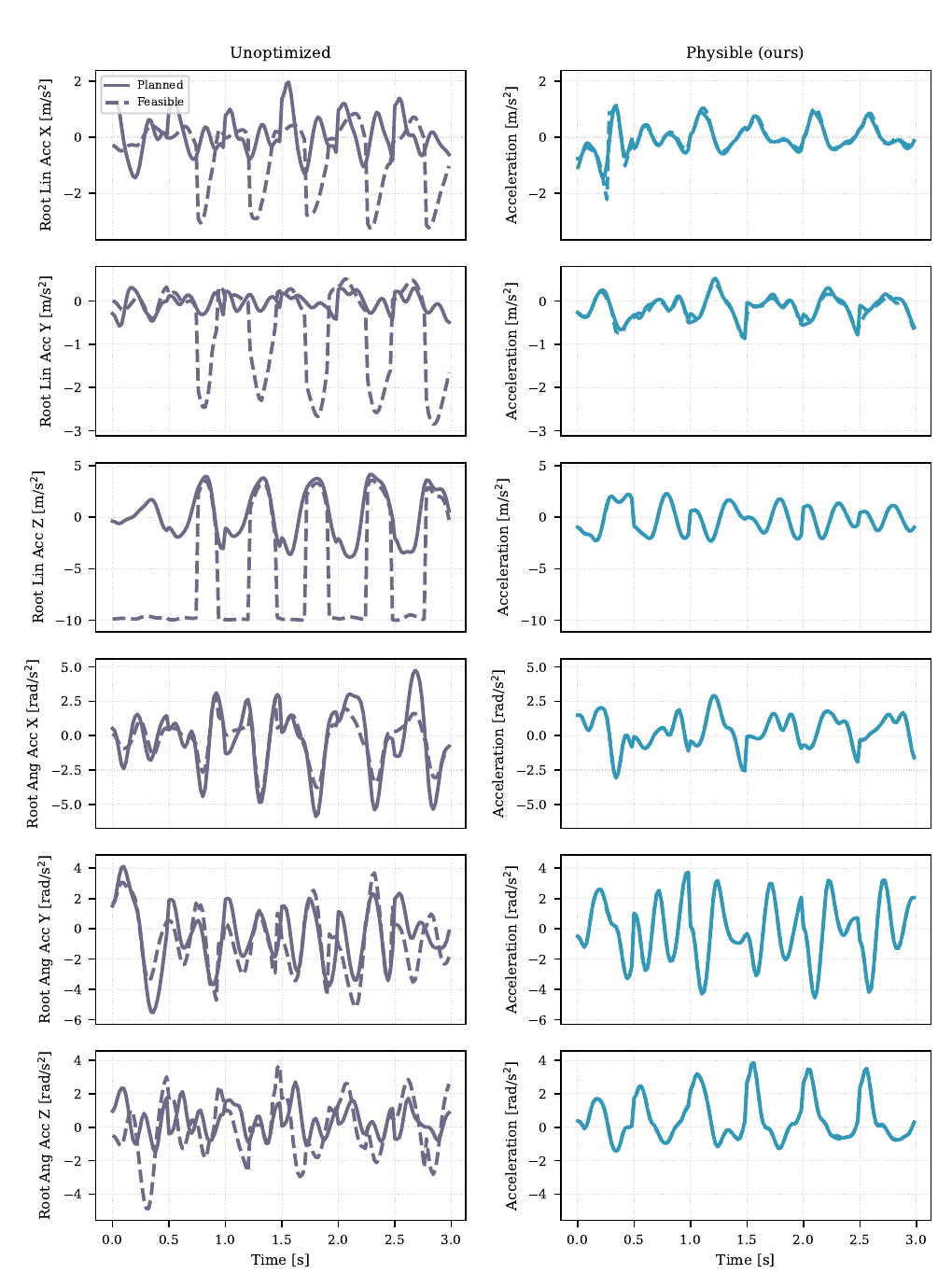}
    \caption{Behavior comparison of the \emph{root acceleration planned vs. feasible} for the dynamic feasibility optimization for $150$ time steps of the \emph{left-hop} walking style on the bipedal robot.}
    \label{fig:dynoptim_results:behavior_comparison:bipedal_robot_lefthop:root_acceleration_planned_vs_feasible}
\end{figure}

\begin{figure}[h!]
    \centering
    \includegraphics[width=1.0\linewidth, trim={5, 5, 5, 5}]{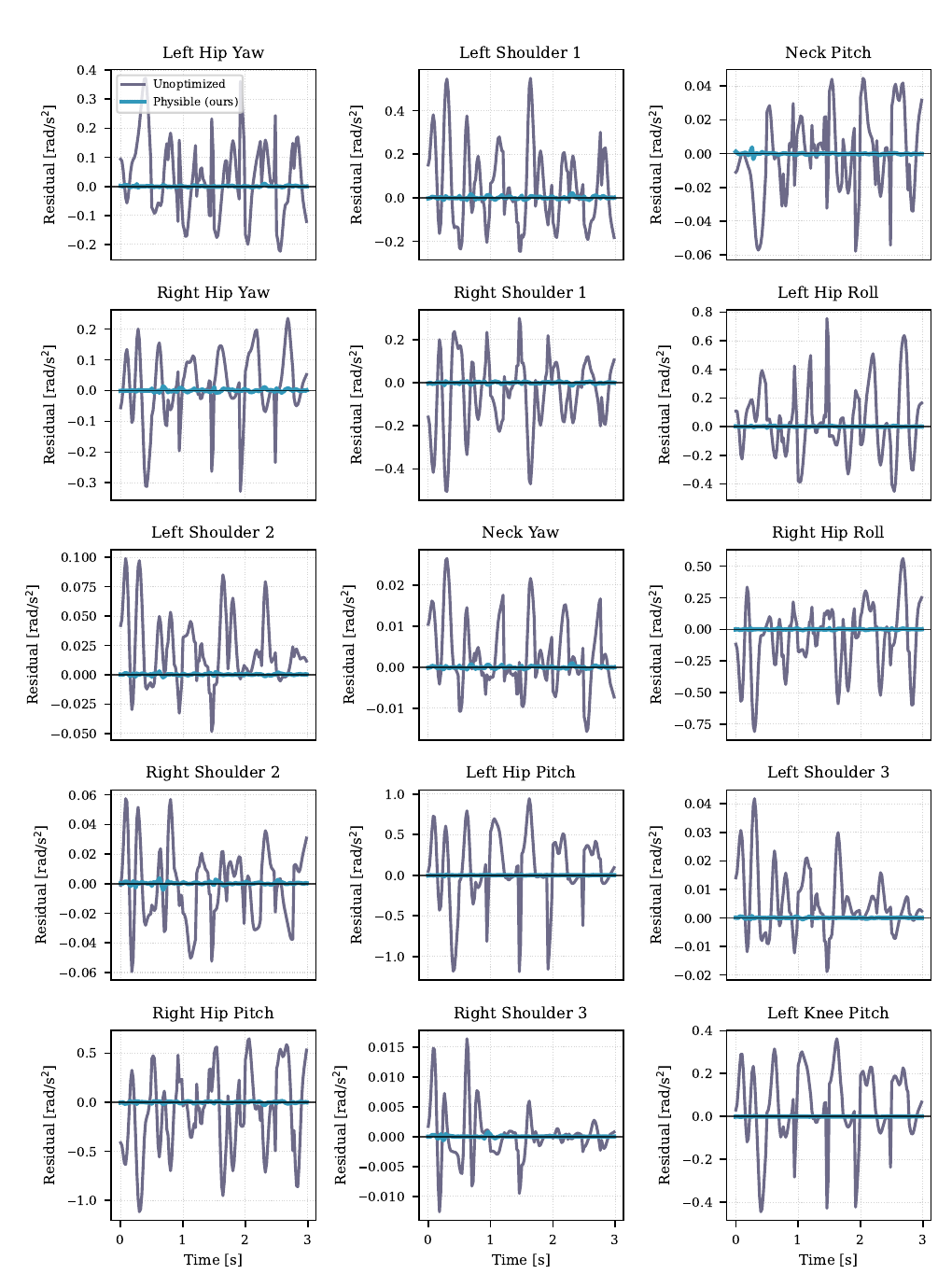}
    \caption{Behavior comparison of the \emph{joint acceleration residual} for the dynamic feasibility optimization for $150$ time steps of the \emph{left-hop} walking style on the bipedal robot.}
    \label{fig:dynoptim_results:behavior_comparison:bipedal_robot_lefthop:joint_acceleration_residual}
\end{figure}

\begin{figure}[h!]
    \centering
    \includegraphics[page=1,width=1.0\linewidth, trim={5, 5, 5, 5}]{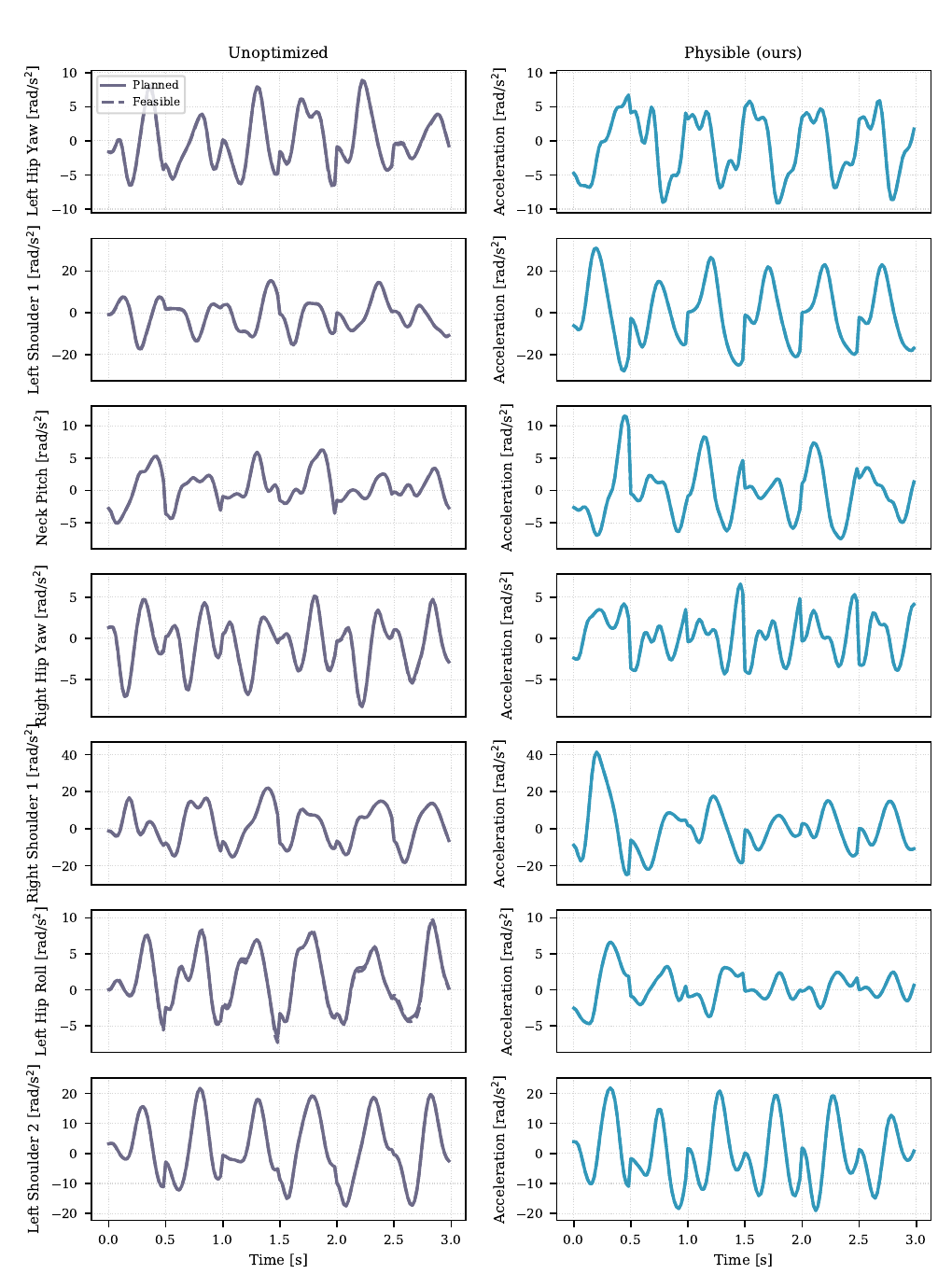}
    \caption{Behavior comparison of the \emph{joint acceleration planned vs. feasible} for the dynamic feasibility optimization for $150$ time steps of the \emph{left-hop} walking style on the bipedal robot - Part 1.}
    \label{fig:dynoptim_results:behavior_comparison:bipedal_robot_lefthop:joint_acceleration_planned_vs_feasible:part1}
\end{figure}

\begin{figure}[h!]
    \centering
    \includegraphics[page=2,width=1.0\linewidth, trim={5, 5, 5, 5}]{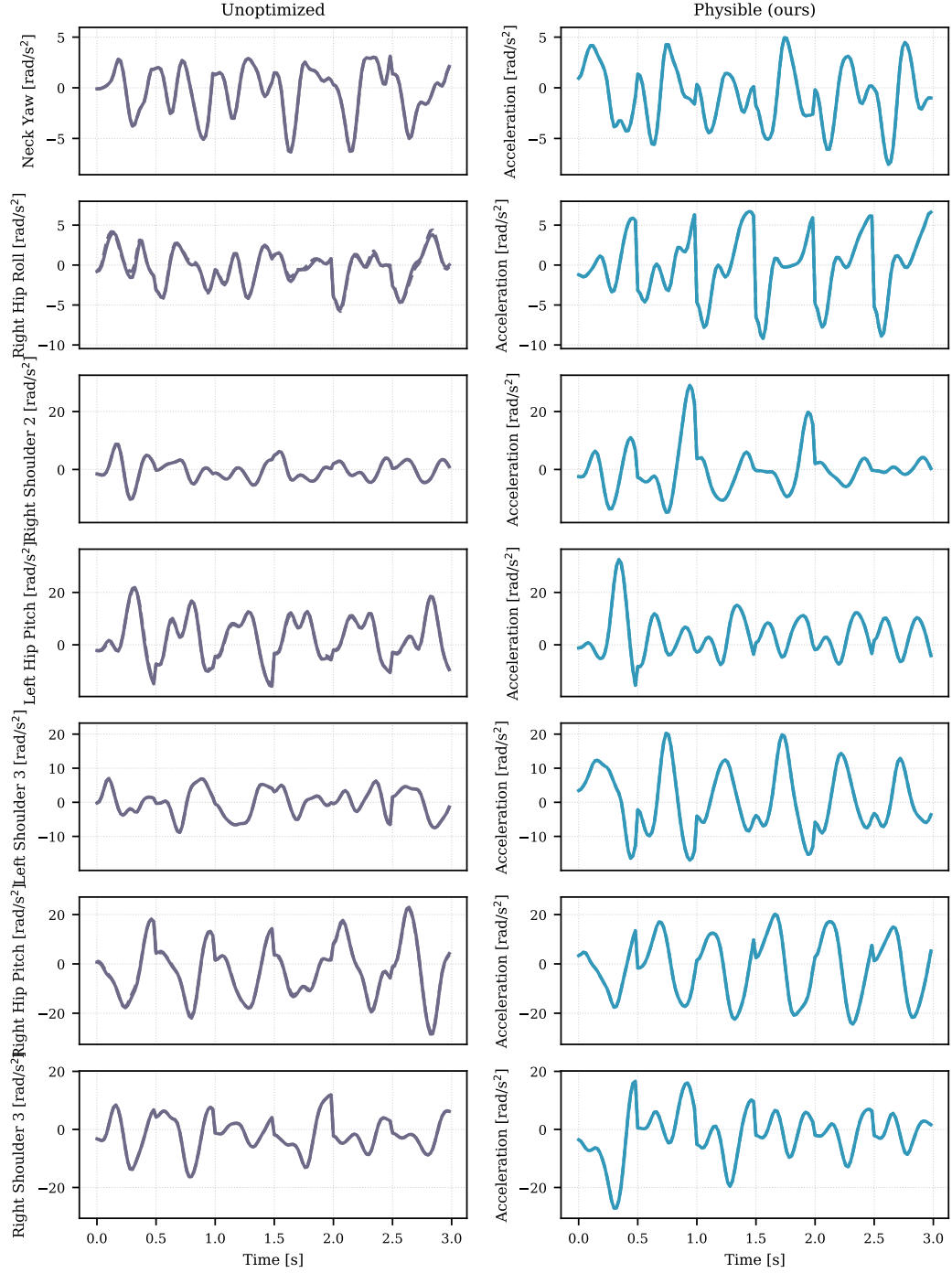}
    \caption{Behavior comparison of the \emph{joint acceleration planned vs. feasible} for the dynamic feasibility optimization for $150$ time steps of the \emph{left-hop} walking style on the bipedal robot - Part 2.}
    \label{fig:dynoptim_results:behavior_comparison:bipedal_robot_lefthop:joint_acceleration_planned_vs_feasible:part2}
\end{figure}

\begin{figure}[h!]
    \centering
    \includegraphics[page=3,width=1.0\linewidth, trim={5, 5, 5, 5}]{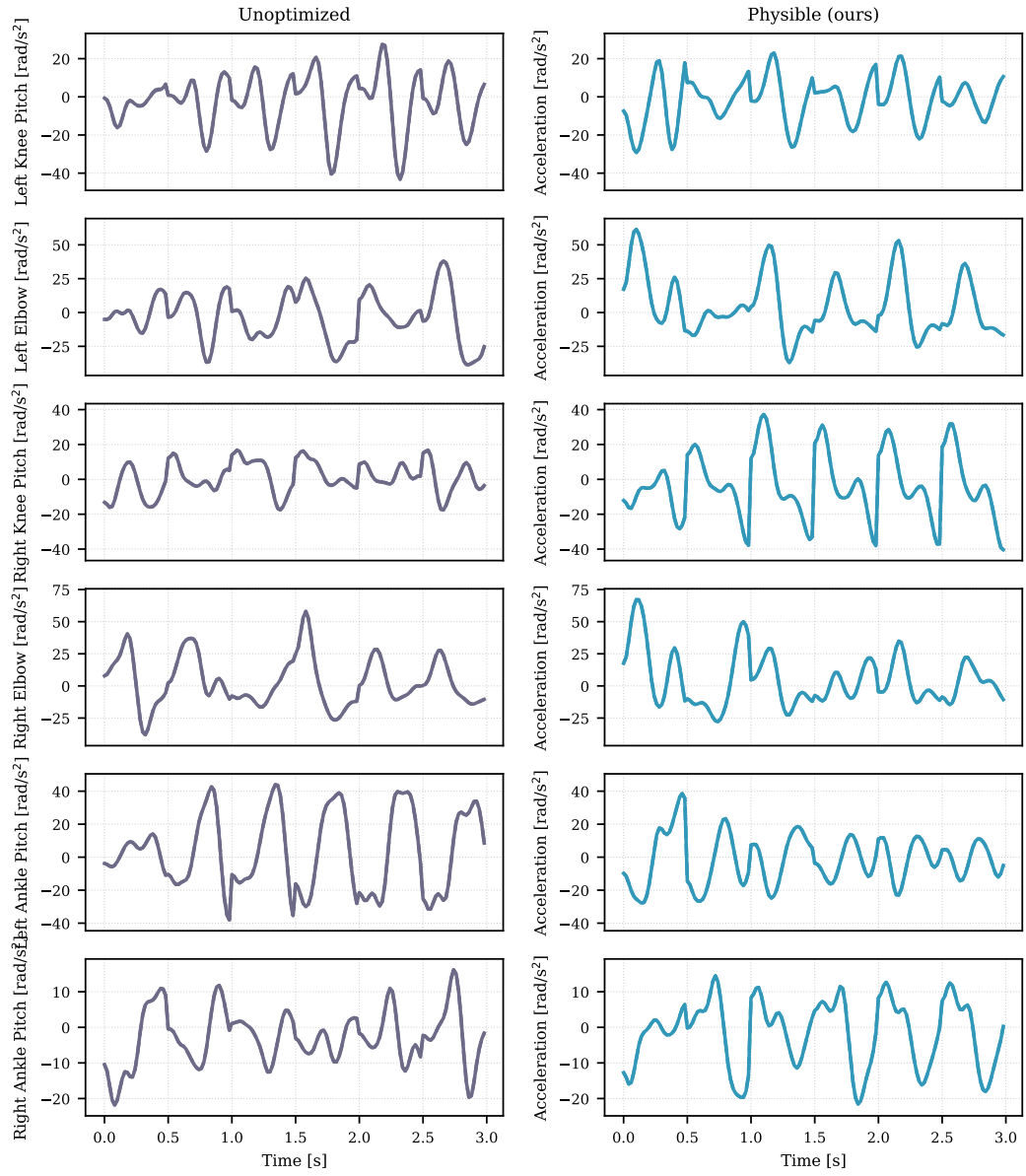}
    \caption{Behavior comparison of the \emph{joint acceleration planned vs. feasible} for the dynamic feasibility optimization for $150$ time steps of the \emph{left-hop} walking style on the bipedal robot - Part 3.}
    \label{fig:dynoptim_results:behavior_comparison:bipedal_robot_lefthop:joint_acceleration_planned_vs_feasible:part3}
\end{figure}

A prominent source of infeasibility in the unoptimized motion is the vertical root acceleration. During repeated intervals, the nominal inverse-dynamics solve reconstructs an acceleration close to free fall while the planned acceleration remains much higher (\figref{fig:dynoptim_results:behavior_comparison:bipedal_robot_lefthop:root_acceleration_planned_vs_feasible}). The resulting vertical acceleration residual repeatedly reaches approximately \SI{10}{\meter\per\second\squared} or more (\figref{fig:dynoptim_results:behavior_comparison:bipedal_robot_lefthop:root_acceleration_residual}). Consistently, the separate relaxed solve requires a large virtual upward force, often of order \SI{100}{N}, together with substantial corrective moments (\figref{fig:dynoptim_results:behavior_comparison:bipedal_robot_lefthop:virtual_wrench}). After optimization, the planned and reconstructed root accelerations nearly coincide, and the virtual force and moment traces remain close to zero over most of the sequence, apart from short transients. The joint-acceleration residuals also decrease markedly (\figref{fig:dynoptim_results:behavior_comparison:bipedal_robot_lefthop:joint_acceleration_residual}); the planned-versus-feasible joint plots retain oscillatory motion while showing changes in its amplitude and timing (Figs.~\ref{fig:dynoptim_results:behavior_comparison:bipedal_robot_lefthop:joint_acceleration_planned_vs_feasible:part1}--\ref{fig:dynoptim_results:behavior_comparison:bipedal_robot_lefthop:joint_acceleration_planned_vs_feasible:part3}).

\begin{figure}[h!]
    \centering
    \includegraphics[page=1,width=1.0\linewidth, trim={5, 5, 5, 5}]{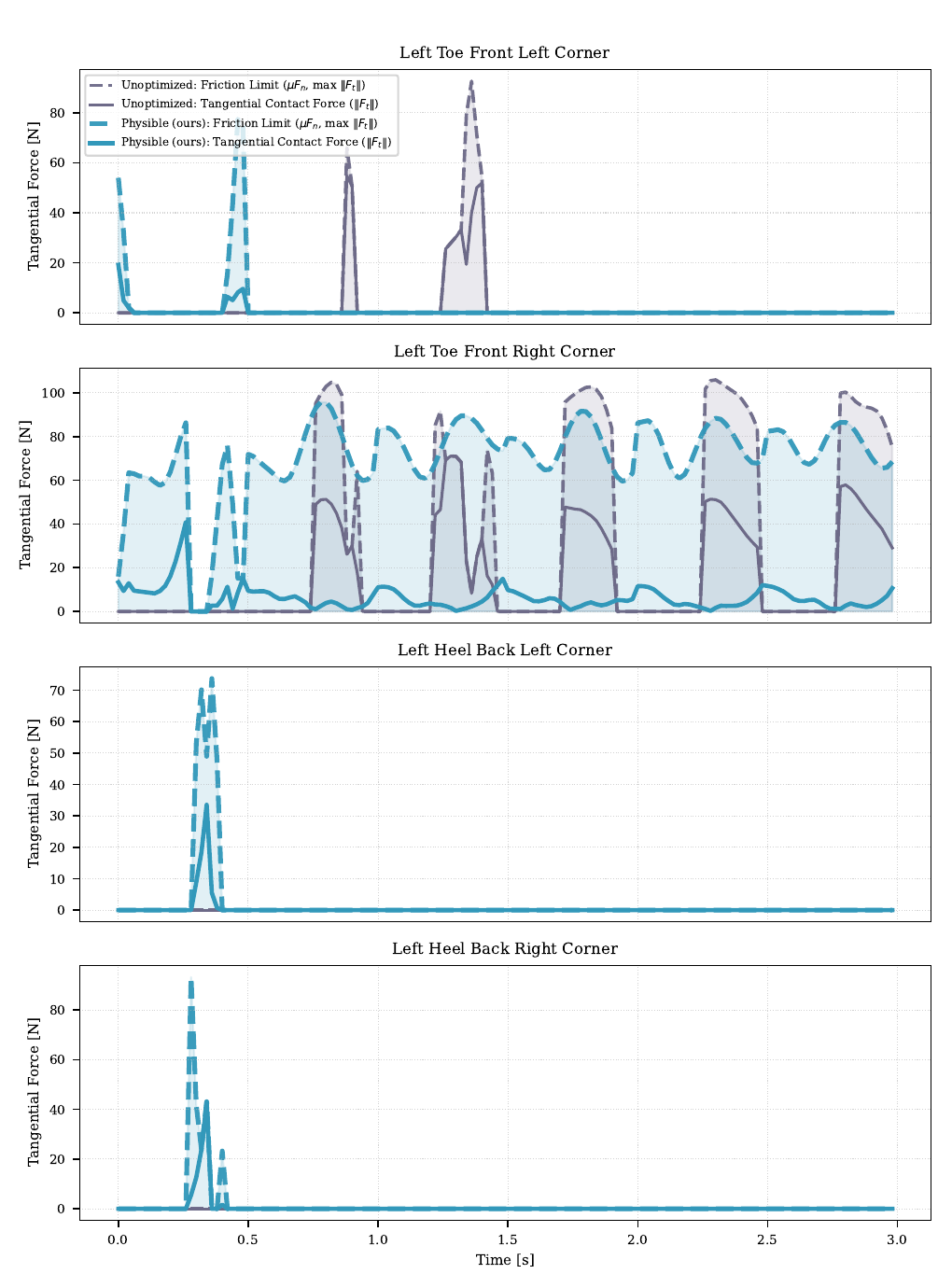}
    \caption{Behavior comparison of the \emph{tangential ground contact forces and the friction cone} for the dynamic feasibility optimization for $150$ time steps of the \emph{left-hop} walking style on the bipedal robot.}
    \label{fig:dynoptim_results:behavior_comparison:bipedal_robot_lefthop:friction_cone}
\end{figure}
\begin{figure}[h!]
    \centering
    \includegraphics[page=1,width=1.0\linewidth, trim={5, 5, 5, 5}]{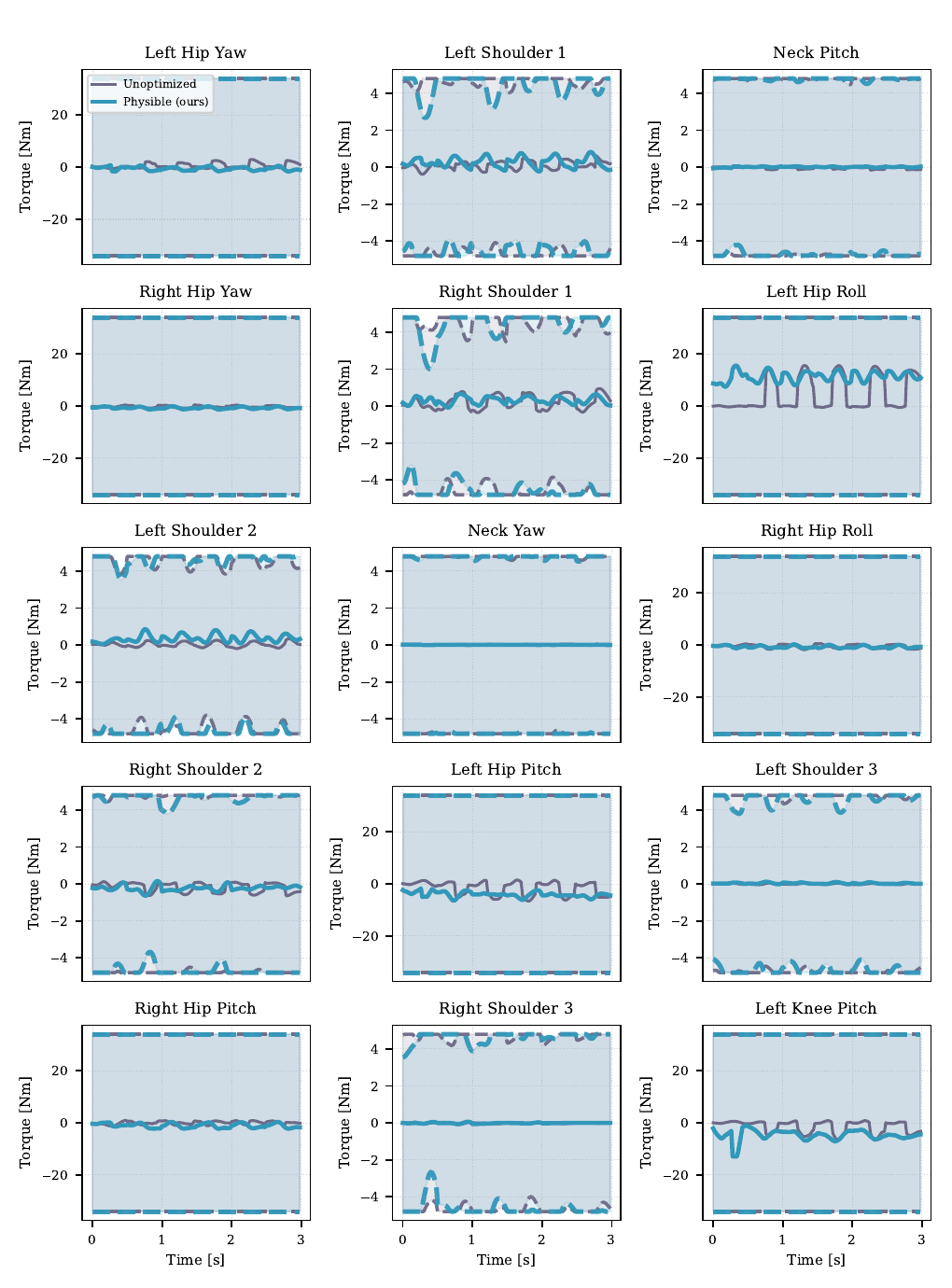}
    \caption{Behavior comparison of the \emph{joint torques (including limits)} for the dynamic feasibility optimization for $150$ time steps of the \emph{left-hop} walking style on the bipedal robot - Part 1.}
    \label{fig:dynoptim_results:behavior_comparison:bipedal_robot_lefthop:joint_torques:part1}
\end{figure}
\begin{figure}[h!]
    \centering
    \includegraphics[page=2,width=1.0\linewidth, trim={5, 5, 5, 5}]{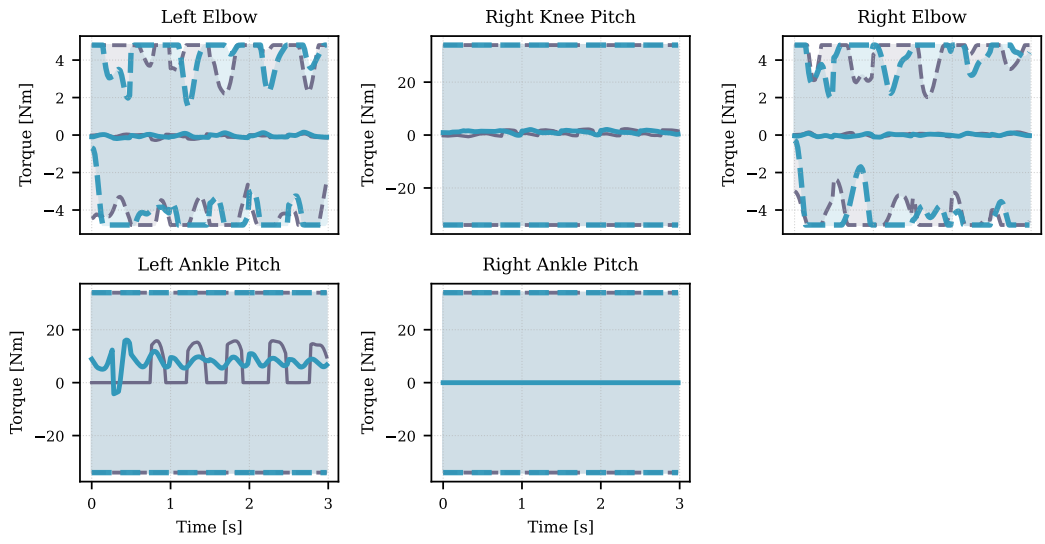}
    \caption{Behavior comparison of the \emph{joint torques (including limits)} for the dynamic feasibility optimization for $150$ time steps of the \emph{left-hop} walking style on the bipedal robot - Part 2.}
    \label{fig:dynoptim_results:behavior_comparison:bipedal_robot_lefthop:joint_torques:part2}
\end{figure}

During sustained toe support, the reconstructed tangential force remains well below its friction limit, and the joint torques remain within the displayed actuator bounds (\figref{fig:dynoptim_results:behavior_comparison:bipedal_robot_lefthop:friction_cone} and Figs.~\ref{fig:dynoptim_results:behavior_comparison:bipedal_robot_lefthop:joint_torques:part1}--\ref{fig:dynoptim_results:behavior_comparison:bipedal_robot_lefthop:joint_torques:part2}). The optimized left hip-roll and ankle-pitch torques become more sustained, consistent with the longer support intervals. 


\begin{figure}[h!]
    \centering
    \includegraphics[width=1.0\linewidth, trim={5, 5, 5, 5}]{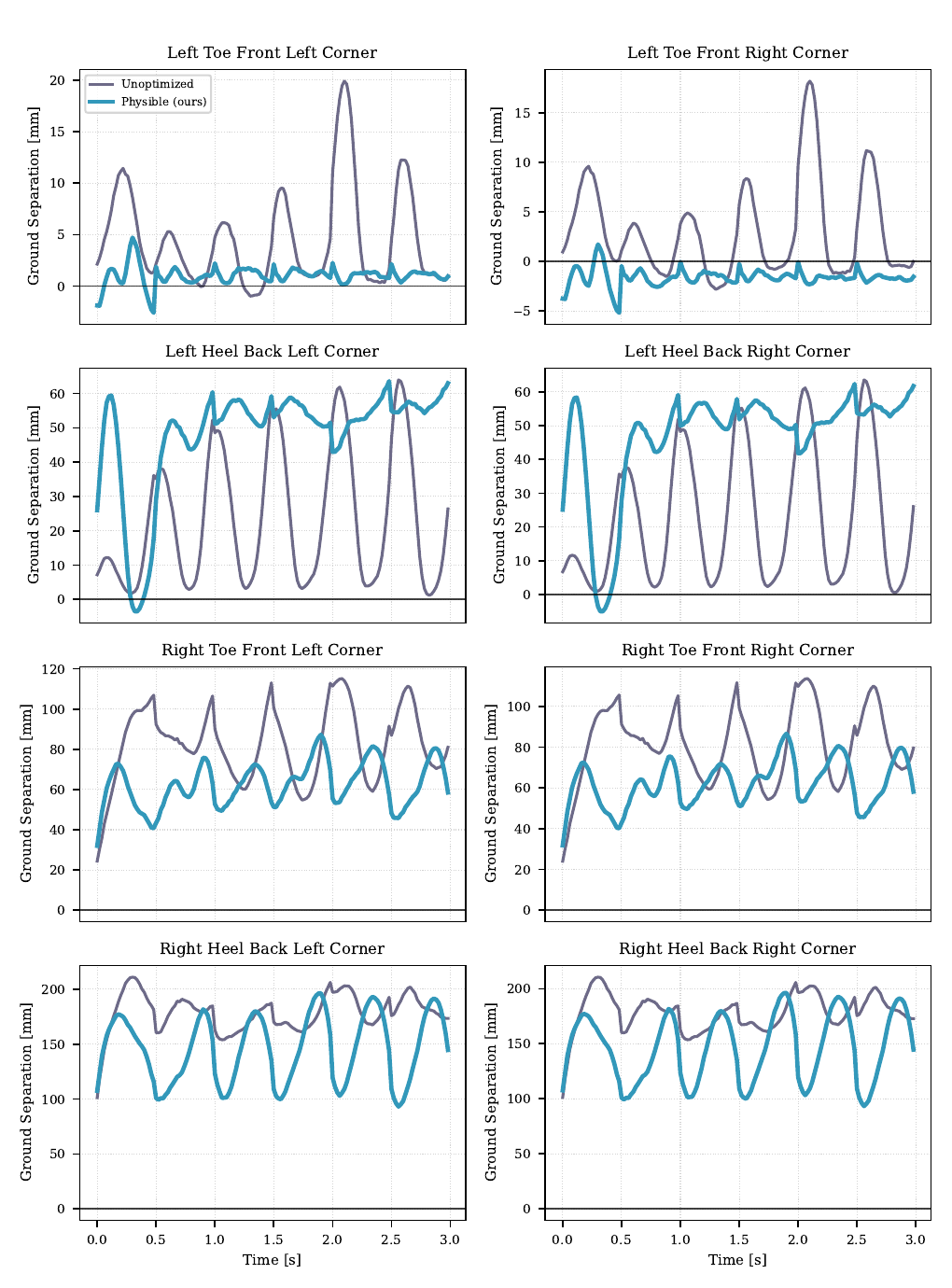}
    \caption{Behavior comparison of the \emph{ground separation} for the dynamic feasibility optimization for $150$ time steps of the \emph{left-hop} walking style on the bipedal robot.}
    \label{fig:dynoptim_results:behavior_comparison:bipedal_robot_lefthop:ground_separation}
\end{figure}

\begin{figure}[h!]
    \centering
    \includegraphics[width=1.0\linewidth, trim={5, 5, 5, 5}]{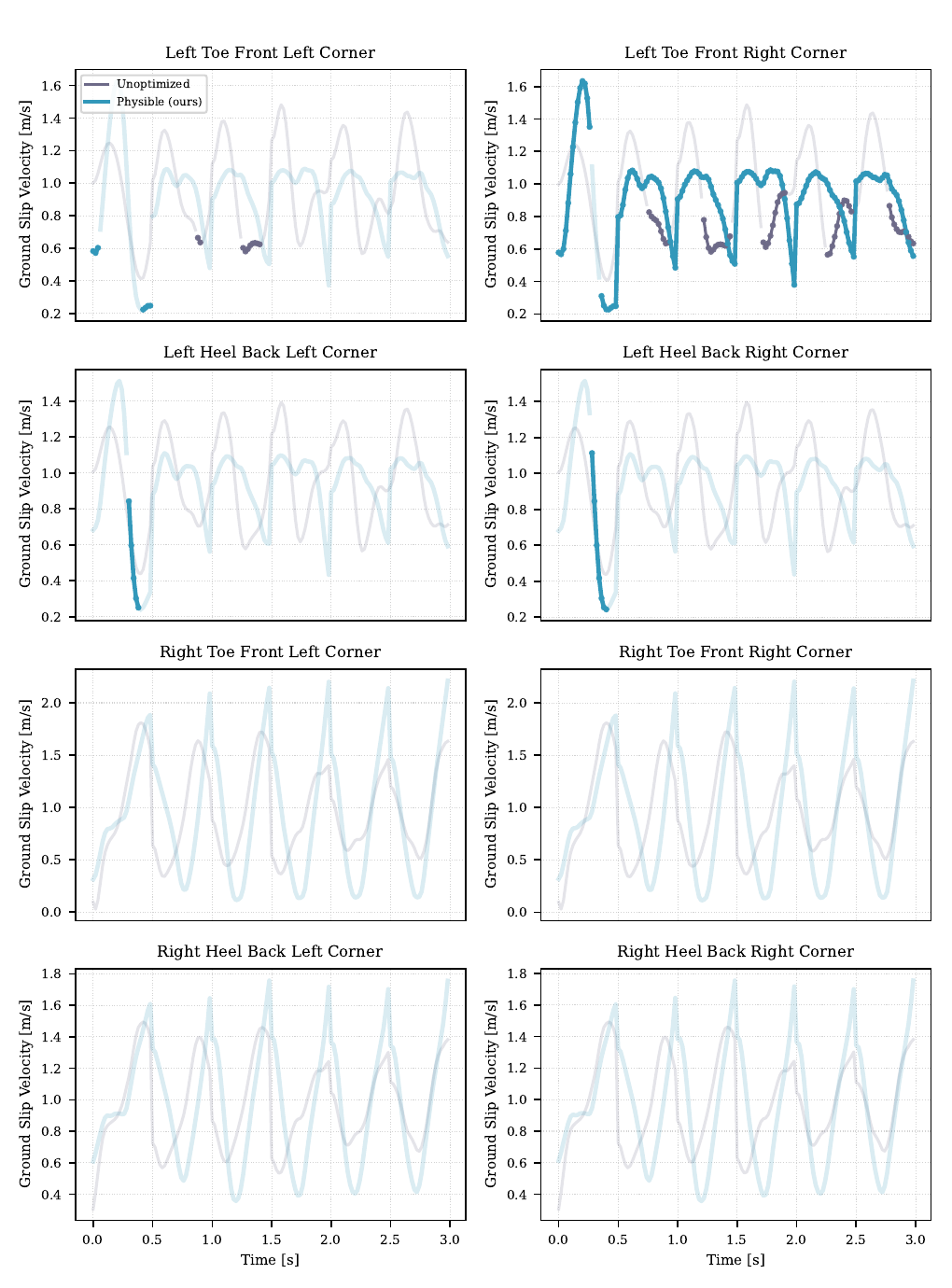}
    \caption{Behavior comparison of the \emph{ground slip velocity} for the dynamic feasibility optimization for $150$ time steps of the \emph{left-hop} walking style on the bipedal robot.}
    \label{fig:dynoptim_results:behavior_comparison:bipedal_robot_lefthop:ground_slip}
\end{figure}

\clearpage
\subsubsection{Uphill and Downhill Walking: Unoptimized vs. Physible}
\label{apx:sub:slope_dynamic_feasibility}
\begin{figure}[p]
    \centering
    \includegraphics[width=\linewidth]{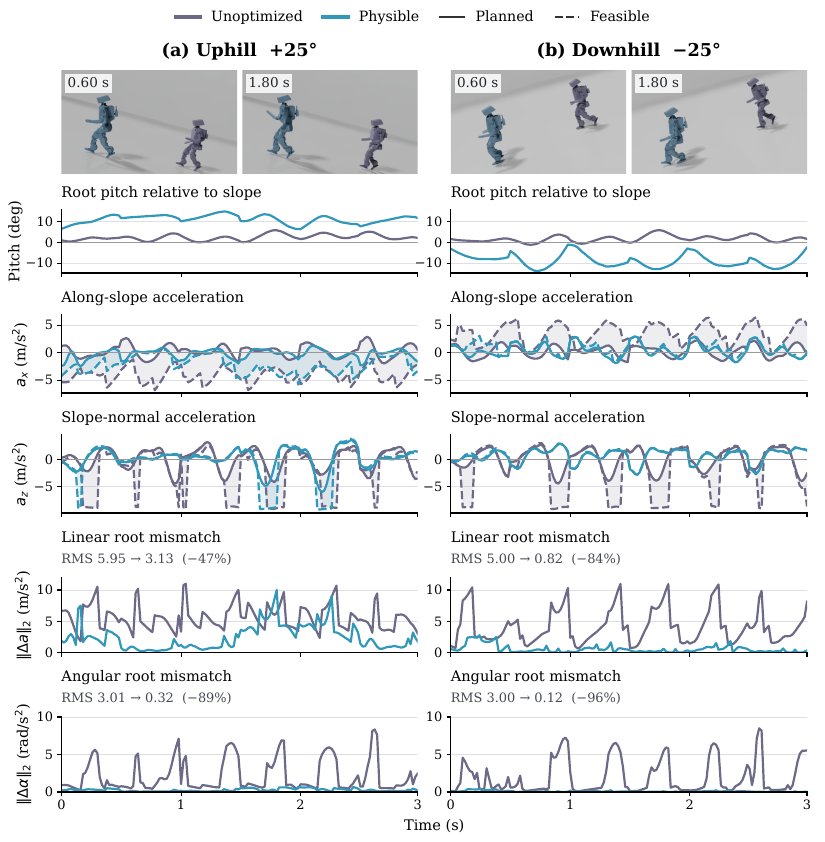}
    \caption{Dynamic feasibility optimization of the bipedal robot in the \emph{neutral} style for (a) uphill ($+25^\circ$) and (b) downhill ($-25^\circ$) motion. Purple denotes unoptimized motion and blue denotes Physible. The snapshots show each pair at video timestamps $0.60$ and $1.80\,\mathrm{s}$. All quantities use the slope-aligned frame with rotated gravity. From top to bottom, the plots show root pitch, along-slope and slope-normal accelerations (solid: planned; dashed: feasible reconstruction), and the norms of the linear and angular root acceleration residuals. Shading between acceleration curves highlights their mismatch. RMS annotations summarize each residual norm over the full sequence.}
    \label{fig:dynoptim-slopes}
\end{figure}

We compare unoptimized and Physible-optimized bipedal robot motions in the \emph{neutral} style for uphill ($+25^\circ$) and downhill ($-25^\circ$) scenarios in \figref{fig:dynoptim-slopes}. Each sequence spans \SI{3}{s} at \SI{50}{Hz}, with friction coefficient $\mu_{\mathrm c,p}=1.0$. The fixed command and experiment settings are given in Appendix~\ref{apx:par:slope_settings}. The renderings and root pitch use a slope-aligned coordinate frame: $x$ points along the direction of travel and $z$ is normal to the surface, so the horizontal ground plane in the renderings represents the slope.

We evaluate the mismatch between planned and feasible reconstructed root accelerations using the Euclidean norms of the three-dimensional linear and angular residuals, $\|\Delta\mathbf a\|_2$ and $\|\Delta\bm\alpha\|_2$. Their root-mean-square values over each sequence decrease from $5.95$ to \SI{3.13}{\meter\per\second\squared} for uphill motion (47\%) and from $5.00$ to \SI{0.82}{\meter\per\second\squared} for downhill motion (84\%). The corresponding angular mismatch decreases from $3.01$ to \SI{0.32}{\radian\per\second\squared} (89\%) and from $3.00$ to \SI{0.12}{\radian\per\second\squared} (96\%), respectively.

Optimization produces opposite pitch adaptations in the two scenarios: root pitch shifts toward positive values uphill and negative values downhill. The planned and reconstructed accelerations agree more closely after optimization, especially downhill. The uphill sequence retains appreciable linear mismatch, visible in both the along-slope and slope-normal components. These examples therefore show improved dynamic consistency under the slope-specific gravity vectors.

\clearpage
\subsection{Downstream RL Tracking}
\label{apx:sub:rl_tracking_results}
We evaluate downstream tracking of the unoptimized and Physible-optimized bipedal two-foot-jump motions using the protocol in Appendix~\ref{apx:sub:rl_tracking_protocol}.
Relative to the policy trained on the unoptimized motion, the policy trained on the optimized motion reduces body position, body orientation, and joint position tracking error by 16.4\%, 11.9\%, and 7.4\%, respectively, indicating that the increased dynamical consistency makes the reference easier to track. In contrast, global root position and orientation error increase by 39.6\% and 41.3\%. We attribute this to the step-wise nature of our dynamic feasibility optimization: accelerations are corrected locally in time rather than through the resulting integrated trajectory, so small residual inconsistencies accumulate as drift in the global root motion, while the more local and repetitive joint and body motion is largely unaffected. Consistent with this interpretation, the RL reward tracks the motion-frame root pose together with root-relative body motion, but it does not explicitly penalize accumulated global root-pose drift. The tracking results therefore show that reducing local dynamic residuals can improve body and joint tracking without improving every aspect of the complete trajectory.

\clearpage
\section{Detailed Limitations and Future Work}
\label{apx:sec:limitations}

Our evaluation is currently limited to bipedal robotic embodiments. Future work should study additional robot topologies and embodiments (e.g., manipulators, quadrupeds, and mobile manipulators).

The method further relies on accurate knowledge of robot dynamics, inertial parameters, contact geometry, and contact characteristics (e.g., friction). In practice, these quantities are uncertain and time-varying, and perception errors can directly affect contact reasoning. A natural next step is to incorporate uncertainty into the optimization (e.g., robust/risk-sensitive feasibility) and to evaluate how performance degrades under dynamics/contact mismatch.

Moreover, we currently assume that all ground contacts are sticking. While this simplifies \ac{ID} and feasibility evaluation, it rules out behaviors that deliberately leverage controlled slipping (e.g., moonwalk-like styles). Future work could therefore co-optimize contact-mode decisions jointly with the motion.

Relatedly, we infer the enforced contact pattern for the \emph{contact pattern preservation} constraint via a majority vote over a batch of unbiased denoised motion sequences. Although effective in practice, this choice is restrictive, as it discourages substantial changes to the initial contact pattern during optimization. Future work should investigate alternatives that simultaneously mitigate ground penetration and ``floating'' artifacts in \emph{kinematic feasibility} optimization without overly constraining contact-pattern changes.

More broadly, the proposed optimization does not guarantee that the injected/optimized noise remains within the diffusion model's training distribution. Therefore, a common failure mode of the optimization is a \emph{style collapse} where objectives and constraints might be satisfied, but the optimized motion does not adhere to the commanded motion style anymore. Beyond noise normalization, future work should investigate additional distribution-preserving mechanisms to better keep optimized motions in-distribution while still improving feasibility.

Furthermore, the approach requires trading off multiple (sometimes competing) objectives and constraints, which can make it challenging to identify suitable weights in the fitness function. In addition, achieving strict satisfaction of physical constraints would require many iterations and can push the optimization into regions where the diffusion model drifts out-of-distribution, degrading motion style.
Future work should therefore explore enforcing physical feasibility constraints earlier in the pipeline---including during motion retargeting and in the construction/selection of the training dataset---so that the unoptimized diffusion output is closer to physical feasibility and requires smaller inference-time corrections.

Building on the previous point, a particularly tricky issue is \emph{ground penetration}. Physically feasible motion (e.g., non-floating behavior, low tangential \emph{slip} velocities near the ground, and dynamically feasible accelerations) relies on active contacts for constraint enforcement. While, in principle, one would like \emph{exactly zero} penetration—ensuring both contact and no interpenetration—this is in practice generally infeasible from an optimization standpoint. We therefore introduce a \emph{contact-pattern preservation} constraint that permits a limited amount of penetration. However, the appropriate penetration threshold is not obvious: setting it too low can substantially increase problem difficulty and slow convergence. Future work should develop more synergistic approaches that better resolve this physical-feasibility ``tension''.

From a practical standpoint, repeated horizon-wide feasibility evaluations (including inner-loop solves of the \ac{ID}) can be computationally expensive for long horizons or real-time use. Promising directions include warm-starting (possibly with a learned modification of the diffusion noise~\citep{eyring2025noise}), multi-resolution optimization~\citep{antonova2023rethinking}, and amortized surrogates for (parts of) the feasibility evaluation to reduce wall-clock time without sacrificing reliability.

\end{document}